%% file: main.tex
\documentclass[acmsmall,nonacm]{acmart}
\usepackage[utf8]{inputenc}

\usepackage{subcaption}
\usepackage{enumitem}

\usepackage{xspace} 

\newcounter{rq}

\newcommand{\ours}{\textsc{GraphMERT}\xspace}

\begin{document}

\title{An Alternative Trajectory for Generative AI}

\author{Margarita Belova}
\authornote{Equal contributions: These author names are listed in alphabetical order.}
\authornote{Corresponding author.}
\email{margarita.bel@princeton.edu}
\affiliation{
  \institution{Princeton University}
  \city{Princeton}
  \state{NJ}
  \country{USA}
}

\author{Yuval Kansal}
\authornotemark[1]
\email{yuvalkansal@princeton.edu}
\affiliation{
  \institution{Princeton University}
  \city{Princeton}
  \state{NJ}
  \country{USA}
}

\author{Yihao Liang}
\authornotemark[1]
\email{yhliang@princeton.edu}
\affiliation{
  \institution{Princeton University}
  \city{Princeton}
  \state{NJ}
  \country{USA}
}

\author{Jiaxin Xiao}
\authornotemark[1]
\email{jx0800@princeton.edu}
\affiliation{
  \institution{Princeton University}
  \city{Princeton}
  \state{NJ}
  \country{USA}
}

\author{Niraj K. Jha}
\email{jha@princeton.edu}
\affiliation{
  \institution{Princeton University}
  \city{Princeton}
  \state{NJ}
  \country{USA}
}

\input{sections/0_Abstract}



\maketitle

\input{sections/0_Introduction}
\input{sections/2_Good_data}
\input{sections/3_Abstraction}
\input{sections/4_Reasoning}
\input{sections/6_Neurosymbolic_AI}
\input{sections/7_DSS_}

\input{sections/8_DSS_society_new}
\input{sections/9_Agents}
\input{sections/10_Energy}
\input{sections/11_Goal}
\input{sections/1_AGI}
\input{sections/Conclusion}

\section*{Author contributions}Margarita Belova: Investigation, Writing – original draft, Writing – review \& editing. Yuval Kansal: Investigation, Writing – original draft, Writing – review \& editing. Yihao Liang: Investigation, Writing – original draft, Writing – review \& editing. Jiaxin Xiao: Investigation, Writing – original draft, Writing – review \& editing. Niraj K. Jha: Project administration, Supervision, Writing – review \& editing.

\section*{Funding}
This research was not supported by any specific grant from funding agencies in the public, commercial, or not-for-profit sectors.

\section*{Declaration of competing interests}
The authors declare competing financial interests. Patent applications have been filed related to building blocks of domain-specific superintelligence, including knowledge graph extraction, knowledge-graph-based synthetic curricula generation, and fine-tuning and reinforcement learning of language models based on such synthetic curricula.

\bibliographystyle{ACM-Reference-Format}
\bibliography{reference}

\end{document}

%% file: sections/0_Abstract.tex
\begin{abstract}
The generative artificial intelligence (AI) ecosystem is undergoing rapid transformations that threaten its sustainability. As models transition from research prototypes to high-traffic products, the dominant energetic burden has shifted from one-time training to recurring, unbounded inference—a phenomenon exacerbated by the emergence of ``reasoning models'' that inflate compute costs by orders of magnitude per query. The prevailing trajectory, i.e, pursuit of artificial general intelligence through continued scaling of monolithic models, is colliding with hard physical constraints: localized grid failures, prohibitive water consumption, and diminishing returns on data scaling. Even though this trajectory yields models with impressive factual recall, it struggles in domains that require in-depth verifiable reasoning. A possible cause is a lack of sufficient abstractions in the underlying training data.

 Current large language models (LLMs) exhibit genuine reasoning depth only in domains like mathematics and coding, where rigorous, pre-existing abstractions provide structural grounding. In other fields that lack such abstractions, the current approach fails to generalize well. We propose an alternative trajectory based on \textit{domain-specific superintelligence (DSS)}. We argue that to achieve robust reasoning in open-world domains, it is advantageous to first construct explicit symbolic abstractions, such as knowledge graphs, ontologies, and formal logic. These abstractions can form the basis for generation of synthetic curricula, training on which can enable small language models to master complex domain-specific reasoning, without encountering the model collapse problem that is typical of current synthetic data methods that use an LLM to supplement training data for next-generation LLMs.

Rather than training a single generalist giant model, we envision ``societies of DSS models'': dynamic ecosystems where orchestration agents route tasks to distinct DSS back-ends. This paradigm shift can help decouple capability from size, enabling intelligence to migrate from energy-intensive data centers to secure, on-device experts. By aligning algorithmic progress with physical constraints, DSS societies can help move generative AI away from an environmental liability to a sustainable force for economic empowerment.
\end{abstract}

%% file: sections/0_Introduction.tex
\section{Introduction}
\label{sec:intro}
Currently, the dominant paradigm in the generative artificial intelligence (AI) industry is a ``top-down'' approach: pursuit of artificial general intelligence (AGI) through massive scaling of generalist large language models (LLMs). The prevailing belief is that by training ever-larger models on 
Internet-scale data, we will eventually converge towards an AI system capable of or surpassing human-level reasoning across all domains. This vision relies heavily on scaling laws, which suggest that increasing model size and data automatically yield proportional gains in intelligence \citep{kaplan2020scaling}. However, this ``bigger-is-better'' trajectory incurs ever-increasing costs. Empirical scaling laws document predictable gains from larger models and datasets but do not instill deeper forms of understanding or reliable reasoning on complex, compositional tasks in LLMs. In addition, the current trajectory faces various headwinds in terms of training and inference energy costs, data quality, and, most critically, depth of reasoning. 
More concretely, three interrelated problems make the current top-down paradigm that the AI industry is following a risky foundation to build future AI systems. 

First, both training and deploying large foundation models demand vast compute and energy resources. Modern deep learning (DL) workflows entail substantial financial and environmental costs, which scale quickly as models grow in size. The need for massive computational resources restricts AI development to a few well-funded entities. These costs create practical, economic, and sustainability constraints \citep{strubell2019energy}. 

Second, whereas LLMs often excel at surface-level pattern matching and retrieval, they struggle with robust, compositional reasoning and practical problem solving. Techniques like chain-of-thought (CoT) prompting can unlock stronger reasoning capabilities in language models, but largely rely on scale and special prompting rather than on principled, abstraction-driven reasoning mechanisms. Recent studies show suboptimal LLM performance on tasks that require combining small components into larger, novel solutions \citep{dedhia2025bottom,kansal2026knowledge, zhao2024exploring}. These studies suggest that mere scaling is not likely to produce the kinds of structured reasoning needed for many real-world domains \citep{wei2022cot}. 

Third, data quality matters. The Internet-scale text corpora used to train generalist models are large but highly unstructured and heterogeneous in quality and coverage. Good reasoning and trustworthy behavior depend on high-quality well-curated training data and on targeted curricula that teach models how to use abstractions. The trajectory of intelligent reasoning, one can argue, is predicated on a specific sequence: abstraction followed by generalization. In this view, the first step is to form structured mental models (abstractions) of the world, understand rules, relationships, and causalities, and then generalize these models to new, unseen situations. This data-centric view of AI argues that improving data quality, curation, and structure often yields larger gains than continually increasing model size alone \citep{jakubik2024data, halevy2009unreasonable}. Current LLM training approaches either invert or bypass this step altogether. They attempt to achieve generalization purely through the statistical breadth of knowledge acquisition. This results in the well-documented brittleness of LLMs \citep{haller2025llm, su2025single, mohsin2025fundamental}. 

We propose an alternative trajectory for generative AI: a ``bottom-up'' approach centered on domain-specific superintelligence (DSS). By superintelligence, we do mean super-human, but qualified by its adjective: domain-specific. The AI industry defines superintelligence as being post-AGI since it is pursuing generalist superintelligence across all domains, rather than in specific domains. However, omniscience in a single LLM monolith may be an overrated attribute. We base our proposed trajectory on DSS because it is practically achievable and also because DSS societies can enable significant success in applications that require inter-domain expertise. Thus, rather than relying on a single generalist LLM monolith,  we envision a future built on specialized small language models (SLMs): smaller, focused models trained on high-quality domain data, paired with explicit abstractions [e.g., knowledge graphs (KGs), formal semantics, inductive program libraries] and modular reasoning engines. This vision of a family of specialist models is highly scalable and mirrors how human societies are organized: Individuals acquire expertise in narrow domains and cooperate to solve complex problems that require inter-domain expertise. This idea has deep roots in cognitive science:  Marvin Minsky's Society of Mind framed intelligence as the coordinated effort of many smaller agents, and work on distributed cognition highlights how cognitive capacity emerges from interactions among people, artifacts, and institutions \citep{agents_minsky_society, hutchins1995cognition}. In the field of AI, past successes buttress this idea; highly specialized systems trained for narrow, hard scientific problems have outperformed generalist approaches within their domain, suggesting that specialized architectures and curated data can achieve superhuman depth without extreme scale \citep{agents_alphageometry, jumper2021highly, lin2025goedel}. 

To fully realize this alternative trajectory, we must concretize the operational mechanics of the DSS society. In a traditional monolithic architecture, a single neural network is burdened with encoding the heterogeneous knowledge of the entire world, leading to hallucinations and diluted reasoning. The DSS society, by contrast, treats intelligence as a dynamic network of specialized compute nodes. Each DSS node is an SLM heavily grounded in its 
domain-specific abstraction: whether a medical KG, a legal ontology, or a mathematical formal language like Lean. 
Operationally, this society functions through an orchestration framework that mimics a team of collaborating human experts. A lightweight, general-purpose SLM serves as the central front-end orchestrator. Rather than attempting to generate the final answer from its own internal weights, this orchestrator acts as a cognitive router. When presented with a complex query, such as evaluating a novel pharmaceutical, it decomposes the prompt and dispatches sub-tasks to relevant experts (e.g., a Legal DSS and a Biochemistry DSS). These backend models
collaborate and reconcile their findings to produce a rigorously fact-checked
output.

This ``Lego-style'' composability fundamentally alters the economics and scalability of AI. In the top-down paradigm, achieving marginal reasoning gains requires exponentially scaling the entire generalist model, leading to unsustainable multi-GigaWatt training compute costs. A DSS society decouples capability from scale. Organizations can dynamically assemble customized systems by swapping in updated DSS models without retraining the entire architecture. Crucially, because only the relevant specialized SLMs are activated for any given query, the inference energy footprint is sparsely distributed and orders of magnitude smaller. This extreme efficiency frees AI from centralized cloud data centers, making it feasible to deploy and run intelligent societies locally on edge devices and smartphones.

This trajectory presents a profound philosophical shift. By continuously expanding the library of DSS models and refining the orchestrator, we can asymptotically approach the breadth of AGI. However, this renders the monolithic conception of AGI redundant. If a society of highly efficient, verifiable expert models can collectively outperform massive generalist models in every domain, a single LLM monolith
is no longer the necessary endpoint of AI research. 

Our bottom-up program to build a DSS combines several elements highlighted above. First, careful data engineering and grounded synthetic curricula emphasize quality over quantity. 
Second, explicit abstractions, such as symbolic representations, program libraries, and KGs, provide training ingredients that help models learn compositional rules that deepen their reasoning. Third, modular societies of small specialist models can enable interoperation: a front-end SLM trained more broadly in a domain can divide a user query into sub-queries, dispatch them to specialist DSS models trained more narrowly in that domain, and integrate their responses \citep{ellis2021dreamcoder}. This lego-like design permits efficient training and easier validation, and promises a scalable, efficient, and more interpretable alternative to monolithic generalists. 

This article outlines how this alternative trajectory can not only address the reasoning gap between specialist and generalist models by placing abstraction back at the center of learning but also does so using vastly smaller training and inference energy footprints. By moving inference to the edge (e.g., running efficient SLMs on smartphones), this trajectory can democratize access to DSS, enhancing the productivity of every worker, from technicians to physicians. 

In this article, we address the following:
\begin{enumerate}
    \item We discuss the limitations of the current scale-driven LLM training paradigm in terms of data quality, reasoning depth, and sustainability, citing empirical and theoretical evidence.

    \item We articulate concrete alternative architectures centered on DSS: how to construct them, how to generate targeted synthetic curricula from abstractions, and how to assemble DSS societies for complex tasks.

    \item We propose evaluation protocols and research directions, including continual learning for agents, energy and deployment analyses, and applications in medicine, engineering, and education that demonstrate how our proposed trajectory is testable and actionable.
\end{enumerate}

The article is organized as follows. We first discuss the principles of good data (Section \ref{sec:good_data_importance}), abstractions (Section \ref{sec:abstraction}), reasoning (Section \ref{sec:reasoning}), and modularity needed for DSS (Sections \ref{sec:dss_building_blocks} and \ref{sec:dss}). Other sections show how synthetic data and neurosymbolic components can bootstrap the AI vision (Section \ref{sec:ns_abstractions}) and how agent architectures enable continual learning and real-world use (Section \ref{sec:agents}). Then, we document the empirical limits of scaling and energy footprint of large models, and discuss broader implications for deployment, sustainability, and social benefits of making expert-level AI available across many domains (Sections \ref{sec:energy} and \ref{sec:final_goals}). Finally, we discuss our vision in the context of the AGI quest  (Section \ref{sec:AGI}). Altogether, these sections build a roadmap for an alternative trajectory for generative AI, one that aims for depth over breadth, quality over quantity, and modular, sustainable intelligence over ever-larger generalist monoliths.

%% file: sections/2_Good_data.tex
\section{Importance of good data}
\label{sec:good_data_importance}
If a good abstraction is the engine of deep reasoning, data provide the fuel. However, the current industrial trajectory has prioritized the volume of this fuel over its purity. The dominant paradigm for training LLMs -- scraping massive quantities of text from the open web and training a single generalist giant on that mixture -- seems to have clear pragmatic advantages (availability, scale, no extra effort required) but is a very poor proxy for a high-quality data training signal. 
Studies on massive web corpora, e.g., CommonCrawl, highlight significant issues related to formatting inconsistencies, irrelevant content, and hidden biases that are inevitably internalized by models \citep{perelkiewicz2024review}. 
Even careful filtering pipelines cannot perfectly remove low-quality or adversarial content, and low rates of problematic examples can have outsized downstream effects on model behavior, e.g., on tasks that require deep, compositional reasoning \citep{havrilla2024understanding}.

In contrast, a bottom-up trajectory prioritizes high-quality, domain-specific data. 
It is far more tractable to gather and curate a smaller 
dataset from authoritative sources, e.g., university textbooks, peer-reviewed research articles, and high-value internal enterprise reports that remain behind firewalls. Although smaller in volume, they offer a far richer signal per token. In many applied domains, such as medicine, law, engineering design, and scientific discovery, organizations already possess high-value internal documents that are never published on the open web for intellectual property (IP) or privacy reasons. Collecting, cleaning, and annotating these sources produces training data that directly target reasoning, conventions, and edge cases that matter the most in the given domain. Empirical work across several modalities shows that models trained on carefully curated, high-quality, ``textbook-quality'' data can outperform those trained on much larger, noisy corpora, especially when the task emphasizes depth, precision, and safety \citep{dong2025scalable, abdin2024phi, dedhia2025bottom}.

However, even though training an SLM (e.g., one containing 3-32B parameters) on very high-quality domain data can produce excellent recall and domain fluency, depth of reasoning, i.e., compositional problem solving, discovery of novel chains of thought, and provable program-like manipulation of concepts often remain elusive. 
This is where abstractions and structured synthetic data come into play. We do not mean blindly generating synthetic data using another LLM, which often leads to model collapse or quickly approaches a performance ceiling. Instead, we propose generating synthetic data derived from ground-truth, verifiable abstractions. Synthetic examples generated from domain abstractions, such as formal ontologies, program libraries, KGs, and proof steps from a theorem prover, can automatically lead to targeted curricula that teach the logic of reasoning and composition to models.

Abstractions come in many complementary forms. 
KGs provide relational facts with provenance that are ideal for generating grounded retrieval examples and multi-hop reasoning traces \citep{dedhia2025bottom}. ``Synthesize-on-Graph'' \citep{ma2025synthesize} demonstrates how navigating a context graph can produce diverse, coherent training samples that teach a model how to connect disparate concepts. Formal systems like Lean and LeanDojo \citep{yang2023leandojo} can generate an unlimited number of correct-by-design mathematical theorems and proofs. Methods like LogicPro \citep{jiang2025logicpro} leverage the strict input-output relationships of code (e.g., LeetCode problems) to synthesize complex logical reasoning puzzles, thereby bridging the gap between coding logic and natural-language reasoning. When such structures are used to generate or curate training examples, the resulting datasets contain both the surface-level knowledge (breadth) and latent operators that enable composition (depth).  
The resulting model possesses the recall of a textbook and the logical depth of a symbolic reasoning system.

%% file: sections/3_Abstraction.tex
\section{Role of abstraction and compositionality in reasoning}
\label{sec:abstraction}
 
Abstraction is a key mechanism for compositional generalization in humans~\citep{doi:10.1126/science.aab3050}. The ``Language of Thought'' hypothesis posits that human intelligence is inherently compositional: We do not store every data instance separately but learn abstract primitives and rules that can be recombined productively~\citep{fodor1975language}. Compositional generalization, in turn, underwrites \emph{productivity}: the ability to generate and comprehend infinitely many novel expressions from a finite set of primitives~\citep{NeSyCoCo},  thus handling out-of-distribution inputs that recombine familiar concepts in new ways. However, scaling-law analyses suggest that, for monolithic networks, the samples required for reliable generalization tend to grow with an effective input dimensionality of the task~\citep{2022sharma}. Structurally modular architectures, built from sparsely connected modules, can mitigate this problem on modular tasks when network decomposition matches the underlying modular structure of the task~\citep{boopathy2025breaking}. On controlled benchmarks, such an alignment can make a constant number of samples sufficient even for high-dimensional inputs. 

\subsection{Abstraction as the missing component in LLM pipelines}
\label{sub:abstraction_missing}

Abstraction is central to DSS because specialization alone is not sufficient for deep reasoning. A model trained on domain-specific text may acquire terminology, factual recall, and stylistic fluency but does not necessarily learn the reusable conceptual structure of the domain. DSS requires this structure to be made explicit. Abstractions provide the stable primitives and operators that allow a model to compose known concepts into new solutions. Thus, abstraction is the mechanism that turns domain adaptation into domain-specific reasoning. Abstractions are also what make DSS societies possible. 
Shared abstractions provide common interfaces.

The industry track has largely assumed that conceptual abstraction will emerge as a by-product of increasing parameter counts, compute, and data volume~\citep{krakauer2025llmsemergence}. In this paradigm, LLMs do not place explicit emphasis on built-in mechanisms for abstraction but acquire distributed representations implicitly through self-supervised pretraining and, sometimes, supervised fine-tuning (SFT) on structured tasks. Yet, recent work suggests a gap between what these representation learners readily acquire and the structural requirements of true abstraction~\citep{Kumar2023DisentanglingAbstraction}.  
Although transformers can learn compositional structure, the outcome is sensitive to initialization scale and data quality, and depends strongly on the training setup and architectural constraints  (e.g., whether a modular structure is discoverable from data)~\citep{schug2024discovering}. 

These observations motivate a shift from ``abstraction as a by-product'' to ``abstraction as a target.'' Recent results indicate that distributional coverage and diversity can matter more than sheer scale for compositional generalization~\citep{uselis2025does}. 
Because compositional generalization is not guaranteed by scaling alone and is sensitive to training distributional coverage, this abstraction-centric route may also reduce compute and energy demands.

\paragraph{Takeaway} Systematic compositional generalization rarely emerges ``for free'' under standard training and typically requires additional pressures or structure. Even at scale, abundant data do not guarantee the emergence of appropriate abstractions when the training distribution lacks coverage of the underlying combinatorial factors needed for out-of-distribution recombination. Empirically, compositional performance improves when training is explicitly aligned with the compositional primitives of the task and encourages reusable abstractions, e.g., via primitive-focused pretraining or architectures that exploit task modularity~\citep{ito2022compositional, boopathy2025breaking}.

\subsection{Domain-specific abstractions}
\label{subsec:domain-specific_abstractions}

Across modern AI, it is useful to distinguish domain-specific abstractions from more general ones. 
Domain-specific abstractions, like formal languages paired with solvers or verifiers, turn reasoning into search over checkable intermediate states. Systems such as AlphaGeometry 
\citep{deepmind2024_alphageometry} and AlphaGeometry2 \citep{chervonyi2025gold} illustrate this in geometry: A geometry-specific formalism plus a symbolic deduction engine, guided by a learned model, can solve Olympiad-level problems with far less dependence on human-written proofs. Similarly, formal theorem proving benefits from a proof-assistant-specific environment and premise-selection scaffolds~\citep{hsiang2025leandojov}, where the abstraction is the proof language together with the proof state and checker.

Next, we list some representative abstractions across various domains.

\paragraph{Computer Science}

\begin{itemize}
\item Programming languages and data structures provide executable, compositional interfaces for computation~\citep{turing_lecture_abstractions}.
\end{itemize}

\paragraph{Abstractions that enable checkable reasoning}
\begin{itemize}
  \item Formal logics: propositional logic and first-order logic as languages with inference rules; probabilistic extensions such as Markov Logic Networks~\citep{richardson2006markov}.
  \item Constraint/declarative formalisms: answer set programming~\citep{gel88}.
  \item Proof languages + checkers: abstraction being the proof language and proof state (instantiated by systems such as Lean, Coq~\citep{paulinmohring1993coq}, and Isabelle~\citep{paulson1994isabelle}).
  \item Pattern languages: regular expressions as a domain-specific language for string structure~\citep{fowler2010dsl}.
\end{itemize}

\paragraph{Structured relational abstractions}
\begin{itemize}
  \item Graphs: KGs~\citep{ABUSALIH2021103076}, causal directed acyclic graphs~\citep{vanderweele2010signed}, ontologies with rules~\citep{ARMARY2025100693}, and task-specific context graphs~\citep{xu2024contextgraph}.
\end{itemize}

\paragraph{Scientific priors and invariances}
\begin{itemize}
  \item Physical laws: Equations and constraints that encode known relationships in AI for science.
  \item Symmetries: Symmetry groups as inductive biases, e.g., invariance/equivariance constraints over admissible transformations.
  \item Physics-inspired heuristics: symbolic regression for discovering governing laws~\citep{2020_ai_feynmann}.
\end{itemize}

%% file: sections/4_Reasoning.tex
\section{Reasoning in AI: Beyond pattern matching}
\label{sec:reasoning}
In the DSS trajectory, reasoning is the point at which high-quality data, explicit abstractions, and neurosymbolic scaffolds become operational. This section, therefore, treats reasoning as a stack of mechanisms: adaptation to domain data, grounding in external or symbolic knowledge, constraint enforcement, and verification. This view makes clear why scale alone is an incomplete solution. 
In open-domain settings, a model can often mask weak reasoning with broad recall and plausible language. In a DSS setting, however, the model is judged by whether it can operate under the rules, evidence standards, and verification practices of a particular field. 

\subsection{What is reasoning in AI?}
\label{subsec:what_is_reasoning}

Reasoning derives new knowledge from existing knowledge by applying a set of transformations and inference operations~\citep{pfister2025}.  It is central to intelligence, enabling step-by-step problem-solving beyond rote pattern matching, and adapting to novel situations~\citep{XU2025101370}. In AI, reasoning is tightly coupled with a representation and inherits its semantics. Concretely, AI reasoning applies logical or learned transformations or inference rules to internal representations to produce new knowledge, actions, or plans.

Recent work organizes AI reasoning into categories that mirror psychological taxonomies, including three inference methods --- deduction, induction, and abduction --- and causal, physical (``naive physics''), and associational reasoning~\citep{goertzel_agi_concept}. Some AI systems support deduction by deriving consequences from a set of premises, e.g., inference in formal languages. 
Machine learning 
targets induction, learning general regularities from examples in order to generalize to new samples from related distributions. Abduction, which infers likely causes from incomplete observations, remains particularly challenging because it requires selecting the most plausible explanation under uncertainty, often by reasoning over latent hypotheses. A prominent recent direction moves beyond standalone LLMs toward LLM-based agents and world-model-based planning, which may enable more hypothesis-driven inference and improve abductive reasoning.

The concept of reasoning depth acknowledges that many real-world problems we delegate to AI are compositional. Reasoning depth is most commonly operationalized as the number of dependent inference steps required to reach an answer. A parallel line of work formalizes depth via computation graphs, defining it as the length of the longest path~\citep{dziri2023faith}. Empirically, performance often degrades with depth: As the required number of steps increases, accuracy typically drops sharply~\citep{patel-etal-2024-multi}, a pattern frequently attributed to error accumulation across steps.

\subsection{State-of-the-art reasoning methods}
\label{subsec:sota_reasoning}
Unlike symbolic systems, LLMs learn implicit patterns from vast text corpora. They excel at recall and producing fluent text but their reasoning is non-deterministic 
rather than rule-based, thus requiring verification. Techniques to enhance their reasoning capabilities can roughly be classified as inference-time and training-time \& post-training methods.

\subsubsection{Inference-time reasoning methods}
\label{subsub:inference-time_reasoning}

Such methods improve reasoning without changing model parameters. Instead, they shape how the model is prompted and generates outputs, and what external information it can access at test time.

\paragraph{Prompting and decoding strategies} Techniques like CoT prompting~\citep{wei2022cot} reveal the knowledge the model has obtained during pretraining. They encourage the model to produce intermediate steps before giving an answer. 
However, CoT can underperform on tasks that require compositional generalization from demonstrations to harder instances 
~\citep{zhou2023leasttomostprompting}.
Follow-up work 
introduces explicit branching and selection over intermediate states, as in tree-of-thought (ToT)~\citep{yao2023tot} and graph-of-thought (GoT)~\citep{Besta_2024}. 
These approaches externalize intermediate reasoning states and support multi-step deliberation, with ToT/GoT enabling structured exploration over partial solutions rather than a single-shot response.
Beyond pure decomposition, ReAct~\citep{yao2023react} interleaves reasoning traces with explicit actions and observations. 
Related approaches perform verifier-guided search by sampling multiple candidate reasoning paths and scoring them with external evaluators. Process reward models 
assess intermediate steps rather than only the final answer \citep{lightman2023let}. 

\paragraph{Retrieval and memory augmentation} Retrieval-augmented generation (RAG) supplies the model with relevant external context at inference time.  
More generally, memory-augmented neural networks and graph neural networks can inject structured signals, helping bridge unstructured text patterns and more explicit forms of reasoning~\citep{patil2025advancingreasoningllms}. Work that integrates LLMs with KGs often frames the model as an agent that navigates a graph, linking reasoning quality to grounding in an explicit relational representation~\citep{sun2024thinkongraph}.

\paragraph{Adaptive compute allocation}
State-of-the-art systems vary inference-time compute based on problem difficulty. 
They avoid overthinking on simple prompts while enabling deeper deliberation on competition-level mathematics or complex coding problems. With sufficient test-time compute, a smaller model can outperform one that is 14$\times$ larger~\citep{snell2025scaling}.

\subsubsection{Training-time and post-training reasoning methods}
\label{subsub:training-time_reasoning}

These methods improve multi-step reasoning by updating model parameters with gradients and, in many cases, by training auxiliary components such as reward models.
A common starting point is reasoning-trace supervision, in which the model is fine-tuned on reasoning-rich data that include step-by-step solutions rather than only final answers~\citep{ouyang2022training, zhang2026instruction}. This can establish a structural grammar for multi-step outputs but also risks surface imitation in which the model reproduces the teacher's style without acquiring the underlying problem-solving procedure. 
Reasoning distillation transfers high-quality traces from a stronger teacher to a smaller student~\citep{xu2024survey}. Effective distillation 
relies on careful data selection so that the student internalizes the implicit problem-solving logic rather than memorize output tokens.

Beyond supervised traces, many approaches rely on reinforcement learning (RL). Reinforcement learning from human feedback (RLHF)~\citep{ouyang2022training} can further improve performance by optimizing toward preferred outputs. 
To reduce the overhead of explicit reward modeling, direct preference optimization~\citep{2023rafailov_dpo} formulates the objective directly in terms of the policy, enabling end-to-end optimization without a separate reward-model training stage.

A more recent direction emphasizes 
verifiable rewards, e.g., in 
group relative policy optimization (GRPO)~\citep{shao2024deepseekmath}. In contrast to proximal policy optimization~\citep{schulman2017proximal}, which uses a memory-intensive critic, GRPO estimates the baseline from group-level statistics over multiple sampled outputs, reducing memory overhead. When paired with process-based reward models that evaluate intermediate steps, this setup supports scalable training focused on step-wise correctness.

Finally, tool-use training expands the capabilities of the model by teaching it to invoke external functions or application programming interfaces (APIs)~\citep{agents_toolformer}. This enables models to overcome inherent limitations in arithmetic, factual retrieval, and real-time awareness.

\subsection{Domain-specific reasoning}
\label{subsec_dss_reasoning}

\begin{figure}[t]
  \centering
  \includegraphics[width=\linewidth]{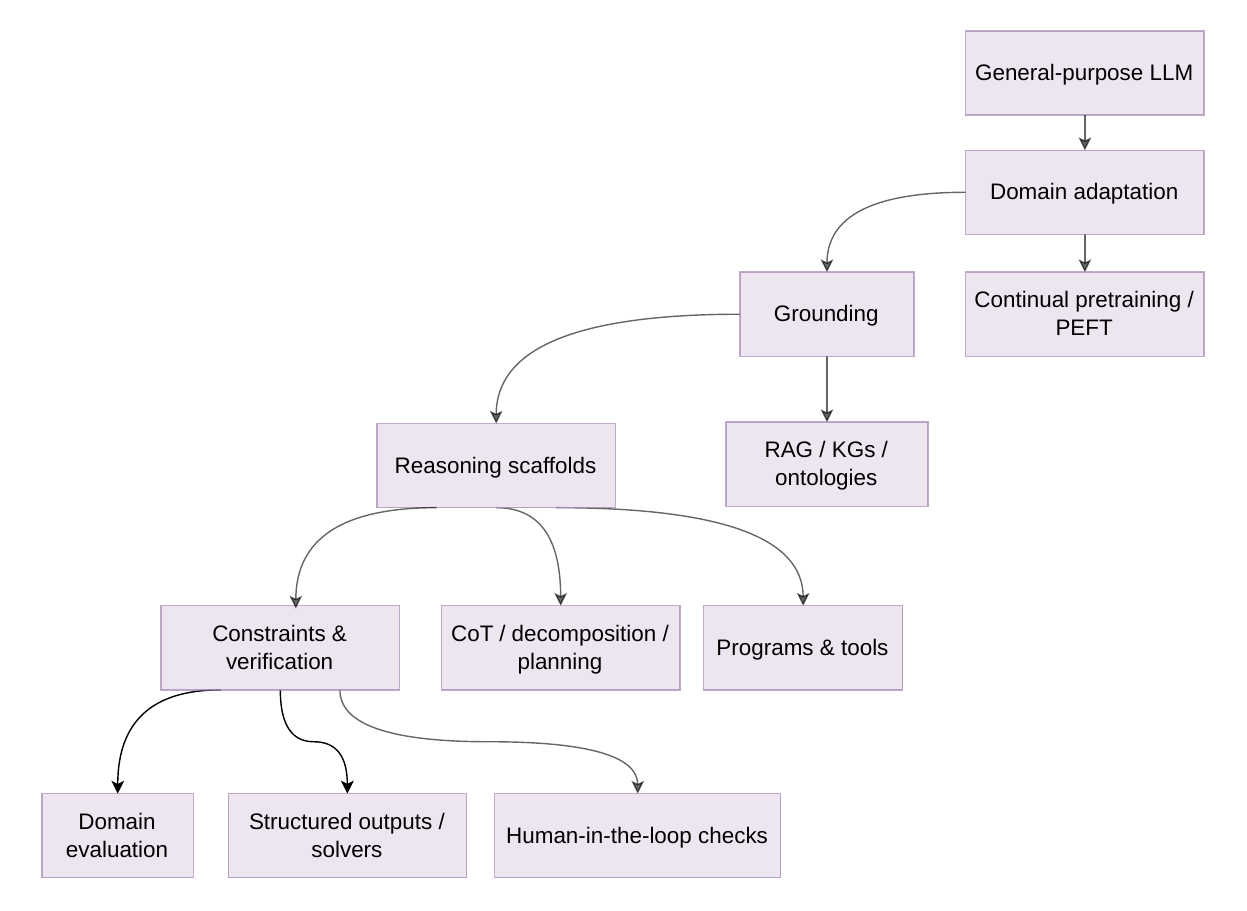}
  \caption{A conceptual workflow for domain-specialized reasoning with LLMs. Stages may be combined or iterated.}
  \Description{A flowchart begins with a general-purpose LLM
  and proceeds to domain adaptation, which branches into
  continual pretraining or PEFT
  and grounding. Grounding connects to RAG, KGs, and ontologies, and leads
  to reasoning scaffolds. These scaffolds branch into
  constraints and verification; chain-of-thought reasoning,
  decomposition, and planning; and programs and tools.
  Constraints and verification further connect to domain
  evaluation, structured outputs and solvers, and
  human-in-the-loop checks.}
  \label{fig:ds_reasoning}
\end{figure}

Domain settings reshape what it means for an LLM to ``reason.'' Compared to open-domain use, a model must operate under domain-specific objectives and constraints, including heterogeneous inputs, proprietary knowledge, and evolving terminology~\citep{Ling2025_doaminspec}. 
As a result, \emph{domain-specific reasoning is increasingly framed as a specialization problem under domain constraints that trades breadth for domain-grounded reliability}.

In practice, the most visible approaches cluster into a small set of strategies (Fig.~\ref{fig:ds_reasoning}). One line of work uses domain-adaptive continual pretraining to absorb domain language and conventions. Another emphasizes parameter-efficient adaptation, including prompt tuning and the low-rank adaptation family of methods, to reduce compute and data requirements. A third line focuses on grounding, typically through retrieval or structured knowledge, such as KGs. Finally, many systems scaffold reasoning with tools, programs, and constrained intermediate representations, aiming to make intermediate steps more controllable and easier to verify.

In high-stakes domains, notably medicine,
evaluation has increasingly moved beyond benchmark accuracy toward expert-aligned criteria, including factuality, comprehension, reasoning quality, harm risk, and bias, often assessed through workflow-grounded studies, which indicate that benchmark success is insufficient as a proxy for real-domain performance. In medicine, domain-adapted models, e.g., Med-PaLM~2~\citep{singhal2025toward} and PMC-LLaMA~\citep{pmc_llama}, have demonstrated that continued pretraining and fine-tuning on medical corpora can mitigate limitations of general models. 
Given the paramount importance of clinical safety and ethical implications, the current trend in medical reasoning is toward explainable AI frameworks that emphasize transparency, responsibility, and accountability in decision-making. 

A further practical constraint is cost. Domain-specific development increasingly favors cost-effective adaptation rather than training from scratch~\citep{xie-etal-2024-efficient}. Since domain adaptation is constrained by hardware and memory~\citep{Wang2025PEFTSurvey}, parameter-efficient fine-tuning (PEFT) methods have become paramount in
training cost while retaining performance.

\subsubsection{Verification}
\label{subsubsec:verification}
Verification supports domain reasoning by checking outputs against formal constraints and external evidence. 
It transforms LLMs from purely probabilistic generators into more reliable decision-making agents. 
The imperative for stepwise verification arises in domains where the reasoning path matters as much as the conclusion. A common failure mode manifests as an apparently correct conclusion supported by flawed intermediate logic, which suggests reliance on statistical shortcuts within the training distribution~\citep{song2025a_survey}. Recent work also highlights a tension between optimizing for instruction-following and preserving factual reliability: Preference-based tuning can increase sycophancy and can yield unfaithful rationales, although some RLHF regimes improve truthfulness on average~\citep{ICLR2024_0105f797}.

Strongest verification is available in domains with formal rules and objectively checkable outcomes. Mathematical problem solving and code generation are the primary testbeds. One direction in reliability is autoformalization: using LLMs to translate informal statements into formal languages that can be checked by a solver. 

\subsubsection{The DeepMind Alpha series as DSS}
\label{subsubsec:alpha_dss}

Specialization can yield striking performance within a narrow scope. The DeepMind Alpha lineage exemplifies this pattern: Scale is used to search and learn deeply within a single domain, rather than many domains. Such a focused capability aligns with our definition of DSS.
Concretely, this design pattern typically involves three engineering solutions:
\begin{enumerate}
    \item Embedding rich domain structure into an environment.
    \item Optimizing a well-defined objective under dense, programmatic feedback signals.
    \item Evaluating against domain-native oracles when available (e.g., compilers, formal proof checkers) and against reference standards when not (e.g., experimental protein structures and benchmarks).
\end{enumerate}
These environments make correctness decidable or measurable with high precision. This enables the domain itself to supervise behavior at scale through mechanisms of self-play, synthetic data generation, or formal verification.

AlphaGo~\citep{silver2016mastering} combines supervised learning from human games with self-play and tree search. 
AlphaGo Zero~\citep{silver2017mastering} reduces reliance on human demonstration data by exploiting self-play under known rules. AlphaZero~\citep{doi:10.1126/science.aar6404} reaches superhuman play in Go, chess, and shogi by learning from game rules alone. Outside games, AlphaFold~\citep{jumper2021highly} makes highly accurate protein structure prediction.
AlphaFold~3~\citep{abramson2024alphafold3} extends the target from single proteins toward biomolecular complexes.
AlphaStar~\citep{vinyals2019alphastar} achieves expert-level performance for StarCraft. AlphaGeometry~\citep{deepmind2024_alphageometry} addresses Olympiad geometry without human demonstrations. It synthesizes large amounts of training data and combines learned components with symbolic reasoning to produce rigorous proofs. 
AlphaProof~\citep{Hubert2025OlympiadFormalMathRL} extends this paradigm to formal mathematics by learning to construct machine-checkable proofs in Lean via RL over large sets of auto-formalized problems. Together with AlphaGeometry2~\citep{chervonyi2025gold}, AlphaProof has reached silver-medal performance on International Mathematical Olympiad problems. 
The strongest results often come from generating high-quality data through interactions, e.g., self-play and synthetic theorem generation, rather than 
large, curated, labeled sets. 

\paragraph{Limitations} 
Alpha-style success largely depends on whether we can formalize a domain into an environment interface with states and actions, clearly specify a prediction target, and check the evaluation signal. Board games and formal proof assistants satisfy these requirements exceptionally well, whereas many open-world domains do not. 
Nonetheless, the Alpha series illustrates what domain-specific structure can enable, reaching performance that can reasonably be described as superintelligent within the domain.

\paragraph{Takeaway}
Gains in reasoning depend on better adaptation schemes, stronger inference-time control, and tighter feedback loops between training and inference~\citep{XU2025101370}. Rich, task-relevant context can matter as much as parameter count: Retrieval augmentation can enable a smaller language model to outperform a larger one on knowledge-intensive queries by grounding answers in an external knowledge base~\citep{2023atlas}. Train-time improvements based on SFT face a bottleneck: Collecting large-scale, high-quality human annotations for instruction-tuning datasets is expensive. This is why rapid progress has concentrated in domains with automatic verification. 

The field is also shifting from scaling 
during training to scaling at inference. 
Simply enlarging the pretraining corpus is no longer sufficient to unlock the next jump in capability. Instead, researchers increasingly allocate more inference-time compute to support deeper, more deliberate reasoning. 
Because multi-step reasoning typically requires multiple forward passes, improving step-level reliability is often more cost-effective than increasing model size.

%% file: sections/6_Neurosymbolic_AI.tex
\section{Unleashing the potential of neurosymbolic AI through abstraction}
\label{sec:ns_abstractions}

Neurosymbolic AI 
reconciles two historically divergent traditions in AI: connectionist learning and symbolic reasoning. Neural networks demonstrate remarkable capacity for modeling high-dimensional, unstructured data. 
However, they often lack explicit structural constraints required for systematic reasoning and formal generalization. In contrast, symbolic AI emphasizes logic, compositionality, and knowledge representation, offering interpretability and deductive rigor, yet often exhibiting brittleness and limited adaptability to noisy, real-world inputs~\citep{garcez2023neurosymbolic}.

The motivation for this integration is grounded in a broader theory of cognition: Robust intelligence requires both the capacity to learn from experience and the ability to reason from what has been learned~\citep{bader511042dimensions, d2009neural, donadello2017logic}. 
Neural models provide scalable statistical learning mechanisms whereas symbolic systems contribute compositional abstraction and logically structured inference. A neurosymbolic architecture, therefore, seeks to synthesize them into unified computational systems in which representation learning and rule-based reasoning mutually reinforce one another 
~\citep{liang2025ai}.

We argue that incorporating a high-level abstraction as the ``symbolic'' part into neurosymbolic systems offers a particularly promising approach to advancing language model capabilities. By unifying data-driven learning with structured reasoning, neurosymbolic systems can realize both flexibility and reliability. In the following sections, we examine how this paradigm is reshaping distinct yet critical domains: KGs and formal languages.

\subsection{Knowledge graphs as semantic abstractions}
Real-world knowledge can be represented in a KG as triples of the form (head entity, relation, tail entity). This encodes how two entities are linked through a specific relationship. In the LLM field, KGs have emerged as an important tool for representing, storing, and effectively managing knowledge~\citep{graph_survey, ns_methods_kg_reasoning, belova2025graphmert}. 

In reasoning-intensive settings, KGs provide a structural counterweight to hallucinations and factual mistakes observed in purely neural architectures~\citep{yu2024cosmo}. Recent frameworks, such as GraphRAG~\citep{dss_graphrag_pharma}, extend RAG by automatically constructing KGs from large text corpora and then extracting subgraphs from them to respond to queries. Generation is conditioned not only on retrieved passages but also on an explicit graph-structured representation, thereby enhancing accuracy. Building on RAG, ToG-2~\citep{ma2024think} tightly couples the processes of context retrieval and graph retrieval. 
Graph-R1~\citep{luo2025graph} further advances the method by introducing lightweight knowledge hypergraph construction and formulating retrieval as a multi-turn agent-environment interaction, optimized through end-to-end reward signals. Graph-constrained-reasoning~\citep{luo2024graph} further bridges the structured knowledge embedded in KGs with the unstructured reasoning of LLMs by integrating the KG structure into the LLM decoding process through a trie-based index that encodes KG reasoning paths. 
KG primitives can also be used to synthesize curricula for domain-specific distillation and finetuning~\citep{dedhia2025bottom}. Moreover, KGs can function as implicit reward models to learn compositional multi-hop reasoning~\citep{kansal2026knowledge}. These developments illustrate how KGs implement neurosymbolic integration at both the retrieval and inference levels.

Because semantic relationships are represented as a graph, the underlying symbolic knowledge remains inherently interpretable to human observers. This ensures that users can directly audit the knowledge base, identify gaps, and easily expand the graph by introducing new nodes and edges as needed. 

\subsection{Formal languages as a mathematical abstraction}
For formal reasoning in modern LLMs, the field has 
converged upon the Lean theorem prover~\citep{de2015lean, moura2021lean} as an 
abstraction layer. Lean is an open-source, extensible, functional programming language and interactive theorem prover 
designed to make the writing of correct, machine-checkable code and mathematical proofs efficient and maintainable
~\citep{lin2025goedel}. This 
shifts the burden of verification away from the probabilistic neural network and 
onto the symbolic compiler~\citep{yang2026formal_rlm_llms}.

The broader ambition of AI mathematical reasoning requires direct engagement with universal formal languages. AlphaProof~\citep{alphaproof} fundamentally reconceptualizes the Lean language as a dynamic environment for RL. Problems are automatically formalized from natural English into Lean, enabling the neural network to generate proof steps whereas the compiler serves as a perfect, hallucination-free reward mechanism. 
To handle complex proofs with convoluted deductive chains, DeepSeek-Prover-V2~\citep{ren2025deepseek} uses recursive proof search via subgoal decomposition. It uses DeepSeek-V3~\citep{liu2024deepseek} to break a theorem into formal subgoals, then deploys a highly optimized 7B prover to solve smaller lemmas before assembling them into a complete formal proof. 
Goedel-Prover-V2~\citep{lin2025goedel} 
effectively creates its own curriculum by generating new subproblems from failed proof attempts. It also applies verifier-guided self-correction, using detailed feedback from the Lean compiler to iteratively revise and debug proofs within a single context window, achieving state-of-the-art results with significantly fewer parameters.

Whereas general formal languages like Lean offer universal applicability, specific geometric domains present unique challenges that require highly specialized architectures. Google DeepMind introduced AlphaGeometry~\citep{deepmind2024_alphageometry}, a system that achieves Olympiad-level performance by explicitly decoupling neural intuition from symbolic deduction. The symbolic core is a Deductive Database and Algebraic Reasoning engine that iteratively applies classical geometric deduction rules structured as definite Horn clauses, taking the logical form $Q(x)\leftarrow P_1(x),P_2(x),\dots,P_k(x)$, where $x$ represents geometric point objects and the predicates represent relationships like colinearity.  
To overcome the bottleneck of scarce human training data, researchers developed AlphaGeometry2~\citep{chervonyi2025gold}, using the symbolic engine to generate a vast pool of synthetic training data comprising 300 million unique, machine-verified geometric examples. 

In summary, formal language can function as a powerful abstraction that enables reliable mathematical reasoning within neurosymbolic systems, rather than relying solely on probabilistic approximations. 

\subsection{Neurosymbolic pathways to small models}
Neurosymbolic approaches significantly enhance parameter efficiency by distilling knowledge from LLMs into highly capable, smaller architectures. E.g., extracting common sense knowledge from GPT-3 into a combined KG and neural model creates a 100$\times$ smaller system that still achieves superior reasoning performance~\citep{west2022symbolic}. Similarly, the Ctrl-G framework explicitly integrates symbolic logic constraints into a distillation pipeline to generate a smaller model~\citep{zhang2024adaptable}. Neurosymbolic AI can also reduce the amount of data and parameters needed by natively embedding symbolic knowledge to fill gaps during both training and inference. Co-designing neural networks to estimate physical variables alongside symbolic equations that govern physical conservation laws for off-road autonomous driving can achieve state-of-the-art performance with significantly reduced training time and parameter count ~\citep{zhao2024physord}. Finally, neurosymbolic frameworks efficiently leverage human subject matter expertise by seamlessly integrating a small number of curated examples directly into symbolic components, such as KGs. This 
successfully combines fast adaptation to novel situations, characteristic of symbolic reasoning, with the high-dimensional generalization capabilities of machine learning~\citep{bhagat2023sample}.
 
This framing turns specialization into reliable DSS. KGs, ontologies, formal languages, simulators, and rule systems expose the reusable primitives of a field in a form that can be traversed, composed, audited, and verified. They can, therefore, serve as grounding sources at inference time, curriculum generators during SFT, and reward or verification mechanisms during post-training. In this sense, neurosymbolic AI supplies the technical bridge between high-quality domain data and DSS: It converts domain knowledge from unstructured evidence into manipulable abstractions that small models can internalize and reason over.

This paradigm is also central to the proposed DSS society because societies of specialist models require shared interfaces, not just independently trained experts. If each DSS encodes knowledge only implicitly in its parameters, collaboration across models would depend on unconstrained natural-language exchange, making routing, verification, and synthesis fragile. Neurosymbolic abstractions provide a common protocol through which different DSS models can communicate: A medical DSS can expose a KG of diseases, symptoms, and treatments; 
a mathematical DSS can expose formal proof states; and an engineering DSS can expose equations, simulations, or design rules. A front-end orchestrator can then decompose a complex query, route subproblems to the appropriate DSS modules, and integrate their outputs through these explicit symbolic structures. 

%% file: sections/7_DSS_.tex
\section{Building blocks of domain-specific superintelligence}
\label{sec:dss_building_blocks}

Next, we describe how the foundational elements described above -- high-quality data, robust abstractions, advanced reasoning methods, and neurosymbolic architectures -- can be practically orchestrated to serve as DSS building blocks. 

\subsection{KG extraction methods}
\label{subsec:dss_builing_blocks}
As the demand for structured knowledge representation has grown, various methods have emerged to automatically extract and construct KGs from unstructured data. These approaches can be categorized into three main groups: task-specific natural language processing (NLP) pipelines, triple embedding-based techniques, and LLM-based extraction.

Traditional 
NLP methods sequentially chain machine learning components, such as named entity recognition, coreference resolution, and relation extraction~\citep{Jaradeh2023}. 
However, they require extensive feature engineering, sophisticated text-preprocessing heuristics, and deep domain expertise~\citep{brissette2024LLMKG}. Furthermore, supervised models within these pipelines require massive, domain-specific labeled datasets, making the extraction process labor-intensive and difficult to scale. Alternatively, triple embedding-based approaches map KG structures into vector spaces to predict missing links and infer new relations~\citep{ns_methods_kg_reasoning}. However, since these methods primarily operate on localized triple patterns, they struggle with multi-hop dependencies, ontological constraints, and external semantic integration~\citep{reasoning_over_kg_with_logic_2025}. They are also limited to the relations, attributes, and ontology of the training KG, making it hard to generalize across different KGs.

Recent KG extraction research has shifted towards using LLMs to generate relational knowledge through targeted prompting~\citep{LLM_for_kg_construction, LI2025104769}. Although LLMs offer impressive flexibility and ease of deployment, their use in KG extraction remains unreliable for rigorous, domain-specific applications. A central limitation is their sensitivity to prompt formulation. Relation extraction accuracy and output consistency can vary substantially in response to minor syntactic changes or small shifts in task framing~\citep{cao-2021-knowledgeable, mousavi-etal-2024-dyknow}. 
Beyond prompt sensitivity, LLM-based extraction is further undermined by intrinsic hallucinations and factual inaccuracies. 
They often produce ungrounded or unfaithful content, which persists regardless of model scale, dataset size, or fine-tuning strategy~\citep{kim2025medicalhallucinations}. In the context of KG construction, LLMs often exhibit the ``reversal curse''~\citep{2024reversalcurse}, failing to perform simple inverse reasoning that is essential for accurately representing bidirectional relations. Moreover, because comprehensive data verification at pretraining scale is infeasible, these models often produce factual errors, particularly for rare or underrepresented entities~\citep{min-factscore}. In high-stakes domains, such as medicine, cybersecurity, and aeronautics, where structural consistency, traceability, and interpretability are essential, this poses a major problem.
Hence, 
we need a reliable KG extraction approach.
\ours
(Graphical Multidirectional Encoder from Transformers)~\citep{belova2025graphmert} is an automatic and scalable framework for extracting domain-specific KGs from text that mitigates the above limitations.

\subsection{\ours: Distillation of factual and ontologically valid domain-specific KGs from raw text}
\label{subsec:graphmert}

\begin{figure}[t]
\centering
\includegraphics[width=\linewidth]{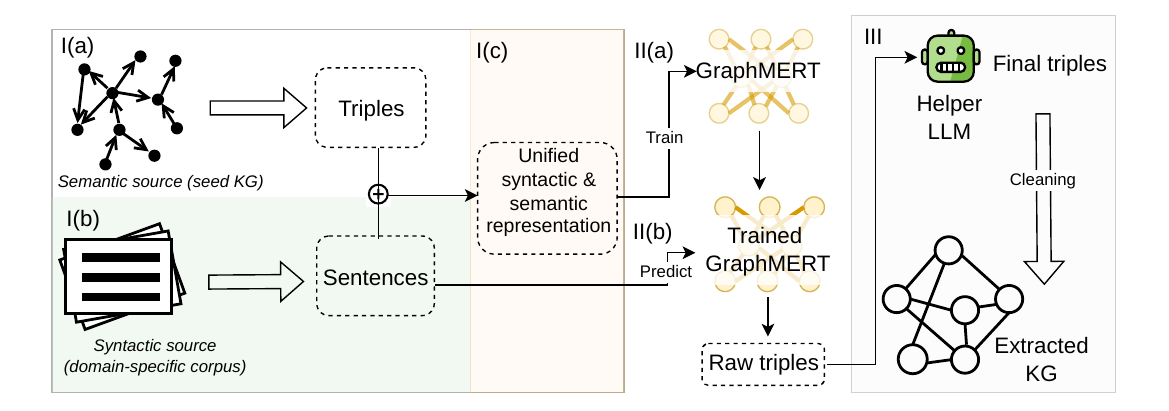}
\caption{Overview of the \ours framework. 
\textbf{(I)}: Unified representation \textbf{(Ic)} combines syntactic knowledge from text \textbf{(Ib)} with semantic examples that include domain-specific relations from a seed KG \textbf{(Ia)}.
\textbf{(II)}: \ours is trained on semantic examples unified with their syntactic context \textbf{(IIa)}. It then predicts novel semantic completions, using their syntactic information as context \textbf{(IIb)}.
An LLM helps refine the linguistic structure of raw triples proposed by \ours \textbf{(III)}. After filtering them by similarity to the source context and dropping duplicate triples, we obtain the final KG
~\citep{belova2025graphmert}.
}
\Description{A left-to-right pipeline with three stages.
  Stage I combines triples from a seed knowledge graph
  with sentences from a domain-specific corpus to form
  a unified syntactic and semantic representation.
  Stage II uses this representation to train GraphMERT.
  The trained model then takes corpus sentences as input
  and predicts raw triples. In Stage III, these triples
  pass through a helper LLM for refinement, followed by
  a cleaning step that produces the extracted knowledge
  graph. Arrows connect the data sources, training,
  prediction, refinement, and cleaning steps.}
\label{fig:graphmert}
\end{figure}

Within the DSS framework, a domain-specific KG functions as a semantic abstraction that augments training data, thereby enabling depth in domain-specific reasoning. 
For a KG to be useful in this role, it must demonstrate high quality along two dimensions: factuality and validity. Factuality captures provenance: whether each triple is supported by, and traceable to, the source text. Validity captures ontological consistency: whether the relation semantics of a triple conform to the domain ontology.
The \ours framework (see Fig.~\ref{fig:graphmert}) is designed to satisfy these requirements while remaining fully automatic. Unlike manual pipelines, it does not rely on hand-crafted features. Instead, it requires only a small seed KG that exemplifies how relations are used under the domain ontology. Given the seed KG and a high-quality domain corpus, it can extract a reliable domain KG, making it a practical first step in the DSS pipeline. 

The \ours model, which is the core of the \ours pipeline, learns graph-structured, ontology-aligned representations in an encoder-only transformer and uses them to predict new triples from local textual context. After training, this  model performs semantic predictions to obtain new KG triples based on context.

During data preparation, the framework augments syntactic corpora with semantic supervision. The semantic supervision consists of KG triples aligned with the domain ontology, providing examples of domain-specific relations. These triples are paired with domain text so that each triple is relevant to its local linguistic context. As a result, at inference time, \ours predicts triples that are consistent with the domain ontology. Each predicted triple can be traced to its origin-supporting provenance. Moreover, the extracted KG can be used to expand the initial seed and iteratively retrain the model, progressively improving coverage and strengthening the domain abstraction over time.

\subsection{\ours as an abstraction-extraction layer}
\label{graphmert_abstraction_extraction}

In the DSS trajectory, \ours addresses the abstraction bottleneck. The central premise is that robust domain-specific reasoning requires explicit abstractions before generalization. Yet, in many domains, such abstractions are incomplete, fragmented, or not available in machine-actionable form. For instance, sciences, engineering design, education, law, finance, manufacturing, and enterprise operations domains contain large amounts of scattered documentation but lack explicit symbolic representations. Thus, without an automatic abstraction-extraction layer, the DSS program would depend on the manual construction of symbolic knowledge, which is slow, expensive, and difficult to scale.

\ours shows how a concrete instantiation of such an abstraction layer can be built. Its input is a high-quality domain corpus together with a small seed KG that specifies the relation semantics of the domain. Its output is an expanded symbolic abstraction in the KG form, whose triples are grounded in the source text and aligned with the ontology expressed by the seed. This changes the role of unstructured text in the DSS pipeline. Instead of treating domain documents merely as data for next-token prediction or as passages for retrieval, \ours converts them into a semantic substrate. This step is crucial because DSS models require more than high-quality text. An SLM trained only on domain documents may acquire fluency and factual recall but is unlikely to internalize the full relational structure of the domain, especially when the task requires multi-hop composition. By contrast, a KG extracted by \ours makes domain primitives explicit. It identifies entities, relations, and local semantic neighborhoods that can be traversed, composed, and checked. These structures can then serve as inputs to synthetic curriculum generation, graph-grounded retrieval, process supervision, and implicit reward modeling. The complete pipeline, therefore, becomes: 
\begin{equation*}
\begin{aligned}
&\text{high-quality corpus } \rightarrow \text{extracted abstraction}
  \rightarrow \text{synthetic curriculum} \\
&\rightarrow \text{SLM training } \rightarrow \text{DSS model.}
\end{aligned}
\end{equation*}

The need for \ours-like methods also follows from the limitations of LLM-based KG generation, which also applies to the extraction of other types of abstractions. 
LLMs often hallucinate relations, reverse relation directions, violate ontology constraints, and generate facts that cannot be traced to source evidence. These failure modes are especially damaging when the extracted KG is not the final output but the foundation on which later training and reasoning stages depend. 
Hence, abstraction extraction requires stricter guarantees than ordinary text generation. \ours constrains KG extraction through semantic relation embeddings learned from a seed KG and by preserving sentence-level provenance for extracted triples. 

\ours currently focuses on KG extraction. However, the underlying idea is broader. Some domains may require richer abstractions, e.g., typed ontologies, causal graphs, event schemas, or constraint systems. However, the \ours approach can be extended to these scenarios as well: A small seed abstraction defines the symbolic vocabulary and admissible operators of the domain; a high-quality corpus supplies factual evidence; a compact neural model learns to align the two; and a symbolic object is extracted as the interface for downstream reasoning.

For ontology extraction, \ours could be modified to extract not only binary relations but also entity types and taxonomic structure. Such an extension would be useful in domains where conceptual hierarchies matter more than individual facts. For attribute-rich KGs, the model could attach qualifiers to triples, e.g., temporal scope, confidence, and modality. This would make the extracted graph more useful for domains where relations are context-dependent rather than universal. For example, in medicine, the efficacy of a drug may depend on what other drugs are being taken.

These extensions would require new training objectives and validators. Triple extraction can be evaluated based on factuality and ontology validity, but event extraction, causal graph extraction, and constraint extraction requires additional checks. Therefore, general abstraction extraction  should not be understood as a single universal model that emits all types of symbolic structures. Rather, it is a methodological template: Pair a compact neural extractor with  seed symbolic schema and a domain-specific validation mechanism.

The benefit of this template is that it makes DSS construction repeatable. Given a new domain, one would not begin by training a large generalist model. Instead, one would collect a small high-quality corpus, define or obtain a seed abstraction, train an extractor, audit the resulting symbolic structure, and only then generate curricula or reward signals.

This perspective also clarifies the limitations of the current \ours framework and turns them into research directions. The current model relies on a seed KG, uses a fixed relation set after training, and still depends on a helper LLM for combining predicted tail tokens. These limitations are not incidental; they identify the technical challenges that must be solved to make abstraction extraction broadly applicable. Future systems should support direct multi-token semantic prediction, richer typed outputs, and stronger domain validators. Solving these problems would move \ours from a reliable KG extraction method toward a general abstraction-extraction engine for DSS.

\subsection{Generating high-quality KG-grounded bottom-up curricula}
Once a reliable abstraction, such as a domain-specific KG, is obtained, the next critical building block is to leverage it as an active training engine rather than a static database. 
By treating a reliable KG as a data foundry, we can systematically traverse its explicit domain primitives (represented as head-relation-tail triples) to synthesize a dense, high-quality reasoning curriculum. This ensures the training data inherently capture the fundamental axioms of the domain.

This paradigm was rigorously demonstrated in \citep{dedhia2025bottom}. It uses a medical KG to generate a vast curriculum designed to impart deep domain expertise. Rather than simply extracting isolated facts, the pipeline synthesizes multi-hop reasoning tasks by connecting disparate medical concepts through logical paths. The result is a comprehensive training set, comprising over $24,000$ reasoning tasks, each paired with a detailed thinking trace that models the step-by-step logic required to reach a conclusion. Fine-tuning a 32B SLM on this explicitly grounded data led to the development of QwQ-Med-3: a model uniquely equipped with internalized medical primitives.

The resulting capabilities of QwQ-Med-3 showcase how this curriculum realizes our core vision of building DSS. Evaluated on ICD-Bench, a rigorous suite featuring 3,675 complex questions across 15 distinct medical domains, the model demonstrates expert-level reasoning that significantly outperforms larger, state-of-the-art generalist models \citep{dedhia2025bottom}. Notably, even though the model is trained on shorter 1-to-3 hop KG paths, it successfully generalizes to unseen, complex 2- to 5-hop scenarios during testing. This shows that by grounding a curriculum in verifiable abstractions, we can efficiently teach SLMs to deeply compose domain concepts.

\subsection{KG-grounded curricula as the data foundry for DSS}
\label{subsec:curricula_data_foundry}

In the broader context of the DSS trajectory, KG-grounded curricula directly address the data quality bottleneck. As argued earlier, the top-down paradigm relies on scraping the open web, which is inherently noisy and increasingly depleted of high-quality human text. While generalist LLMs require trillions of tokens to statistically approximate reasoning, the DSS approach posits that intelligence can be engineered much more efficiently if the training data are explicitly structured. KG-grounded curricula serve as the ``data foundry'' for this vision. By systematically converting verified domain abstractions into multi-hop reasoning traces, we manufacture the exact high-signal fuel necessary to train specialized systems. 

This mechanism is critical for populating the DSS society. A society of experts is only as capable as its individual members. If we are to construct thousands of specialized SLMs, ranging from molecular biology blocks to neuroscience, physics, and engineering blocks, relying on human experts to manually curate distinct training datasets for each domain is economically and temporally impossible. The KG-to-curriculum pipeline automates the generation of expert-level training data. It guarantees that the resulting SLM is not merely memorizing domain trivia but is actively practicing the compositional logic unique to that field. 

Ultimately, this curriculum generation strategy subverts the traditional heterogeneous scaling of unstructured data for training. It demonstrates that the path to superhuman domain capability does not require scaling the model size to hundreds of billions of parameters but rather scaling the \textit{quality and relational depth} of the training data. By grounding the curriculum in a verifiable abstraction, the resulting DSS module achieves the deep, specialized competence required to act as a reliable back-end expert within the larger society of models.

\subsection{Abstractions as implicit reward models}
Whereas fine-tuning on a KG-grounded curriculum successfully grounds the model in domain facts, SFT alone is often insufficient to elicit robust, zero-shot compositional reasoning. Models trained purely on supervised data can still default to superficial pattern matching when faced with highly complex, unseen multi-hop scenarios. To unlock true logical depth, the model must undergo RL that provides rigorous process supervision, evaluating and rewarding the method of reasoning, not just the final answer. However, traditional RL environments are either too rigid to scale easily 
or require careful expert curation or are too niche to generalize.

The solution lies in using the underlying abstractions themselves as implicit reward models. Because a KG encodes the causal and relational links between entities, the multi-hop paths within the KG can serve as a ground-truth, verifiable logical chain. In \cite{kansal2026knowledge}, a scalable SFT+RL post-training pipeline is demonstrated, in which these KG paths provide novel reward signals.
During the RL phase, the model is dynamically evaluated against these path-derived signals. If the model successfully composes the intermediate axioms, navigating the correct sequence before arriving at the conclusion, it receives a high reward. This shifts the optimization landscape, forcing the model to explicitly prioritize coherent logical composition over hallucinated shortcuts.

This approach bridges the gap between simple recall and true compositional reasoning. In experiments within the medical domain, a 14B language model trained exclusively on short 1-to-3 hop reasoning paths can zero-shot generalize to more complex 4- and 5-hop queries
\cite{kansal2026knowledge}. The pipeline enables an SLM to significantly outperform much larger frontier systems, such as GPT-5.2 and Gemini 3 Pro, on the most difficult reasoning tasks, while also demonstrating robust resistance to adversarial option-shuffling stress tests. Thus, using KGs as implicit reward models provides the scalable, deterministic supervision necessary to transform an SLM into a rigorous domain expert.

\subsection{Implicit reward models as verifiable reasoning scaffolds for DSS}
\label{subsec:reward_models_scaffolds}

Just as KG-grounded curricula solve the data supply problem for DSS models, implicit reward models address the reasoning supervision bottleneck. In the conventional AI trajectory, aligning models to perform complex reasoning requires RLHF or process reward models explicitly annotated by humans. In specialized domains, e.g., oncology 
or quantum mechanics, crowdsourcing human evaluators to grade intermediate reasoning steps is very expensive and unscalable. This creates a ceiling on how deeply a generalist model can be aligned to reason in niche fields.

Using abstractions as implicit reward models shatters this ceiling, fundamentally enabling the neurosymbolic vision of DSS. When a KG or formal logic system acts as the reward mechanism, the training paradigm shifts from human preference to axiomatic correctness. The abstraction provides a deterministic, verifiable scaffold. During the RL phase, the model is forced to explicitly prioritize coherent, logically sound steps over probabilistic shortcuts or hallucinated links. This ensures that the SLM internalizes the strict compositional rules of its domain, transforming it from a fluent text generator into a rigorous deductive engine.

Within the DSS society, this rigorous process supervision is the foundation of trust and interoperability. When the front-end orchestrator routes a complex query to a back-end Medical DSS, it must trust that the specialist will not hallucinate an interaction pathway. Furthermore, when multiple DSS models debate a solution, they must communicate logically, not just persuasively. Because each DSS agent has been reinforced via an implicit reward model grounded in factual abstraction, their internal monologues and outputs remain tethered to verifiable ground truths. This creates an AI ecosystem where deep reasoning is mathematically steered, ensuring the entire society operates with the precision and reliability demanded by high-stakes real-world applications.

%% file: sections/8_DSS_society_new.tex
\section{Domain-specific superintelligence}
\label{sec:dss}

Having shown the limits of scale and by synthesizing the three pillars discussed thus far: high-quality data, robust abstractions, and advanced neurosymbolic reasoning, as well as DSS building blocks, we arrive at the core proposal of this alternative trajectory: the creation of DSS societies. The previous section introduced DSS as specialized systems that achieve superhuman depth within bounded domains. This section focuses on the next level of organization: how many such DSS models can be composed, coordinated, and verified as a collective intelligence system.

A single DSS can provide exceptional competence within one domain but real-world problems rarely respect clean disciplinary boundaries. A clinical decision may involve endocrinology, nephrology, cardiology, pharmacology, radiology, medical guidelines, insurance constraints, and patient behavior. A scientific discovery task may require literature understanding, mathematical modeling, simulation, robotics, experimental design, and safety assessment. 
Thus, individual DSS models solve the problem of depth but a broader system must also solve the problem of breadth. We propose that this breadth should not be achieved by forcing all capabilities into one monolithic generalist model but by organizing many DSS models into a coordinated society.

\subsection{DSS society as an organized intelligence system}

A DSS society is not merely a collection of independently trained expert models. It is an organized computational ecosystem in which multiple DSS models cooperate through explicit roles, interfaces, communication protocols, and verification mechanisms. Each DSS model is responsible for a bounded domain in which it can maintain deep expertise, efficient adaptation, and domain-specific alignment. The society-level system then coordinates these models to solve complex tasks that exceed the scope of any single DSS. In this sense, the unit of intelligence is no longer an isolated model, but a structured system of interacting specialists.

This distinction is crucial. A loose ensemble of domain models does not automatically produce reliable collective intelligence. If specialist models simply exchange unconstrained natural-language messages, the resulting system may suffer from ambiguity, 
inconsistent assumptions, duplicated work, and unresolved conflicts. A DSS society must, therefore, specify how tasks are decomposed, how subproblems are assigned, how intermediate results are represented, how models communicate, how disagreement is resolved, and how final outputs are verified. The ``society'' metaphor should thus be understood architecturally: It has division of labor, shared abstractions, routing mechanisms, and governance procedures that allow many specialized intelligences to operate as a coherent whole.

The motivation for such an organization is analogous to the organization of human expertise. Modern civilization does not rely on one universal expert to master medicine, law, engineering, science, economics, and policy simultaneously. Instead, it relies on networks of specialists connected through institutions, standards, protocols, documents, tools, and verification procedures. Physicians, engineers, scientists, 
and policymakers coordinate through structured artifacts such as medical records, mathematical models, 
scientific papers, technical specifications, and regulatory standards. A DSS society follows the same principle computationally. 

\subsection{A workflow architecture for DSS society}

A DSS society should preserve the original workflow intuition of a front-end orchestrator, decomposition and routing, and inter-DSS collaboration, while adding an explicit verification stage. The goal is not to simply call multiple specialist models but organize them into a reliable pipeline in which user intent is interpreted, complex tasks are decomposed, domain experts are activated, specialists communicate, and the final response is checked before being returned. This workflow can be described through four tightly connected components.

\paragraph{Front-end orchestrator.}
An SLM trained on the whole domain (e.g., all topics in medicine or physics) serves as the user interface and the entry point to a query. It acts as the first-level router that parses the user's intent, identifies the broad domain of the request, and dispatches the query to the appropriate ``back-end'' DSS models. For example, a query related to diabetes would first be recognized as a medical query and then routed by a medical DSS or medical orchestrator to a diabetes DSS. 
Similarly, a
query about circuit design would be routed to an electrical-engineering DSS. The orchestrator is not expected to replace the back-end DSS models or contain all specialized knowledge itself. Rather, its role is to understand the user's goal, determine which expertise is needed, and initiate the proper society-level workflow.

The front-end orchestrator also plays an important role in interaction management. Many user queries are underspecified, ambiguous, or context-dependent. Before invoking specialist DSS models, the orchestrator may need to ask clarifying questions, retrieve relevant user context, normalize terminology, or convert the original request into a more structured task representation. In this sense, the orchestrator serves as the society's conversational and control interface: It connects human intent to the internal organization of DSS expertise.

\paragraph{Decomposition and routing.}
For complex, multi-faceted inquiries, the orchestrator breaks the problem into sub-queries and routes each sub-query to the appropriate DSS model. This step is essential because many high-value tasks require multiple types of expertise. 
A question about a novel way to model the dynamics of a physical system can be directed to physics, mathematics, and simulation DSS models. A clinical question about diabetic kidney disease may require a diabetes DSS, a nephrology DSS, a cardiology DSS, a pharmacology DSS, and a clinical-guideline DSS.

Effective decomposition requires the orchestrator to identify not only the relevant domains but also their dependencies. Some subproblems can be solved in parallel whereas others require sequential reasoning. E.g., a pharmacology DSS may need the diagnosis and kidney-function assessment from other medical DSS models before checking for drug safety. 
Thus, routing is not a static lookup table from query type to expert model. It is a dynamic planning process: which DSS models should be called, in what order, with what context, and with what expected output format.

\paragraph{Inter-DSS collaboration.}
In advanced scenarios, DSS models do not merely report independently to the user or the orchestrator; they communicate with each other. The engineering DSS might request clarification from a physics DSS before finalizing its assessment. A cardiology DSS might ask a pharmacology DSS whether a proposed medication conflicts with cardiovascular risk factors. 
This interchange of findings mimics a team of human experts consulting on a case: Each specialist contributes domain-specific reasoning while responding to the questions from other specialists.

Inter-DSS collaboration is especially important when the boundary between domains is not clean. Many real-world problems contain coupled constraints: medical decisions interact with patient behavior and insurance policy; engineering designs interact with safety regulation and manufacturing cost; scientific hypotheses interact with experimental feasibility and available instruments. In these cases, the output of one DSS model becomes the input of another. The DSS society, therefore, needs mechanisms for iterative consultation, intermediate feedback, and revision. Rather than producing a one-shot answer, the DSS society can refine its reasoning through structured expert dialog.

\paragraph{Society-level synthesis and verification.}
After the relevant DSS models have produced their outputs, the system needs an explicit synthesis and verification stage before returning the final answer. This stage integrates the findings from multiple DSS models, checks whether they are mutually consistent, resolves conflicts, estimates uncertainty, and verifies the response against domain-specific constraints. Without such a stage, a DSS society would risk becoming a collection of expert opinions without a reliable mechanism for deciding which opinions should be trusted.

Verification can take different forms depending on the domain. In medicine, the final answer may be checked against clinical guidelines, contraindication rules, patient-specific constraints, and evidence provenance. In engineering, it may be checked against physical laws, simulation results, design margins, and safety standards. In mathematics, it may be checked through formal proof systems, such as Lean. The key point is that verification should not be left only to the generative ability of the same model that produced the answer. It should use domain-specific abstractions, symbolic constraints, tools, and validators whenever possible.

This final stage turns a DSS society from a collaborative model ensemble into a trustworthy intelligence system. The orchestrator and specialist models generate candidate reasoning paths but the synthesis and verification stage audits those paths before producing the final response. It can identify unsupported claims, expose disagreements among DSS models, request additional consultation, or escalate to a human expert when uncertainty is high. 

\subsection{Lego-like composability of DSS models}

A central advantage of DSS society is ``Lego-like'' composability. Each DSS model can be viewed as a specialized unit of intelligence with a defined domain boundary, input-output interface, and verification standard. Once these interfaces are standardized, models can be added, removed, updated, or recombined without retraining the entire system. This is a major advantage over monolithic generalist models, where improving one domain may require expensive global retraining and may unintentionally affect unrelated capabilities.

For example, a medical DSS society can incorporate a new rare-disease DSS model when new data, guidelines, or expert models become available. The rest of the system does not need to be rebuilt from scratch. The orchestrator only needs to learn when this new DSS model is relevant, how to route subproblems to it, and how to integrate its outputs with other medical models. Similarly, a scientific-discovery society can add a new molecular-simulation DSS model, a new robotic-lab DSS model, or a new safety-evaluation DSS model as independent components. 

Composability also supports continual improvement. Domains evolve at different speeds. Medical guidelines may change yearly, 
scientific fields may change through new discoveries, and software tools may change through new releases. A modular DSS society enables each domain component to be updated according to its own schedule and validation process. 
This modularity makes a DSS society scalable in both technical and institutional terms. Technically, it avoids the need to centralize all computation in one giant model. Institutionally, it allows different communities to build, audit, and maintain DSS models for their own domains. Hospitals, research labs, universities, and companies can contribute specialized models with explicit standards and interfaces. 

\subsection{DSS society as a practical path toward broad intelligence}

The long-term significance of DSS society is that it offers a practical path toward broad intelligence through composition rather than monolithic scaling. Instead of requiring one general-purpose model to master every domain, a DSS society can approach broad competence by accumulating, coordinating, and verifying many narrow superintelligences. Breadth emerges from the organized interaction of depth-specialized models.

This path has several advantages. First, it is more efficient. Individual DSS models can be smaller than generalist foundation models because they operate within bounded domains and can rely on domain-specific data and abstractions. Second, it is more reliable. Each model can be evaluated using domain-specific benchmarks, constraints, and expert standards. Third, it is more interpretable. Failures can often be localized to a specialist model, an orchestration decision, a communication interface, or a verification step. Fourth, it is more updatable. New knowledge can be incorporated into the relevant DSS model without disturbing the entire system.

This architecture also reframes the question of AGI. If a society of DSS models can outperform human experts in every specific domain, diagnosing diseases better than physicians, 
coding better than engineers, discovering materials better than scientists, and tutoring students better than teachers, then the practical need for a single unified ``general'' intelligence becomes less clear. The utility lies in the excellence of the parts and the coordination of the whole, not necessarily in the existence of one model that internally contains all capabilities.

In this sense, a DSS society approaches broad intelligence asymptotically. As the society expands to cover more knowledge domains, and as the orchestration and verification layers become more capable, the collective system can exhibit increasingly general intelligence. However, this generality is achieved by modular growth rather than by collapsing all knowledge into one undifferentiated model. The society becomes broader by adding new specialists, improving existing specialists, standardizing interfaces, and strengthening mechanisms for communication and verification.

This perspective also provides a more grounded route toward safe and useful AI. A monolithic generalist system is difficult to inspect, align, and govern because its knowledge and capabilities are deeply entangled. A DSS society, by contrast, makes specialization, responsibility, and verification more explicit. Each model can be aligned with the standards of its domain whereas the society can enforce routing, evidence tracking, uncertainty estimation, conflict resolution, and human oversight when needed. 

%% file: sections/9_Agents.tex
\section{AI agents based on DSS societies}
\label{sec:agents}

The agent layer is the operating substrate of the DSS trajectory. A trained DSS SLM can be more effective if surrounded by modules for perception, memory, world modeling, and action. The agent loop, in which perception, cognition, and action repeat over time, is the structure that lets a DSS SLM turn its internalized abstractions into closed-loop, grounded behavior. 
The SLM contributes cognition, symbolic abstractions contribute semantic memory and a world model, and specialized machine learning models contribute perception and tool use. 

The agent layer is also what turns isolated DSS SLMs into a functioning DSS society. Orchestrator-and-experts routing, multi-agent debate, emergent specialization, and standardized inter-agent protocols 
are the 
mechanisms by which heterogeneous specialists negotiate, collaborate, and arbitrate disputes across domains. Because DSS SLMs are small, these protocols also enable the society to be physically decentralized, federated across edge devices rather than being concentrated in a single cloud. A closed-loop AI Scientist pattern and system-level continual learning extend the picture: New facts flow into shared KGs and episodic memory, new specialists are composed on demand, and the society's collective intelligence grows without retraining any single model. The agent layer is where the \textit{Society of Mind} architecture sketched earlier becomes a concrete operational pattern.

\begin{figure}[ht]
    \centering
    \includegraphics[width=\textwidth]{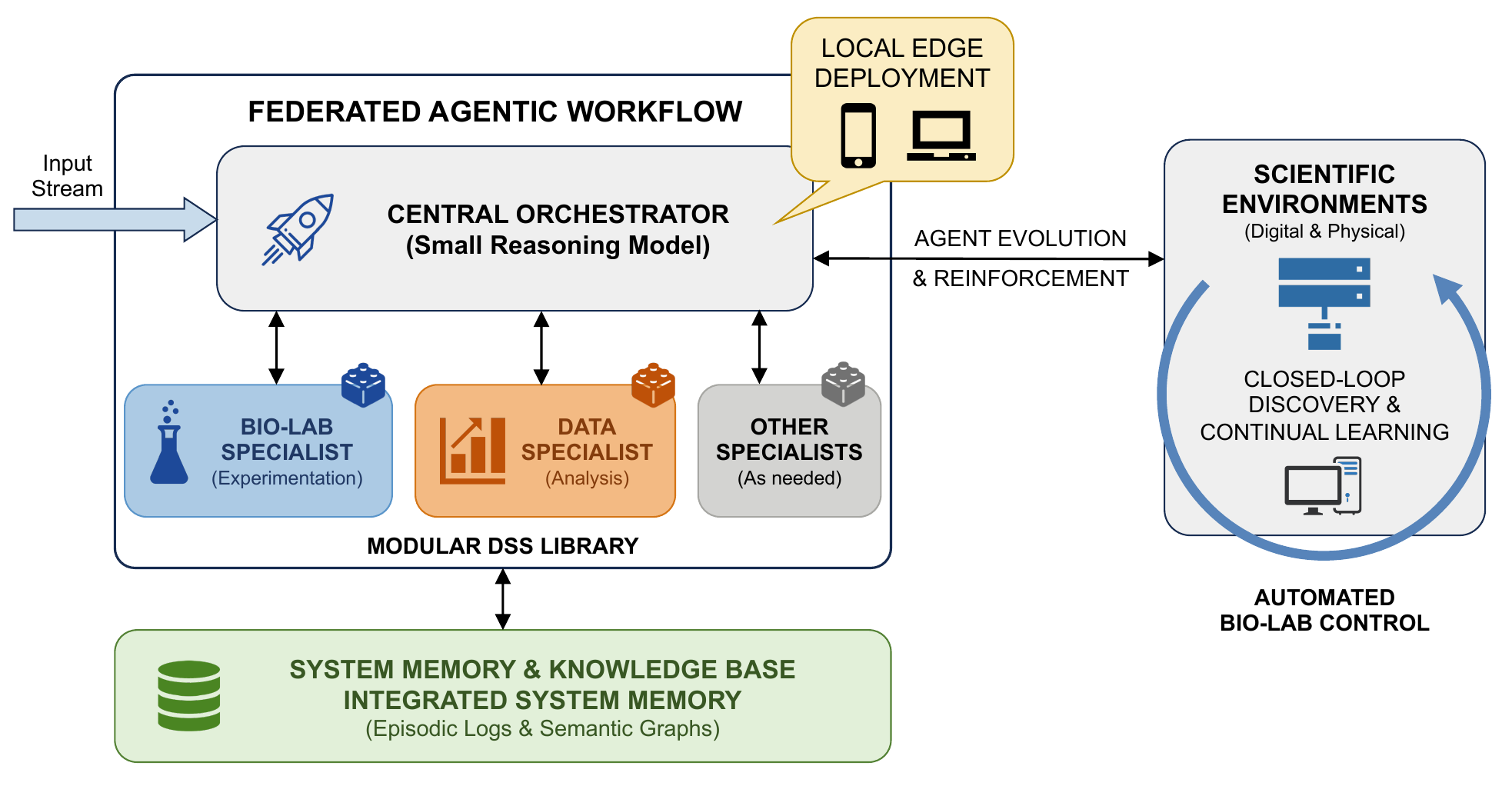}
    \caption{A possible agentic workflow based on a DSS society. A central orchestrator (a fine-tuned SLM) routes tasks to a 
    library of modular DSS specialists. The architecture supports local edge deployment for energy efficiency and integrates a continuous ``AI Scientist'' closed-loop where automated physical/digital experiments continually update the shared system memory.}
    \Description{An input stream enters a central orchestrator,
  implemented as a small reasoning model within a federated
  agentic workflow. Bidirectional arrows connect the
  orchestrator to a modular library containing a bio-lab
  specialist for experimentation, a data specialist for
  analysis, and other specialists as needed. The workflow
  exchanges information with shared system memory containing
  episodic logs and semantic graphs. A callout indicates
  local deployment on phones and computers. A bidirectional
  connection labeled agent evolution and reinforcement links
  the orchestrator to digital and physical scientific
  environments, where circular arrows represent closed-loop
  discovery, continual learning, and automated bio-lab control.}
    \label{fig:dss_architecture}
\end{figure}

\subsection{Architecture: Societies of DSS models}

No single DSS SLM, no matter how deep its expertise, can handle real-world tasks that span multiple domains. As 
illustrated in Fig.~\ref{fig:dss_architecture}, the solution is to integrate various modules around a society of DSS models: A central front-end orchestrator divides a user query into sub-queries, dispatches them to a modular library of specialist DSS back-ends, and integrates their responses. 

\subsubsection{Why a heterogeneous DSS society may outperform a monolith model as the agent brain}

The key advantage of a DSS society is heterogeneity: Each back-end SLM is trained on different domain data and paired with a different symbolic abstraction (e.g., a KG); hence, the society as a whole covers far more ground than any single model~\citep{agents_multiagent_collab}. This heterogeneity operates at multiple levels. At the persona level, DSS SLMs take on distinct specialist roles: a programmer, product manager, and tester in a software development team (MetaGPT~\citep{agents_metagpt}, ChatDev~\citep{agents_chatdev}), or cardiologists, oncologists, and pediatricians in a medical consultation. Each is optimized for its function, yet collectively they outperform a single massive model~\citep{agents_multiagent_collab}. At the action-space level, the front-end orchestrator SLM decomposes tasks and routes queries to the appropriate specialist back-end. MasRouter~\citep{agents_masrouter} shows that intelligent routing to appropriately sized models achieves 1.8-8.2\% performance gains while reducing computational overhead, consistent with the DSS prediction that capability can be decoupled from model size.

\paragraph{Emergent specialization}
Even initially homogeneous agents can spontaneously develop distinct roles through repeated interactions. 
In Project Sid~\citep{agents_project_sid}, hundreds of agents evolve professions, trade networks, and even cultural practices without explicit programming. This echoes Minsky's ``Society of Mind'' hypothesis~\citep{agents_minsky_society}: Intelligence emerges from the interactions of many specialized sub-agents.

\paragraph{Robustness through diversity.}
Even when one DSS back-end fails, the DSS society and agent persist. 
This avoids ``correlated failures,'' where a single blind spot in a monolithic model causes total system failure, a risk inherent in concentrating all knowledge in one set of weights. 

\subsubsection{How DSS models collaborate}

The front-end/back-end routing described above is the simplest task-oriented collaboration pattern: The orchestrator SLM decomposes a high-level objective into sub-tasks and dispatches them to specialist DSS back-ends, each processing upstream artifacts (intermediate code, data structures, evaluation results) and producing downstream outputs~\citep{agents_multiagent_collab}. When a problem requires reconciling conflicting domain perspectives, DSS societies can also use consensus-oriented debate: Specialist SLMs with different domain grounding debate a solution and each must justify its reasoning against its own verifiable symbolic abstraction, which reduces hallucination and bias. The RECONCILE~\citep{agents_reconcile} and MAD~\citep{agents_mad_debate} frameworks show measurable improvements in decision quality through such structured argumentation.

\paragraph{Communication protocols}
For DSS societies to scale, the modules need standardized interfaces~\citep{agents_protocol_survey}: MCP for centralized tool integration, ANP for decentralized cross-platform agent networks, A2A for web-native agent-to-agent communication, and IoA for task-based team coordination. Because individual DSS SLMs are small enough to run on edge devices (Section~\ref{sec:energy}), these decentralized protocols also enable societies where some back-ends run on-device rather than in the cloud, distributing thermal load and reducing dependence on centralized data centers.

\subsection{Implementation: Mapping DSS components to agentic frameworks}

Translating the DSS society into working code requires concrete agentic frameworks. The \textit{Foundation Agent} framework~\citep{agents_foundation_survey} and the \textit{CoALA} architecture~\citep{agents_coala} provide two blueprints, both of which map well onto DSS components.

\subsubsection{The agent loop and its DSS realization}

The Foundation Agent framework~\citep{agents_foundation_survey} defines each agent as a system of interacting modules organized in an \textit{Agent Loop}, a continuous cycle of Perception~($\mathrm{P}$), Cognition~($\mathrm{C}$), and Action Execution~($\mathrm{E}$). At each time step, the agent perceives the environment state $s_t$ to produce observation $o_t = \mathrm{P}(s_t, M_{t-1})$, updates its mental state, selects an action through cognition $(M_t, a_t) = \mathrm{C}(M_{t-1}, a_{t-1}, o_t)$, and executes the action to transition the environment.

In DSS terms, the SLM implements the reasoning function $\mathrm{R}$ within cognition whereas the symbolic abstraction, such as a KG, implements the semantic memory $M_t^{\text{mem}}$ and world model $M_t^{\text{wm}}$. The perception function $\mathrm{P}$ is where multimodal DSS models play a role: A medical agent, for instance, routes text inputs to an electronic health record specialist SLM and imaging inputs to a radiology-specialist SLM, fusing their outputs into a joint observation $o_t$. This separation of reasoning from knowledge storage follows the ``abstraction first, then generalization'' principle: Symbolic backends, such as the KG, provide structured abstractions and the SLM generalizes from them. The execution module grounds actions in the physical or digital world by invoking external APIs and specialized models~\citep{agents_foundation_survey}. It connects to the Machine Learning as a Tool (MLAT) paradigm described ahead.

\subsubsection{CoALA: A natural home for DSS}

\textit{CoALA (Cognitive Architectures for Language Agents)}~\citep{agents_coala} can easily be adapted as a blueprint for DSS agents. Instead of centering on an LLM, CoALA could center on a fine-tuned SLM as the orchestrator and surround it with DSS models and modular memory that maps onto DSS components:

\begin{itemize}
  \item \textbf{Procedural knowledge} resides in the SLM weights (implicit) and in executable code (explicit). This is the domain reasoning capability trained via the synthetic curricula described in Section~\ref{sec:good_data_importance}.
  \item \textbf{Semantic knowledge} is stored in the symbolic abstractions described in Section~\ref{sec:abstraction} that ground SLM reasoning in verifiable facts.
  \item \textbf{Episodic knowledge} accumulates from past interactions, enabling the agent to learn from experience without retraining the SLM.
\end{itemize}

\noindent CoALA also distinguishes internal actions (reasoning, retrieval, learning by writing to memory) from external actions (interacting with the physical world, digital interfaces, or humans). This gives a clean interface between the DSS reasoning core and the tools it orchestrates.

\subsubsection{MLAT: Machine learning as a tool}

A natural extension of tool use is MLAT~\citep{agent_mlat}. Rather than asking a single generalist model to perform every cognitive task, a DSS agent acts as an orchestrator that calls specialized machine learning models as tools, each a domain expert like a DSS SLM but implemented as a task-specific model:

\begin{itemize}
  \item Instead of an LLM trying to predict a protein structure (which it is bad at), it calls AlphaFold~\citep{jumper2021highly}. ChemCrow~\citep{agents_chemcrow} integrates chemistry-specific tools for molecular design and HoneyComb~\citep{agents_honeycomb} provides domain-specific tool integration for materials science.
  \item Instead of forecasting stock prices via token prediction, the agent calls a specialized autoregressive integrated moving average or long short-term memory time-series model. This reduces hallucination and computational cost while increasing precision.
\end{itemize}

\subsection{Case study: The AI scientist and closed-loop discovery}

A demanding test case is scientific discovery, which requires precision grounded in domain knowledge, long-term planning, and synthesis of large structured knowledge bases. These are capabilities that monolithic LLMs lack but that DSS agents, with their KG-stored knowledge and SLM reasoning cores, are designed to provide. The Foundation Agents survey~\citep{agents_foundation_survey} introduces the concept of the ``AI Scientist'' and provides a formal intelligence-theoretic framework for how agents improve through scientific inquiry.

\subsubsection{The closed-loop innovation cycle}

The AI Scientist is a self-sustaining innovation cycle where agents continuously expand the boundaries of knowledge without human intervention. The Foundation Agents survey~\citep{agents_foundation_survey} formalizes agent intelligence as a Kullback-Leibler divergence measure: $\text{IQ}_t^{\text{agent}} \equiv -D_{\text{KL}}(\theta, M_t^{\text{mem}})$, where intelligence grows monotonically as the agent acquires knowledge that reduces the divergence between its internal model and the true world distribution. This formalization identifies scientist agents as curiosity-driven agents whose expected intelligence gain is highest when new measurements are most unexpected.

The closed-loop process works as follows:

\begin{enumerate}
  \item \textbf{Hypothesis generation:} The agent uses its world model (KG) to identify gaps in current knowledge and uses reasoning (SLM) to formulate testable hypotheses~\citep{agents_scientist_survey}. In controlled studies, language model-generated research ideas were rated as more novel than human expert ideas ($p<0.05$)~\citep{agents_si_llm_ideas}. SciAgents~\citep{agents_sciagents} uses ontological KGs to generate hypotheses along with testing methods for materials science.
  \item \textbf{Experiment design and execution:} In digital domains, the agent writes and executes code~\citep{agents_ai_scientist_sakana}. In physical domains, it controls robotic laboratories~\citep{agents_embodied_systems}. The Genesis system~\citep{agents_genesis} is aimed at controlling 1,000 micro-bioreactors and running 1,000 hypothesis-driven closed-loop experimental cycles per day; earlier proof-of-principle work advanced a yeast diauxic shift model by +92 genes (+45\%) and +1,048 interactions (+147\%).
  \item \textbf{Analysis and feedback:} The agent analyzes results using specialized tools, like MLAT, and compares them against predictions from its world model~\citep{agents_scientist_survey}.
  \item \textbf{Knowledge consolidation:} The agent updates its KG with new findings~\citep{agents_ai_scientist_sakana}. This learning is explicit and permanent, unlike the transient context of an LLM. ChemAgent~\citep{agents_chemagent} demonstrates self-updating memory with dynamic updates that incorporate correct chemistry answers, achieving performance gains up to 46\% on four SciBench chemical reasoning datasets.
  \item \textbf{Publication and dissemination:} The agent writes a paper that summarizes its findings to share with other agents, enabling collective knowledge accumulation~\citep{agents_ai_scientist_sakana}. Agent Laboratory~\citep{agents_agent_laboratory} autonomously conducts literature review, experimentation, and report writing, achieving a human-evaluated experiment quality score of 3.2 out of 5 across five research questions in computer vision and NLP.
\end{enumerate}

\paragraph{Physical-world closed loops}
The closed-loop paradigm extends beyond digital experiments. Autonomous mobile synthesis systems integrate mobile robots, automated synthesis equipment, and characterization instruments into a synthesis-analysis-decision cycle. In one demonstration, five geographically distributed laboratories across three continents connect to a cloud-based experiment planner that continuously learns from incoming data, discovering 21 state-of-the-art organic solid-state laser materials~\citep{agents_delocalized_discovery}. ChemOS 2.0~\citep{agents_chemos2} integrates \emph{ab initio} calculations, experimental orchestration, and Bayesian optimization as a world model $M_t^{\text{wm}}$ to make updates based on experimental results.

\subsubsection{Continual learning}

For an AI system to function in the real world, it cannot rely solely on static knowledge acquired during training. It must be capable of acquiring new skills, incorporating new information, and adapting to changing environments over its operational lifetime~\citep{lifelong_learning}. This is known as continual learning (CL), lifelong learning, self-evolution, or incremental learning. In the context of LLMs, CL is often framed as a computational maintenance challenge: how to update the model's parameters to reflect new data without catastrophic forgetting of previously learned capabilities~\citep{cl_survey}. This tension between incorporating new information (plasticity) and retaining prior knowledge (stability) constitutes the central stability-plasticity dilemma~\citep{zheng2026lifelong}.
However, as we argue for an alternative trajectory focused on societies of SLMs, our proposed architecture implicitly enables CL. As mentioned in the survey~\citep{agents_foundation_survey}, ``All manually designed agentic systems will eventually be replaced by learnable, self-evolving systems,'' agents can optimize their own prompts, workflows, and tool usage based on feedback. Unlike a static LLM frozen after training, a DSS agent 
creates a flywheel: Better tools lead to better experiments, which produce better data, which yield smarter agents~\citep{agents_self_evolving_survey}. This reinforces the data-quality principle from Section~\ref{sec:good_data_importance}.

For agentic systems built atop DSS models, CL shifts from a model-level problem (updating parameters while avoiding catastrophic forgetting) to a system-level one. Rather than through costly retraining, agents adapt by accumulating episodic memory, expanding their underlying symbolic abstractions with newly discovered facts, and refining tool usage strategies. This decouples knowledge acquisition from parameter optimization, sidestepping the stability-plasticity dilemma~\citep{zheng2026lifelong} that plagues monolithic models. A further possibility is that the expanded abstractions and accumulated episodic memory can seed new synthetic curricula (Section~\ref{sec:good_data_importance}), enabling the SLM itself to be continually and periodically fine-tuned on high-quality, grounded agent-generated data at far lower cost than full retraining.

\paragraph{Mechanisms}
The survey~\citep{agents_foundation_survey} identifies three primary mechanisms:

\begin{itemize}
  \item \textbf{Memory-based learning:} Agents record outcomes and world states, then identify successful versus failed strategies through reflection. This improves performance without changing SLM weights. In clinical simulations, doctor agents improved treatment through experiential memory.
  \item \textbf{Shared memory-based learning:} Agents pool experiences via shared memory. G-Memory~\citep{agents_gmemory} implements a hierarchical three-tier memory system (insight, query, and interaction graphs) that achieves a 20.89\% improvement in complex embodied tasks through cross-trial learning.
  \item \textbf{Collective intelligence optimization:} EvoAgent~\citep{agents_evoagent} uses genetic algorithms to evolve agent configurations and outperforms manually designed systems.
\end{itemize}

\paragraph{Challenges and limitations}
The AI Scientist paradigm still has capability gaps: Operating physical laboratory devices remains at Technology Readiness Levels (TRLs) 4-6~\citep{agents_foundation_survey}, i.e., validated or demonstrated in a laboratory or relevant environment on the standard 1-9 TRL scale, but not yet in real-world operation, and reconciling conflicting information from multiple sources (including paywalled data and empirical expert heuristics not captured in text) is an open problem.

%% file: sections/10_Energy.tex
\section{Energy efficiency and economic sustainability}
\label{sec:energy}

Energy is not an afterthought to the DSS trajectory; it is the constraint that motivates it. Frontier model training already requires large amounts of electricity~\citep{ier_orion_energy}, but inference now accounts for most AI computing~\citep{ai_inference_costs_2025}. Reasoning models can raise the energy cost of each query further, although the size of the increase varies widely with the model, the workload, and the infrastructure serving it~\citep{ai_hungry_benchmark_2025}. Once the optimization target moves from capability at any cost to capability-per-Watt and capability-per-dollar, compact domain-grounded DSS SLMs become competitive with, and frequently dominate, monolithic generalists. Specialized hardware (FP4 GPUs, LPUs with on-chip SRAM, mobile NPUs delivering tens of TOPS at single-digit Watts) gives DSS a concrete physical substrate. A 7-14B model paired with a KG can deliver expert-level reasoning at roughly 0.005-0.01~J/token on a smartphone, against tens of Joules per token in the cloud. Thus, energy efficiency is one of the main reasons to pursue DSS in the first place.

The energy and economic analysis is also what makes a DSS society possible at planetary scale. A Stargate-class monolith
(joint venture among OpenAI,
Microsoft, Oracle, and SoftBank) concentrates several GW of demand at a handful of sites, with the grid, water, and geopolitical bottlenecks already visible in Northern Virginia and Ireland. A DSS society inverts this geometry: Billions of edge devices, each drawing only a few Watts and federated through lightweight agent protocols, distribute thermal load across existing infrastructure rather than overloading any single node. This is what makes Sovereign AI, the \textit{Pocket Sage}, and resilience through diversity deployable at the scale of billions of users. The economic asymmetries documented in this section, including the 11.4$\times$ price gap between premium reasoning APIs and efficient alternatives, the bifurcation between commodity and premium intelligence, and Jevons-paradox-driven demand growth, determine who can participate in the DSS society and on what terms. 

\subsection{Lifecycle accounting: Energy and carbon}

To accurately assess the environmental impact of AI systems, we account for \textit{energy} and \textit{carbon} separately.

\paragraph{Energy accounting}
Total lifecycle energy consumption is the sum of training and cumulative inference:
\begin{equation}
E_{\text{total}} = E_{\text{train}} + \sum_{t=0}^{T} E_{\text{inference}}(t),
\end{equation}
where $E_{\text{train}}$ is a one-time expenditure (order of GWh to TWh for frontier models) and $E_{\text{inference}}(t)$ scales with query volume and complexity over the deployment period $T$.

\paragraph{Carbon accounting}
Total lifecycle carbon emissions incorporate grid carbon intensity and facility overhead:
\begin{equation}
C_{\text{total}} = C_{\text{embodied}} + \mathrm{CI}_{\text{train}} \cdot \mathrm{PUE}_{\text{train}} \cdot E_{\text{train}} + \sum_{t=0}^{T} \mathrm{CI}(t) \cdot \mathrm{PUE}(t) \cdot E_{\text{inference}}(t),
\end{equation}
where $C_{\text{embodied}}$ is the embodied carbon of hardware manufacturing, $\mathrm{CI}(t)$ is the grid carbon intensity (kg CO$_2$e/kWh), and power usage effectiveness, $\mathrm{PUE}(t)$, captures facility overhead. Typical PUE ranges from $\sim$1.1 for hyperscale facilities to $\sim$1.5-1.6 for legacy data centers~\citep{pew_datacenter_energy, google_env_2025}.

\paragraph{The inference crossover point}
This is the time, $t_{\text{cross}}$, at which cumulative inference energy equals training energy:
\begin{equation}
t_{\text{cross}} \approx \frac{E_{\text{train}}}{N_{\text{req/day}} \cdot e_{\text{req}}},
\end{equation}
where $N_{\text{req/day}}$ is the daily request volume and $e_{\text{req}}$ is the energy consumed per request.

For high-traffic deployments, this crossover occurs rapidly. 
Across the industry, inference is estimated to consume 80-90\% of all AI computing power~\citep{ai_inference_costs_2025}, which supports our focus on efficient, modular inference architectures over monolithic training runs.

\paragraph{Energy scaling and architectural asymmetry}
Training energy scales with several interacting factors:
\begin{equation}
E_{\text{train}} \propto \frac{D \cdot C}{\eta}, \qquad C \approx 6N,
\end{equation}
where $D$ is the number of training tokens, $C$ is the training compute per token, which already scales with the parameter count $N$ (about $6N$ FLOPs for the forward and backward passes of a dense Transformer~\citep{kaplan2020scaling}), and $\eta$ is the effective hardware efficiency (useful FLOPs per Joule). Larger models also lower $\eta$ because they increase memory bandwidth pressure and communication overhead. Long-context inference further introduces quadratic complexity in attention mechanisms. In practice, scaling model parameters by 10$\times$ does not yield a proportional gain in reasoning depth but increases memory footprint near-proportionately and training energy at least proportionately: 10$\times$ at a fixed $D$; more when $D$ grows with $N$ or when $\eta$ drops. This asymmetry undermines the economic rationale of indefinite scaling. In addition, the success of large models paradoxically accelerates their energetic burden as they become embedded in consumer products and enterprise systems.

\subsection{Economic analysis: Unit economics and market bifurcation}

Commercializing AI requires sound unit economics. At product scale, inference cost determines gross margins: If the marginal inference cost approaches 
revenue per query, deployment becomes economically fragile. Large generalist models impose persistent high per-token compute costs, incentivizing centralization among a few capital-rich actors.

\paragraph{The hidden token cost structure}
Reasoning models generate thousands of invisible CoT tokens that are billed but not shown to users. For Claude's Opus 4.6 model, these are priced as output tokens at \$25/1M. A single complex query could cost several dollars in reasoning tokens alone, making flat-rate subscriptions economically unsustainable~\citep{claude_pricing}.
In contrast, DeepSeek-R1 launch price is $\sim$\$2.19/1M output tokens: an 11.4$\times$ price differential that reflects not merely margin compression but fundamental differences in training efficiency and model architecture~\citep{deepseek_cost, deepseek_r1_pricing}.

\paragraph{Market bifurcation}
The market is thus fragmenting into two distinct segments:
\begin{itemize}
  \item \textbf{Commodity intelligence}: DeepSeek-style distilled models running cheaply on edge devices or older hardware~\citep{deepseek_vs_gpt_leanware}. Users pay primarily for electricity, bypassing API margins.
  \item \textbf{Premium reasoning}: Orion/Stargate-class systems that remain centralized, expensive, and energy-intensive, reserved for high-value enterprise tasks~\citep{datacenter_cost_structure}.
\end{itemize}

\paragraph{The Jevons paradox}
Economic theory predicts that efficiency gains increase demand elastically~\citep{ jevons_paradox_ai, jevons_paradox_digitopoly}. If inference cost drops 90\% due to efficient SLMs, we are unlikely to see a 90\% reduction in energy use; instead, new applications (always-on agents, video generation, ubiquitous assistants) will absorb the savings.
Efficiency alone does not guarantee sustainability; it must be paired with deliberate constraints or carbon pricing.

\subsection{Productivity and labor implications}

AI's economic impact extends to the labor market, where electrical energy substitutes for human cognitive effort~\citep{staffing_ai_jobs, ai_job_impact}.

\paragraph{White-collar turbulence}
Generative AI primarily augments or automates cognitive, white-collar tasks. Reasoning models specifically target high-order tasks, such as software engineering and mathematical problem-solving. Roles involving routine cognitive work (coding, translation, basic analysis) face the highest displacement risk. A ``reasoning wage premium'' is emerging that benefits workers who can orchestrate these models, while a ``growth without hiring'' trend sees companies expanding output without proportional headcount increases.

\paragraph{Blue-collar resilience}
Physical jobs (plumbing, nursing, construction) remain largely unaffected due to \textit{Moravec's Paradox}: Automating dexterous sensorimotor skills is harder than automating high-level reasoning. The net effect may be \textit{wage compression}, where skilled trades become relatively more valuable compared to mid-tier cognitive labor.

\subsection{Domain-specific superintelligence: A sustainable path forward}

Energy constraints, economic pressure, and diminishing returns from monolithic scaling together point toward our alternative trajectory: 
DSS. Rather than pursuing AGI through ever-larger monolithic models, DSS replaces general-purpose giants with a coordinated ecosystem of specialized, efficient models---a \textit{Society of Mind} where domain experts collaborate. 
Thus, intelligence is not defined by parameter count but by task-relevant performance per unit of energy.

\paragraph{The DSS architecture}
DSS relies on SLMs trained to reach expert-level depth in narrow domains (e.g., organic chemistry, 
avionics diagnostics) rather than broad, shallow recall:
\begin{itemize}
  \item \textbf{Data efficiency}: Instead of scraping the entire Internet, DSS models are trained on high-quality, curated datasets (textbooks, proprietary corporate data, scientific papers). This ``high-signal'' training enables models with 7B-14B parameters to outperform LLMs in their specific niche domain.
  \item \textbf{Neurosymbolic scaling}: DSS uses abstractions like KGs to ground reasoning (see Section~\ref{sec:abstraction}). This enables verifiable, symbolic reasoning (GraphRAG, MCTS) that is more efficient and accurate than the probabilistic output of standard LLMs. By offloading factual recall to a KG, SLMs can focus on reasoning, reducing its parameter count and, crucially, avoiding the ``inference inflation'' of massive CoT generation.
\end{itemize}

\paragraph{Energy advantages of DSS}
The energy profile of a DSS-based ecosystem improves 
upon the Stargate paradigm:
\begin{itemize}
  \item \textbf{Training}: Training a 7B-parameter DSS requires a fraction of the energy of that needed for a frontier model. DeepSeek-V3's \$5.5M training run shows that competitive intelligence can be built without GW-scale campuses, lowering the barrier to AI development.
  \item \textbf{Inference}: SLMs are small enough to run on edge devices. An 8B-parameter model running on a Snapdragon X Elite laptop processor answers a 64-token prompt in roughly 0.4 s at $\sim$7 W~\citep{offline_ai_guide}. This eliminates the round-trip energy cost to a data center.
  \item \textbf{The edge advantage}: Such a short query processed locally consumes only a few Joules ($\sim$7 W $\times$ 0.4 s $\approx$ 3 J, i.e., around 0.001 Wh) and no data-center cooling water. Long reasoning queries in the cloud sit at the other end of the scale: One benchmark estimates that a long query (10,000 input and 1,500 output tokens) sent to DeepSeek-R1 on DeepSeek's infrastructure consumes $\sim$29 Wh and more than 200 mL of water~\citep{ai_hungry_benchmark_2025}. The two workloads differ too much in size for these figures to give a like-for-like ratio. For billions of daily transactions, DSS is the only environmentally viable path for widespread AI adoption.
\end{itemize}

\paragraph{Agentic workflows and the Society of Mind}
The DSS vision moves toward the AGI goal not as a single monolith, but as a ``society'' of specialized agents:
\begin{itemize}
  \item \textbf{Modular intelligence}: A ``general'' AI agent (likely a cloud-based router) acts as a conductor, breaking complex queries into sub-tasks and dispatching them to specialized DSS models. A medical diagnosis query might be routed to a ``radiology expert'' and a ``pathology expert,'' whose outputs are synthesized.
  \item \textbf{Hierarchical efficiency}: This enables the system to use the ``least capable model'' necessary for each sub-task. Routine language tasks are handled by inexpensive, low-energy SLMs whereas heavy reasoning is reserved for specific DSS models. By routing a large fraction of queries to efficient SLMs and only a much smaller fraction to heavy reasoning models, the system maintains high capability while reducing aggregate energy use significantly.
\end{itemize}

\subsection{Strategic outlook}

The current trajectory---centralized, massive-scale inference on general-purpose hardware---is economically and environmentally unsustainable. The industry is approaching physical limits that software efficiency alone cannot resolve:

\begin{enumerate}
  \item \textbf{The energy wall}: The electricity demand of frontier training runs, such as OpenAI's Orion, is a harbinger~\citep{ier_orion_energy}. Next-generation models may require dedicated nuclear reactors or massive off-grid renewable arrays. Stargate's  10 GW target acknowledges this reality.
  \item \textbf{Inference dominance}: Inference already accounts for most AI computing~\citep{ai_inference_costs_2025}. Its share is likely to grow as reasoning models spread. How much energy a reasoning model consumes per query depends on the setup: For one benchmark's long-prompt workload (10,000 input and 1,500 output tokens), the estimated energy per query is $\sim$29 Wh for DeepSeek-R1 served by DeepSeek, $\sim$7 Wh for the same model served on Azure, and $\sim$3 Wh for GPT-4o~\citep{ai_hungry_benchmark_2025}.
  \item \textbf{Cost fracture}: The market will split between ``commodity intelligence'' (DeepSeek-style distilled models on edge
  hardware) and ``premium reasoning'' (Orion/Stargate-class, centralized, expensive, energy-intensive).
\end{enumerate}

Over the next decade, competitive advantage will depend as much on hardware efficiency and energy access as on algorithmic innovation.
By shifting from top-down monoliths to bottom-up, modular societies of models, we can reconcile the demand for superintelligence with the constraints of our planet's resources. Energy efficiency is, therefore, not an auxiliary engineering consideration---it is a defining criterion for the next generation of AI systems.

%% file: sections/11_Goal.tex
\section{Goal of DSS: Enhancing the productivity of every worker}
\label{sec:final_goals}

This section grounds the DSS trajectory in its purpose. The earlier sections built the technical machinery: abstractions, neurosymbolic reasoning, SLMs, KGs, agentic loops, and edge-class energy budgets. A technical proposal alone, however, does not justify a new direction for the field. Here, we argue that the design target of DSS is task-relevant sufficiency, not maximal generality. A model that reaches expert-level performance in a bounded domain under strict energy and deployment budgets is preferable to a generalist with marginally broader recall at orders of magnitude greater cost. The \textit{Pocket Sage}, GraphRAG-grounded reasoning, proofs of reasoning, and KG-grounded synthetic curricula are the concrete forms DSS takes once this objective is accepted. 

A DSS society is a socioeconomic object, not only a technical one. This section describes the institutional scaffolding required for it to function: an orchestrator with expert back-ends as a productivity multiplier for every worker, an \textit{Internet of Agents} that prevents vendor lock-in, data coalitions and data dignity as the basis of an equitable data economy, sovereign and indigenous DSS as the alternative to compute-divide colonialism, and human-in-the-loop governance as the guardrail on irreversible action. The new occupational tiers of data curators, synthetic data engineers, and ontology architects describe how a DSS society would create meaningful work rather than destroy it. Educational reform centered on abstraction and KG-grounded tutoring shows how the next generation will be prepared to inhabit such a society. 

\subsection{From generality to sufficiency}

The prevailing narrative assumes that increasingly general models will eventually subsume all domain expertise. Yet, generality carries intrinsic costs: higher inference energy, larger memory footprints, longer latency, and centralized infrastructure dependency.
For most real-world applications, maximal generality is neither required nor economically optimal. The objective should be \textit{task-relevant sufficiency}: achieving or exceeding expert-level performance within a constrained domain under strict energy and deployment budgets. Rather than maximizing capability in isolation, we propose optimizing:

\begin{equation}
M^* = \arg\max_{M \in \mathcal{M}}
\frac{P_{\text{domain}}(M)}
{E(M) + \lambda\, C(M)},
\end{equation}

\noindent
where $P_{\text{domain}}$ denotes model performance within a defined task space, $E(M)$ is lifecycle energy consumption, $C(M)$ is economic cost, and $\lambda$ encodes deployment constraints. Under this objective, larger models are not inherently superior. A model that achieves sufficient domain expertise with orders-of-magnitude lower energy footprint is preferable to a generalist model with marginally higher cross-domain ability.

\subsection{Domain-specific superintelligence: The ``Society of Mind'' architecture}

DSS differs from traditional AGI aspirations in three respects:

\begin{itemize}
    \item \textbf{Constraint-aware optimization:} Intelligence is evaluated under energy, latency, and cost budgets, not in isolation.
    \item \textbf{Architectural modularity:} Specialized models replace monolithic universal networks, activating only the expertise required for each task.
    \item \textbf{Deployment locality:} Whenever feasible, inference occurs near the point of use, reducing infrastructure load and latency.
\end{itemize}

This paradigm does not reject general intelligence research; it recognizes that sustainable large-scale deployment favors specialization over universalization.

\subsubsection{The orchestrator and the experts}

In the DSS architecture, intelligence is modular. A lightweight \textit{orchestrator agent} (typically a fine-tuned SLM) is the executive control center. It does not attempt to solve every problem itself; instead, it parses complex user intent, decomposes it into sub-tasks, and routes these tasks to specialized \textit{expert agents}~\citep{ier_orion_energy}.

For example, a complex query regarding a supply chain disruption might be decomposed into:

\begin{enumerate}
    \item \textbf{Weather expert:} An SLM connected to meteorological APIs to forecast storm impacts.
    \item \textbf{Logistics expert:} An agent grounded in the company's Enterprise Resource Planning database and a logistics ontology to identify affected shipments.
    \item \textbf{Legal expert:} A neurosymbolic agent trained on contract law to determine liability for delayed deliveries.
\end{enumerate}

The orchestrator then synthesizes the outputs of these experts into a coherent response. This architecture has several advantages:

\begin{itemize}
    \item \textbf{Asymptotic competence:} The system's intelligence grows by adding new specialized modules (e.g., a ``customs compliance expert'') rather than retraining a trillion-parameter giant~\citep{agents_multiagent_collab}.
    \item \textbf{Verifiable specialization:} Each expert agent is constrained by a curated corpus and a rigorous domain ontology. The legal expert does not ``hallucinate'' case law; it retrieves verifiable precedents from a legal KG~\citep{dss_legal_ai}.
    \item \textbf{Hierarchical efficiency:} Routine tasks are handled by small, efficient models, whereas energy-intensive reasoning is reserved for complex integration. This reduces the aggregate energy cost per task by orders of magnitude compared to using a frontier model for everything~\citep{ier_orion_energy}.
\end{itemize}

\subsubsection{Interoperability and the Internet of agents}

For a \textit{Society of Mind} to function, agents must share a common language for negotiation and data exchange. Standardized protocols are emerging that act as the ``USB-C for AI.'' The MCP and A2A standards enable diverse agents---built by different vendors, running on different clouds or devices---to discover each other, advertise their capabilities, and collaborate on tasks~\citep{dss_agent_protocols}.

These protocols enable an \textit{Internet of Agents} where a healthcare agent from a hospital can securely query an insurance agent from a payer and a scheduling agent from a patient's phone, orchestrating a complex medical appointment without a single central platform owning the entire interaction~\citep{dss_internet_of_agents}. This decentralization supports a competitive ecosystem that prevents vendor lock-in and promotes ``antifragility'': if one model fails, the network can route around it.

\subsection{The Pocket Sage: Smartphone-resident intelligence}
\label{sec:pocket_sage}

A natural endpoint of DSS is the \textit{Pocket Sage}: a domain-specific expert that resides entirely on a user's smartphone, requiring no cloud connectivity, no subscription fees, and no data exfiltration. 

\subsubsection{From cloud dependency to edge sovereignty}

Modern smartphone processors---such as Apple's A18 Pro and Qualcomm's Snapdragon 8 Elite---now include dedicated NPUs (e.g., the A18 Pro's 16-core Neural Engine is rated at 35 INT8 TOPS)~\citep{apple_a18_pro,snapdragon_8_elite}, and 4-bit-quantized 8B-parameter models, such as Llama-3.1-8B-Instruct, are already deployable on Snapdragon 8 Elite smartphones~\citep{qualcomm_llama}. Combined with the quantization and distillation techniques, a high-quality DSS can be compressed to fit within the memory constraints of a typical smartphone (8-16 GB RAM) while retaining 95\%+ of its reasoning capability.
This decouples intelligence from connectivity. Farmers in rural Kenya do not need a 5G connection to a distant data center; a \textit{Maize Agronomy DSS} trained on local soil types, pest patterns, and weather cycles can run entirely on their device. Factory technicians in a subterranean facility---where connectivity is impossible---can consult a \textit{Hydraulic System DSS} trained on equipment manuals and failure modes. This eliminates the ``capability tax'' imposed by cloud subscriptions and bandwidth requirements.

\subsubsection{The energy arithmetic of edge intelligence}

The energy economics strongly favor edge deployment. As discussed in Section~\ref{sec:energy}, a long query (10,000 input and 1,500 output tokens) to a cloud-hosted reasoning model (DeepSeek-R1, served by DeepSeek) is estimated to consume approximately 29 Wh~\citep{ai_hungry_benchmark_2025}: more than enough to fully charge a smartphone. Separately, an 8B-parameter model running locally answers a short (64-token) prompt in roughly 0.4 s at 7 W~\citep{offline_ai_guide}, i.e., under 0.001 Wh. These two workloads are not matched. Hence, the figures do not measure the size of the saving: They show how little energy a short on-device query needs. At this rate, the energy required for one full smartphone charge can cover thousands of such short prompts.

Edge deployment also eliminates the water consumption associated with data center cooling, carbon emissions from long-haul data transmission, and privacy risks inherent in sending sensitive queries to third-party servers. By design, \textit{Pocket Sage} is sustainable, private, and under user control.

\subsection{The neurosymbolic bridge: Restoring reliability with knowledge graphs}

The technical foundation of DSS is neurosymbolic AI (as discussed in Section \ref{sec:ns_abstractions}): the fusion of linguistic fluency of neural networks (``System 1'') with the logical rigor of symbolic systems (``System 2''). 

\subsubsection{GraphRAG: Beyond the vector}

GraphRAG improves on standard vector-based RAG. Vector databases retrieve isolated chunks of text based on fuzzy semantic similarity; GraphRAG instead retrieves \textit{structured relationships} from a KG~\citep{dss_graphrag_comparison}.
In the DSS context, a KG is the ``ground truth.'' It explicitly models entities (e.g., ``protein,'' ``drug,'' ``statute'') and their relationships (e.g., ``inhibits,'' ``is\_precedent\_for''). When an expert agent receives a query, it does not merely scan for keywords; It traverses the graph to find causal chains. For instance, in a pharmaceutical context, GraphRAG enables multi-hop reasoning: \textit{``Drug A targets Protein B. Protein B is upregulated in Disease C. Therefore, Drug A may treat Disease C''}~\citep{dss_graphrag_pharma}. Benchmarks indicate that GraphRAG significantly outperforms vector methods in complex reasoning tasks, particularly where schemas and precise constraints are involved~\citep{dss_graphrag_benchmark}.

\subsubsection{Proof of reasoning and audit trails}

In high-stakes domains, ``black box'' decisions are legally and ethically unacceptable. A medical DSS cannot simply output a diagnosis; it must provide a ``proof of reasoning'': an immutable, cryptographic log of the logical steps taken to reach its conclusion~\citep{dss_proof_of_reasoning}.

This proof might look like:

\begin{enumerate}
    \item Observation: Patient exhibits symptoms X, Y, Z.
    \item Retrieval: KG query identifies three potential conditions that match symptoms.
    \item Exclusion: Condition A is excluded due to Lab Result W (based on rule \#421 in Medical Ontology).
    \item Conclusion: Diagnosis is Condition B with 95\% confidence.
\end{enumerate}

This audit trail enables human professionals to verify AI logic, making the user an active auditor of reasoning rather than a passive consumer of outputs~\citep{ier_orion_energy}. This ``glass box'' approach is essential for regulatory compliance and for keeping professionals accountable in the decision loop.

\subsection{Augmenting productivity of every worker: DSS as a productivity multiplier}
\label{sec:worker_empowerment}

DSS is designed to amplify what workers can do rather than replace them, giving professionals access to expert-level guidance in their domain regardless of location or socioeconomic status. This applies to both white-collar knowledge workers and blue-collar skilled tradespeople.

\subsubsection{White-collar productivity augmentation: Cognitive offloading with accountability}

For knowledge workers, DSS is 
a second opinion that is always available, never fatigued, and grounded in verifiable knowledge.
For example, a ``clinical DSS'' resident on a physician's tablet can cross-reference patient symptoms against the latest diagnostic criteria, flag potential drug interactions, and retrieve relevant case studies from the medical literature. In rural clinics with limited specialist access, this provides a diagnostic capability approaching that of a tertiary research hospital. The DSS does not replace the physician's judgment; it supplements it with instant access to KG-encoded medical knowledge~\citep{dss_neurosymbolic_audit}.



\subsubsection{Blue-collar empowerment: Multimodal procedural guidance}

For skilled trades, DSS operates through \textit{multimodal grounding}: combining visual recognition, sensor data, and procedural manuals to provide real-time
guidance.

\paragraph{Factory technicians}
A ``maintenance DSS'' trained on equipment manuals, failure mode databases, and repair histories can diagnose machine errors via camera input and provide step-by-step repair instructions overlaid on an augmented reality (AR) display. A technician facing an unfamiliar error code no longer needs to wait for a specialist; the DSS provides immediate, verified guidance. The target metric is the reduction of mean-time-to-repair and the elimination of catastrophic errors caused by procedural mistakes~\citep{dss_augmented_worker}.

\paragraph{Agricultural workers}
A ``precision agriculture DSS'' running on a farmer's smartphone can analyze soil samples (via camera-based spectroscopy apps), cross-reference local weather forecasts, and recommend optimal planting schedules, irrigation timing, and pest management strategies. This DSS is trained not on generic agricultural data but on a KG of local conditions: specific crop varieties suited to the region, endemic pest species, and traditional farming practices that have proven effective over generations~\citep{dss_agriculture_ai,offline_ai_guide}.

\paragraph{Construction workers}
A ``building codes DSS'' can verify that construction practices comply with local regulations, flagging potential violations before they become costly rework. By integrating with building information models, it can provide spatial guidance for complex installations.

\paragraph{Electricians and plumbers}
Specialized ``trade DSS'' agents trained on wiring diagrams, plumbing schematics, and safety codes can guide tradespeople through unfamiliar installations, reducing 
code violations and safety hazards.

\subsubsection{The human-in-the-loop guardrail}

DSS systems are designed as \textit{recommenders}, not \textit{deciders}. For irreversible physical actions---cutting a wire, administering a medication, filing a legal motion---the system enforces mandatory ``human-in-the-loop'' verification steps. The worker remains the active agent; the DSS provides information but the human bears final responsibility. This design prevents ``automation bias'' and keeps professional judgment at the center of decision-making~\citep{dss_hitl_future}.

\subsection{Education: Restoring abstraction as the pedagogical center}
\label{sec:education}

A direct application of DSS is in education. As argued in Section~\ref{sec:intro}, intelligence begins by forming structured mental models (abstractions). By grounding AI tutors in domain-specific KGs, we restore abstraction as the prerequisite for learning, creating personalized experiences that combine the rigor of structured curricula with the adaptability of one-on-one instruction.

\subsubsection{Every textbook as a knowledge graph}

In the DSS paradigm, every textbook, manual, and corpus of domain knowledge is accompanied by an explicit KG that encodes its conceptual structure. This graph represents facts and, more importantly, \textit{relationships} between concepts: prerequisites, implications, analogies, and common misconceptions.

For example, a KG for introductory physics might encode:
\begin{itemize}
    \item Concept: Newton's Second Law ($F = ma$)
    \item Prerequisites: concepts of force, mass, and acceleration
    \item Implications: conservation of momentum, projectile motion, orbital mechanics
    \item Common misconceptions: confusing mass with weight; assuming $F = ma$ applies only to constant forces
    \item Abstraction level: intermediate (requires algebra, no calculus)
\end{itemize}

This structured representation enables the AI tutor to diagnose gaps in understanding by traversing the prerequisite graph. If a student struggles with projectile motion, the tutor can identify whether the root cause is a misunderstanding of Newton's Second Law, a gap in vector algebra, or a misconception about gravitational acceleration.

\subsubsection{Unlimited high-quality question-answer-reasoning trace generation}

A practical advantage of KG-grounded education is the ability to generate essentially unlimited high-quality question-answer-reasoning trace triplets. Unlike static question banks, a DSS tutor can dynamically construct novel problems by:

\begin{enumerate}
    \item \textit{Sampling} a target concept from the KG.
    \item \textit{Instantiating} the concept with novel parameters (e.g., different masses, velocities, contexts).
    \item \textit{Generating} a question that tests understanding of the concept.
    \item \textit{Producing} a step-by-step reasoning chain that derives the answer from first principles.
    \item \textit{Creating} plausible distractors based on common misconceptions encoded in the KG.
\end{enumerate}

This capability is useful to both test-givers (educators, certification bodies) and test-takers (students). Educators gain access to an inexhaustible supply of valid, calibrated assessment items. Students gain access to unlimited practice problems, each accompanied by a detailed explanation: not a canned solution but a reasoning trace that teaches the underlying logic and rules of abstraction.

\subsubsection{Teaching abstractions: Key to deep learning}

The DSS approach emphasizes abstraction as the core skill to be cultivated. Rather than memorizing facts or procedures, students learn to recognize patterns, generalize from examples, and apply principles to novel situations. The KG explicitly encodes abstraction hierarchies: 
how specific examples instantiate general principles and how general principles are derived from 
fundamental axioms.
This is DL in the pedagogical sense: understanding that $F = ma$ is not just an equation to be memorized but an instance of the broader principle that changes in motion require external influence. A DSS tutor trained on such a KG can guide students up and down the abstraction ladder, ensuring that procedural fluency is always grounded in conceptual understanding~\citep{turing_lecture_abstractions}.

\subsubsection{The AI tutor as a world expert}

The long-term goal of DSS 
is broad access to high-quality instruction. Today, the quality of education a student receives depends heavily on geography and socioeconomic status. A student in a well-funded suburban school may have access to expert teachers; a student in an under-resourced rural school may not.
DSS can narrow this gap: A ``mathematics DSS'' trained by leading mathematicians and grounded in a rigorous mathematical ontology can provide Socratic tutoring to any student with a smartphone. This does not replace human teachers; it frees them. Educators can shift from routine drill-and-practice to mentorship and creativity, where human judgment matters most.

\subsection{Breaking the data wall: Synthetic curricula and ``textbooks''}

To overcome the data wall and the threat of model collapse, DSS adopts a new paradigm for model training: \textit{Synthetic Curricula}. Rather than relying on the noisy, unstructured web, it uses KGs and high-performing teacher models to 
generate vast amounts of high-signal training data~\citep{ier_orion_energy}.

\subsubsection{The ``textbooks are all you need'' paradigm}

The efficacy of this approach was demonstrated by Microsoft's Phi~\citep{abdin2024phi} series of models. By training on ``textbook-quality'' synthetic data---carefully curated to teach reasoning principles rather than just rote facts---the 1.3B-parameter Phi-1 outperformed models 100$\times$ its size on coding benchmarks~\citep{dss_textbooks_all_you_need}.

DSS extends this ``Textbooks Are All You Need'' philosophy to every domain. An ``organic chemistry DSS'' would be trained not on scraped PDFs but on a synthetic curriculum generated from a rigorous chemistry ontology. This curriculum would systematically cover every reaction type, molecular structure, and bonding rule, presenting them in diverse contexts and difficulty levels~\citep{dss_synthllm}. This ``high-signal'' training enables SLMs to reach expert-level depth in their specific niche without the massive parameter count required to ``memorize the Internet.''

This method also mitigates model collapse. Because the synthetic data are generated from a structured, verifiable ontology, they  
preserve variance and factual integrity~\citep{ma2025synthesize}. The KG ensures that the model trains on the logic of the domain, not just statistical artifacts of previous model outputs.

\subsubsection{Connecting to training efficiency}

The DSS approach 
reduces training costs through three mechanisms:

\begin{enumerate}
    \item \textbf{Smaller models:} By focusing on narrow domains, DSS requires models with 1-13B parameters rather than hundreds of billions, reducing training compute by 10-100$\times$.
    \item \textbf{Higher-quality data:} Synthetic curricula generated from KGs provide higher information density than web scrapes, meaning fewer training tokens are needed to achieve equivalent capability.
    \item \textbf{Transfer learning:} Domain-specific models can be fine-tuned from efficient base models (e.g., Phi, Llama), rather than trained from scratch, further reducing energy expenditure.
\end{enumerate}

The result is that training a world-class DSS for a specific domain requires a fraction of the energy, water, and capital needed to train a frontier generalist model. This opens AI development to a wider range of institutions: Universities, professional associations, and individual enterprises can create expert systems tailored to their needs.

\subsection{Economic restructuring: From top-down dystopia to bottom-up opportunity}
\label{sec:economic}

A dominant narrative in public discourse frames AI as an engine of mass unemployment: a technology that will automate white-collar jobs. This ``top-down dystopia'' envisions a future where a small techno-elite controls omniscient AGI systems while rendering a majority of workers economically superfluous.
We argue that this framing is a consequence of the scaling paradigm. The bottom-up trajectory of DSS offers a different path, one that favors productivity augmentation over replacement, job creation alongside displacement, and distributed over concentrated economic gains.

\subsubsection{Historical precedent: Productivity as the engine of progress}

Throughout modern history, increases in productivity have been the primary driver of improvements in living standards. The DSS framework is designed to continue this pattern by raising the productivity floor across all occupations, rather than concentrating gains among a small elite. When every worker is empowered by a domain-specific expert on the phone, the aggregate wealth of society increases, creating demand for new goods and services.

\subsubsection{New labor market: Roles in the DSS ecosystem}

The deployment of DSS networks creates demand for new categories of labor: jobs that draw on human expertise in ways that scale-driven AI cannot replicate. We identify three tiers of emerging roles:

\paragraph{Entry level: Data curators (high school graduates)}
The DSS ecosystem requires vast quantities of high-quality, domain-specific data. Unlike the ``clickwork'' of simple image labeling, \textit{data curation} for DSS demands contextual awareness and domain familiarity. These roles provide meaningful employment for workers without advanced degrees, with wages significantly above minimum wage due to the specialized knowledge required~\citep{dss_data_labeling}.

\paragraph{Intermediate level: Synthetic data engineers (college graduates)}
Creating the synthetic curricula that train DSS requires a new profession: the ``synthetic data engineer.'' These professionals design the templates, verification protocols, and quality metrics that govern curriculum generation. They work at the intersection of domain expertise and machine learning, translating expert knowledge into structured formats.

\paragraph{Advanced level: Ontology architects (Ph.D students and researchers)}
At the apex of the DSS labor market are ``ontology architects'': the experts who define the conceptual structure of KGs. These are domain experts---physicians, lawyers, scientists---who encode their tacit knowledge into explicit representations. This is demanding work that commands high compensation, as it directly determines the quality and reliability of the resulting AI system~\citep{dss_expert_data,dss_train_ai_side_hustle}.

\subsubsection{Data dignity and the data-centric vision}

The shift to DSS requires a fundamental restructuring of the data economy. 
In the DSS economy, structured, verifiable expertise is more valuable than raw compute. Companies like Surge AI are already paying PhD-level experts hundreds of dollars per hour for specialized data curation. This establishes a reasoning wage premium, shifting economic value back to the human creators of knowledge who provide high-quality curation that models need to generalize.

\subsubsection{Data coalitions and collective bargaining}

To operationalize data dignity, we envision the rise of data coalitions: collective bargaining entities that pool the data of professionals and negotiate licensing terms with AI developers~\citep{dss_data_coalitions}. A radiologist data coalition might license a high-quality dataset of annotated scans, so that the doctors receive a royalty stream. This creates a self-reinforcing cycle: Experts are incentivized to maintain the high-quality KGs that power the DSS, which in turn preserves the long-term 
integrity of the system.

\subsection{Global equity: Sovereign AI and the compute divide}

The monolithic AI paradigm exacerbates inequality. The immense cost of training frontier models (\$100M+) creates a ``compute divide,'' restricting AI sovereignty to a handful of wealthy nations and corporations~\citep{dss_compute_divide}. Developing nations face the risk of becoming ``digital colonies,'' dependent on foreign ``black box'' models that do not understand their languages, legal codes, or cultural contexts~\citep{dss_digital_divide}.

\subsubsection{Epistemic sovereignty via edge AI}

DSS provides a route to epistemic sovereignty. Because specialized SLMs are small and efficient, they can be trained and deployed on local infrastructure, or even on consumer-grade edge devices~\citep{ier_orion_energy}. This enables nations to build their own ``Sovereign AI'' ecosystems tailored to their specific needs without relying on hyperscale data centers~\citep{dss_sovereign_ai}.

Consider a ``hyper-local agriculture DSS'' for a farmer in rural Kenya. Instead of querying a US-based cloud model trained on Midwest industrial farming data, the farmer uses an SLM running locally on a smartphone. This model is trained on a curated KG of local soil types, crop varieties, and weather patterns~\citep{offline_ai_guide}. It functions offline, which preserves privacy and eliminates latency. This decouples intelligence from connectivity and capital, making expert-level knowledge accessible without cloud infrastructure.

\subsubsection{Cultural preservation and indigenous AI}

SLMs are well-suited for cultural preservation. Indigenous communities can train small, high-quality models on their own languages and oral histories, creating digital archives that revitalize endangered cultures~\citep{dss_indigenous_languages}. Because these models are lightweight, they can be owned and governed by the community itself, preventing the extraction and commodification of their cultural heritage by external actors. This aligns AI with the principles of self-determination, so that the ``digital future'' includes the voices of the oldest cultures on earth.

\subsection{Autonomous discovery and self-driving laboratories}

DSS may have its largest impact on the pace of scientific discovery. The complexity of modern disciplines---genomics, materials science, high-energy physics---has surpassed the capacity of unaided human cognition. DSS agents, integrated into self-driving laboratories, open a path toward autonomous discovery.

The A-Lab at Berkeley Lab is a concrete example. This autonomous materials science laboratory combines AI reasoning with robotic synthesis to discover new materials faster than manual methods permit. In a 17-day run, the A-Lab synthesized 36 target compounds, operating 24/7 with minimal human intervention~\citep{dss_deepmind_materials, Szymanski2026}.

The ``brain'' of the A-Lab is a DSS composed of specialized agents: one to scour the literature and generate hypotheses (hypothesis agent), one to plan the synthesis recipe (experiment agent), and one to control the robotic arms and analyze the results (execution agent)~\citep{dss_autonomous_agents}. With KGs, these agents can reason over the entirety of published chemical literature, finding connections and ``dark data'' that no single human can synthesize.

This approach could accelerate progress in clean energy materials, drug discovery, and climate mitigation while maintaining epistemic integrity. The DSS does not just guess a new material; it produces a verifiable synthesis recipe and an audit trail of its reasoning, enabling human scientists to validate and build upon the machine's work~\citep{dss_knowledge_graphs_clarivate}.

\subsection{Governance and the digital social contract}

The transition to a society of autonomous, reasoning agents requires a new digital social contract. As AI moves from a passive tool to an active agent in the economy and society, issues of liability, bias, and control become central.

\subsubsection{Participatory governance and human-in-the-loop}

We advocate for participatory governance models where diverse stakeholders---including workers, marginalized communities, and domain experts---are actively involved in the design and oversight of AI systems~\citep{ier_orion_energy}. For high-stakes decisions, the ``human-in-the-loop''  principle must be strictly enforced. The AI acts as a recommender, providing verifiable evidence and reasoning, but the final decision---and the moral liability---rests with a human~\citep{ier_orion_energy}.
This approach transforms the workforce rather than replace it. It creates an ``augmented worker'' paradigm where human judgment is amplified by machine precision. 

\subsubsection{Comparative analysis of AI paradigms}

Table~\ref{tab:paradigm_comparison} summarizes the fundamental shift from the current monolithic paradigm to the proposed DSS framework.

\begin{table}[htbp]
\centering
\caption{Comparative analysis of AI paradigms}
\label{tab:paradigm_comparison}
\resizebox{\textwidth}{!}{%
\begin{tabular}{@{}p{3cm}p{5.5cm}p{5.5cm}@{}}
\toprule
\textbf{Feature} & \textbf{Monolithic AGI Paradigm (Current Trend)} & \textbf{Domain-Specific Superintelligence (Proposed)} \\
\midrule
Primary Architecture & Single, massive ``foundation model'' (trillions of parameters) & Ecosystem of modular ``expert agents'' (SLMs + KGs) \\
Scaling Mechanism & Brute-force scaling (parameters \& data) & Neurosymbolic scaling (logic \& structure) \\
Energy Profile & Centralized, GW-scale (nuclear/fossil) & Decentralized, W-scale per device (edge/battery) \\
Data Strategy & Indiscriminate scraping of the open web & Curated ``textbooks'' \& KGs \\
Reasoning Type & Probabilistic, opaque (``System 1'') & Verifiable, transparent (``System 2'') \\
Economic Model & Subscription/API rent-seeking & Data dignity \& sovereign ownership \\
Key Risk & Hallucination, model collapse, energy wall & Integration complexity, coordination overhead \\
Accessibility & Centralized control (high barrier) & Broad access (low barrier) \\
\bottomrule
\end{tabular}%
}
\end{table}

%% file: sections/1_AGI.tex
\section{AGI as a vision: Is it worth pursuing now?}
\label{sec:AGI}

Although this article argues for DSS rather than scale-driven AGI, the AGI question remains important because it clarifies what kind of intelligence the proposed trajectory is meant to approximate. From this perspective, AGI serves as a useful reference point: It defines the breadth of capability that the field has traditionally sought, whereas DSS specifies an alternative path toward that breadth through modular specialization and orchestration.

If a DSS society can cover enough domains, resolve conflicts among specialists, and transfer structured information across them, then it may exhibit many of the useful properties associated with AGI without requiring a single omniscient model. Thus, this section explains why DSS societies may be a more attainable and sustainable way to approach the practical benefits that AGI is expected to deliver.

\subsection{What is AGI and why does the field care about it so much?}
\label{subsec:what_is_agi}

Since its inception, the field of AI has aspired to create machines that possess intelligence equal to or greater than that of humans. Most scholarly and historical accounts credit Ben Goertzel with popularizing the term AGI around 2002, contrasting it with ``narrow AI.''  
One of the definitions of intelligence is the capacity to handle uncertainty and acquire skills to complete tasks for which one was not explicitly trained~\citep{chollet2019measureintelligence}. AGI is envisioned as satisfying this requirement. Thus, the expectation is that AGI should demonstrate human-like cognition across a wide range of unformatted tasks and acquire human-like abilities to learn, apply, and transfer knowledge. Though there is little agreement on a precise definition of AGI, there is consensus that it has not yet been achieved. The timeline for its inception and reaching the so-called ``singularity'' continues to be postponed each year. 

\subsection{Why AGI is unlikely to emerge from LLM scaling alone}
\label{subsec:agi_is_not_feasible_with_llms}

The path to AGI shaped during the contemporary LLM era depends on scale, data, and emergent capabilities to reach human-parity performance. However, humans can learn from 
a few examples and readily generalize their learning to novel situations. In this regard, LLMs are still far behind human-like intelligence. One argument is that the human brain is better at forming and maintaining compact abstract representations, whereas LLMs primarily encode knowledge as a vast collection of context-weighted ``special cases'' that support pattern reuse rather than abstraction as the default mode~\citep{goertzel2023generativeaivsagi}.
Recent discussions of AGI and brain-inspired AI emphasize that progress towards AGI is now constrained less by abstract definitions than by concrete bottlenecks in learning and deployment~\citep{brain-inspired_AGI}. We summarize some of these discussions here. 
\begin{itemize}
    \item {\bf Data efficiency:} Learning from limited data remains an open challenge in the march towards AGI. Currently, LLMs require vast amounts of data to achieve human-level performance on individual tasks.

    \item {\bf Computational cost:} LLMs require substantial computational resources for both training and deployment, limiting their applicability. They require the construction and maintenance of data centers and other infrastructure, which may have negative externalities.

    \item {\bf Environmental perspective:} The substantial energy consumption for LLM inference raises serious concerns about their long-term environmental sustainability.
\end{itemize}

\subsection{Switching aspirations from AGI to DSS as an alternative}
\label{subsec:switching_to_dss}
For decades, AI was constrained by computing and data availability. Scaling in the LLM era on commercial-scale resources enabled various breakthroughs. As data/compute limits come into sharper focus, we enter the era of diminishing returns~\citep{Bennett_2025}. At the same time, the financial and environmental costs of training and deploying ever-larger models have become increasingly salient~\citep{strubell2019energy}, making specialization attractive even if AGI remains a long-term goal. Now, the field needs to move toward a new paradigm that embraces efficiency and sustainability.

\paragraph{Why DSS is a good alternative}
We encounter far more data-scarce problems than those with abundant training examples, and scaling no longer seems to be the most straightforward path to further progress~\citep{Bennett_2025}. In such settings, parameter-efficient adaptation of smaller language models on domain-specific data can be considerably more energy-efficient, and the required data are often available.

%% file: sections/Conclusion.tex
\section{Conclusion}

DSS is a shift from black-box scaling to structured, verifiable reasoning. It takes the physical limits of energy and water, the finite supply of high-quality training data, and the economic value of human expertise seriously. By distributing intelligence across specialized agents rather than concentrating it in monolithic models, DSS embeds expert-level reasoning into daily workflows.

The practical vision is concrete: a doctor in a remote village with diagnostic capability approaching that of a research hospital; a farmer with access to agronomic knowledge tailored to local conditions; a factory technician who can consult an expert system for any piece of equipment; a student working with an AI tutor that provides grounded explanations. GW-scale data centers like Stargate may serve frontier research, but the \textit{Pocket Sage}---an efficient, verifiable expert running on a phone---is the form factor that scales to billions of users.

DSS grounds intelligence in verifiable reasoning, makes it accessible without cloud connectivity, and aligns it with environmental sustainability and human expertise.

The history of computing repeatedly demonstrates that monolithic systems give way to modular, optimized architectures once deployment scale increases. AI appears to be approaching a similar transition point. Scaling laws enabled rapid capability gains during the pre-deployment phase, but as systems become infrastructural, energy efficiency, economic sustainability, and deployment constraints reshape the optimization landscape. The next phase of AI development will be defined not by the largest models but by the most efficient ones: the disciplined allocation of computational resources toward bounded, high-value tasks under real-world constraints.

%% file: reference.bib
@article{jakubik2024data,
  title={Data-centric artificial intelligence},
  author={Jakubik, Johannes and V{\"o}ssing, Michael and K{\"u}hl, Niklas and Walk, Jannis and Satzger, Gerhard},
  journal={Business \& Information Systems Engineering},
  volume={66},
  number={4},
  pages={507--515},
  year={2024},
  publisher={Springer}
}

@article{haller2025llm,
  title={{LLM knowledge is brittle: Truthfulness representations rely on superficial resemblance}},
  author={Haller, Patrick and Ibrahim, Mark and Kirichenko, Polina and Sagun, Levent and Bell, Samuel J.},
  journal={arXiv preprint arXiv:2510.11905},
  year={2025}
}

@article{su2025single,
  title={A single character can make or break your {LLM} evals},
  author={Su, Jingtong and Zhang, Jianyu and Ullrich, Karen and Bottou, L{\'e}on and Ibrahim, Mark},
  journal={arXiv preprint arXiv:2510.05152},
  year={2025}
}

@article{mohsin2025fundamental,
  title={On the fundamental limits of {LLM}s at scale},
  author={Mohsin, Muhammad Ahmed and Umer, Muhammad and Bilal, Ahsan and Memon, Zeeshan and Qadir, Muhammad Ibtsaam and Bhattacharya, Sagnik and Rizwan, Hassan and Gorle, Abhiram R. and Kazmi, Maahe Zehra and Mohsin, Ayesha and others},
  journal={arXiv preprint arXiv:2511.12869},
  year={2025}
}

@article{halevy2009unreasonable,
  title={The unreasonable effectiveness of data},
  author={Halevy, Alon and Norvig, Peter and Pereira, Fernando},
  journal={IEEE Intelligent Systems},
  volume={24},
  number={2},
  pages={8--12},
  year={2009},
  publisher={IEEE}
}

@article{kansal2026knowledge,
  title={{Knowledge graphs are implicit reward models: Path-derived signals enable compositional reasoning}},
  author={Kansal, Yuval and Jha, Niraj K.},
  journal={arXiv preprint arXiv:2601.15160},
  year={2026}
}

@article{zhao2024exploring,
  title={Exploring the limitations of large language models in compositional relation reasoning},
  author={Zhao, Jinman and Zhang, Xueyan},
  journal={arXiv preprint arXiv:2403.02615},
  year={2024}
}

@article{jumper2021highly,
  title={Highly accurate protein structure prediction with {AlphaFold}},
  author={Jumper, John and Evans, Richard and Pritzel, Alexander and Green, Tim and Figurnov, Michael and Ronneberger, Olaf and Tunyasuvunakool, Kathryn and Bates, Russ and {\v{Z}}{\'\i}dek, Augustin and Potapenko, Anna and others},
  journal={Nature},
  volume={596},
  number={7873},
  pages={583--589},
  year={2021},
  publisher={Nature Publishing Group UK London}
}

@book{hutchins1995cognition,
  title={{Cognition In The Wild}},
  author={Hutchins, Edwin},
  year={1995},
  publisher={MIT Press}
}

@inproceedings{ellis2021dreamcoder,
  title={{DreamCoder}: Bootstrapping inductive program synthesis with wake-sleep library learning},
  author={Ellis, Kevin and Wong, Catherine and Nye, Maxwell and Sabl{\'e}-Meyer, Mathias and Morales, Lucas and Hewitt, Luke and Cary, Luc and Solar-Lezama, Armando and Tenenbaum, Joshua B.},
  booktitle={Proceedings of the 42nd ACM SIGPLAN International Conference on Programming Language Design and Implementation},
  pages={835--850},
  year={2021}
}

@inproceedings{perelkiewicz2024review,
  title={A review of the challenges with massive web-mined corpora used in large language models pre-training},
  author={Pere{\l}kiewicz, Micha{\l} and Po{\'s}wiata, Rafa{\l}},
  booktitle={Proceedings of the International Conference on Artificial Intelligence and Soft Computing},
  pages={153--163},
  year={2025},
  organization={Springer}
}

@article{havrilla2024understanding,
  title={Understanding the effect of noise in {LLM} training data with algorithmic chains of thought},
  author={Havrilla, Alex and Iyer, Maia},
  journal={arXiv preprint arXiv:2402.04004},
  year={2024}
}

@article{dong2025scalable,
  title={Scalable vision language model training via high-quality data curation},
  author={Dong, Hongyuan and Kang, Zijian and Yin, Weijie and Liang, Xiao and Feng, Chao and Ran, Jiao},
  journal={arXiv preprint arXiv:2501.05952},
  year={2025}
}

@article{abdin2024phi,
  title={Phi-4 technical report},
  author={Abdin, Marah and Aneja, Jyoti and Behl, Harkirat and Bubeck, S{\'e}bastien and Eldan, Ronen and Gunasekar, Suriya and Harrison, Michael and Hewett, Russell J. and Javaheripi, Mojan and Kauffmann, Piero and others},
  journal={arXiv preprint arXiv:2412.08905},
  year={2024}
}

@article{dedhia2025bottom,
  title={{Bottom-up domain-specific superintelligence: A reliable knowledge graph is what we need}},
  author={Dedhia, Bhishma and Kansal, Yuval and Jha, Niraj K.},
  journal={arXiv preprint arXiv:2507.13966},
  year={2025}
}

@article{ma2025synthesize,
  title={{Synthesize-on-Graph: Knowledgeable synthetic data generation for continue pre-training of large language models}},
  author={Ma, Shengjie and Jiang, Xuhui and Xu, Chengjin and Yang, Cehao and Zhang, Liyu and Guo, Jian},
  journal={arXiv preprint arXiv:2505.00979},
  year={2025}
}

@inproceedings{yang2023leandojo,
  title={{LeanDojo}: Theorem proving with retrieval-augmented language models},
  author={Yang, Kaiyu and Swope, Aidan and Gu, Alex and Chalamala, Rahul and Song, Peiyang and Yu, Shixing and Godil, Saad and Prenger, Ryan J. and Anandkumar, Animashree},
  booktitle={Proceedings of the 37th International Conference on Neural Information Processing Systems},
  pages={21573--21612},
  year={2023}
}

@inproceedings{jiang2025logicpro,
  title={{LogicPro}: Improving complex logical reasoning via program-guided learning},
  author={Jiang, Jin and Yan, Yuchen and Liu, Yang and Wang, Jianing and Peng, Shuai and Cai, Xunliang and Cao, Yixin and Zhang, Mengdi and Gao, Liangcai},
  booktitle={Proceedings of the 63rd Annual Meeting of the Association for Computational Linguistics (Volume 1: Long Papers)},
  pages={26200--26218},
  year={2025}
}

@article{
brain-inspired_AGI,
title = {When brain-inspired {AI} meets {AGI}},
journal = {Meta-Radiology},
volume = {1},
number = {1},
pages = {100005},
year = {2023},

author = {Lin Zhao and Lu Zhang and Zihao Wu and Yuzhong Chen and Haixing Dai and Xiaowei Yu and Zhengliang Liu and Tuo Zhang and Xintao Hu and Xi Jiang and Xiang Li and Dajiang Zhu and Dinggang Shen and Tianming Liu},
}

@article{chollet2019measureintelligence,
  title={{On the measure of intelligence}},
  author={Chollet, Fran\c{c}ois},
  journal={arXiv preprint arXiv:1911.01547},
  year={2019}
}

@article{
Hubert2025OlympiadFormalMathRL,
  title   = {Olympiad-level formal mathematical reasoning with reinforcement learning},
  author  = {Hubert, Thomas and Mehta, Rishi and Sartran, Laurent and Horv{\'a}th, Mikl{\'o}s Z. and {\v Z}u{\v z}i{\'c}, Goran and Wieser, Eric and Huang, Aja and Schrittwieser, Julian and Schroecker, Yannick and Masoom, Hussain and Bertolli, Ottavia and Zahavy, Tom and Mandhane, Amol and Yung, Jessica and Beloshapka, Iuliya and Ibarz, Borja and Veeriah, Vivek and Yu, Lei and Nash, Oliver and Lezeau, Paul and Mercuri, Salvatore and S{\"o}nne, Calle and Mehta, Bhavik and Davies, Alex and Zheng, Daniel and Pedregosa, Fabian and Li, Yin and von Glehn, Ingrid and Rowland, Mark and Albanie, Samuel and Velingker, Ameya and Schmitt, Simon and Lockhart, Edward and Hughes, Edward and Michalewski, Henryk and Sonnerat, Nicolas and Hassabis, Demis and Kohli, Pushmeet and Silver, David},
  journal = {Nature},
  year    = {2025},

  note    = {Published online 12 Nov 2025}
}

@article{goertzel2023generativeaivsagi,
  title={{Generative AI vs. AGI: The cognitive strengths and weaknesses of modern LLMs}},
  author={Goertzel, Ben},
  journal={arXiv preprint arXiv:2309.10371},
  year={2023}
}

@article{richardson2006markov,
  author  = {Richardson, Matthew and Domingos, Pedro},
  title   = {Markov logic networks},
  journal = {Machine Learning},
  year    = {2006},
  volume  = {62},
  number  = {1},
  pages   = {107--136},
}

@book{fowler2010dsl,
  author    = {Fowler, Martin},
  title     = {Domain-Specific Languages},
  publisher = {Addison-Wesley Professional},
  year      = {2010},
}

@article{ABUSALIH2021103076,
title = {Domain-specific knowledge graphs: A survey},
journal = {Journal of Network and Computer Applications},
volume = {185},
pages = {103076},
year = {2021},
author = {Bilal Abu-Salih},
}

@article{ARMARY2025100693,
title = {Ontology learning towards expressiveness: A survey},
journal = {Computer Science Review},
volume = {56},
pages = {100693},
year = {2025},
author = {Pauline Armary and Cheikh Brahim El-Vaigh and Ouassila {Labbani Narsis} and Christophe Nicolle},
}

@article{vanderweele2010signed,
  author  = {VanderWeele, Tyler J. and Robins, James M.},
  title   = {Signed directed acyclic graphs for causal inference},
  journal = {Journal of the Royal Statistical Society: {Series B} (Statistical Methodology)},
  year    = {2010},
  volume  = {72},
  number  = {1},
  pages   = {111--127},
}

@article{xu2024contextgraph,
  title={{Context graph}},
  author={Xu, Chengjin and Li, Muzhi and Yang, Cehao and Jiang, Xuhui and Tang, Lumingyuan and Qi, Yiyan and Guo, Jian},
  journal={arXiv preprint arXiv:2406.11160},
  year={2024}
}

@article{
schulman2017proximal,
  title={Proximal policy optimization algorithms},
  author={Schulman, John and Wolski, Filip and Dhariwal, Prafulla and Radford, Alec and Klimov, Oleg},
  journal={arXiv preprint arXiv:1707.06347},
  year={2017}
}

@inproceedings{ouyang2022training,
  title={Training language models to follow instructions with human feedback},
  author={Ouyang, Long and Wu, Jeffrey and Jiang, Xu and Almeida, Diogo and Wainwright, Carroll and Mishkin, Pamela and Zhang, Chong and Agarwal, Sandhini and Slama, Katarina and Ray, Alex and others},
  booktitle={Proceedings of the 36th International Conference on Neural Information Processing Systems},
  pages={27730--27744},
  year={2022}
}

@book{
fodor1975language,
  author    = {Fodor, Jerry A.},
  title     = {{The Language Of Thought}},
  volume    = {5},
  publisher = {Harvard University Press},
  year      = {1975}
}

@inproceedings{
NeSyCoCo,
author = {Kamali, Danial and Barezi, Elham J. and Kordjamshidi, Parisa},
title = {{NeSyCoCo}: A neuro-symbolic concept composer for compositional generalization},
year = {2025},
booktitle = {Proceedings of the AAAI Conference on Artificial Intelligence},
}

@article{
turing_lecture_abstractions,
author = {Aho, Alfred and Ullman, Jeffrey},
title = {Abstractions, their algorithms, and their compilers},
year = {2022},
publisher = {Association for Computing Machinery},
volume = {65},
number = {2},

journal = {Communications of the ACM},
pages = {76--91},
}

@article{doi:10.1126/science.aab3050,
author = {Brenden M. Lake  and Ruslan Salakhutdinov  and Joshua B. Tenenbaum },
title = {Human-level concept learning through probabilistic program induction},
journal = {Science},
volume = {350},
number = {6266},
pages = {1332-1338},
year = {2015},

}

@article{
2022sharma,
  author  = {Utkarsh Sharma and Jared Kaplan},
  title   = {{Scaling laws from the data manifold dimension}},
  journal = {Journal of Machine Learning Research},
  year    = {2022},
  volume  = {23},
  number  = {9},
  pages   = {1--34},
}

@inproceedings{
boopathy2025breaking,
title={Breaking neural network scaling laws with modularity},
author={Akhilan Boopathy and Sunshine Jiang and William Yue and Jaedong Hwang and Abhiram Iyer and Ila R. Fiete},
booktitle={Proceedings of The Thirteenth International Conference on Learning Representations},
year={2025},
}

@inproceedings{
schug2024discovering,
title={Discovering modular solutions that generalize compositionally},
author={Simon Schug and Seijin Kobayashi and Yassir Akram and Maciej Wolczyk and Alexandra Maria Proca and Johannes von Oswald and Razvan Pascanu and Joao Sacramento and Angelika Steger},
booktitle={Proceedings of The Twelfth International Conference on Learning Representations},
year={2024},
}

@inproceedings{
uselis2025does,
title={{Does data scaling lead to visual compositional generalization?}},
author={Arnas Uselis and Andrea Dittadi and Seong Joon Oh},
booktitle={Proceedings of the Forty-second International Conference on Machine Learning},
year={2025},
}

@article{krakauer2025llmsemergence,
  title={{Large language models and emergence: A complex systems perspective}},
  author={Krakauer, David C. and Krakauer, John W. and Mitchell, Melanie},
  journal={arXiv preprint arXiv:2506.11135},
  year={2025}
}

@article{
Kumar2023DisentanglingAbstraction,
  title   = {{Disentangling abstraction from statistical pattern matching in human and machine learning}},
  author  = {Kumar, Sreejan and Dasgupta, Ishita and Daw, Nathaniel D. and Cohen, Jonathan D. and Griffiths, Thomas L.},
  journal = {PLOS Computational Biology},
  year    = {2023},
  volume  = {19},
  number  = {8},
  pages   = {e1011316},
}

@inproceedings{
ito2022compositional,
title={Compositional generalization through abstract representations in human and artificial neural networks},
author={Takuya Ito and Tim Klinger and Doug H. Schultz and John D. Murray and Michael Cole and Mattia Rigotti},
booktitle={Advances in Neural Information Processing Systems},
year={2022},
}

@inproceedings{
hsiang2025leandojov,
title={{LeanDojo-v2: A comprehensive library for AI-assisted theorem proving in Lean}},
author={Ryan Hsiang and William Adkisson and Robert Joseph George and Anima Anandkumar},
booktitle={Proceedings of The 5th Workshop on Mathematical Reasoning and AI at NeurIPS 2025},
year={2025},
}

@article{
2020_ai_feynmann,
author = {Silviu-Marian Udrescu  and Max Tegmark },
title = {{AI} {F}eynman: {A} physics-inspired method for symbolic regression},
journal = {Science Advances},
volume = {6},
number = {16},
pages = {eaay2631},
year = {2020},
}

@inproceedings{
gel88,
title={{The stable model semantics for logic programming}},
author={Michael Gelfond and Vladimir Lifschitz},
booktitle={Proceedings of International Logic Programming Conference and Symposium},
publisher={MIT Press},
pages={1070-1080},
year={1988}
}

@book{
paulson1994isabelle,
  title={{Isabelle}: A Generic Theorem Prover},
  author={Paulson, Lawrence C.},
  series={Lecture Notes in Computer Science},
  volume={828},
  year={1994},
  publisher={Springer}
}

@article{
paulinmohring1993coq,
  title={Synthesis of {ML} programs in the system {Coq}},
  author={Paulin-Mohring, Christine and Werner, Benjamin},
  journal={Journal of Symbolic Computation},
  volume={15},
  number={5--6},
  pages={607--640},
  year={1993}
}

@article{
singhal2025toward,
  author  = {Singhal, Karan and Tu, T. and Gottweis, J. and Sayres, R. and Wulczyn, E. and Amin, M. and Hou, L. and Clark, K. and Pfohl, S. R. and Cole-Lewis, H. and Neal, D. and Rashid, Q. M. and Schaekermann, M. and Wang, A. and Dash, D. and Chen, J. H. and Shah, N. H. and Lachgar, S. and Mansfield, P. A. and Prakash, S. and Green, B. and Dominowska, E. and Ag{\"u}era y Arcas, B. and Toma{\v{s}}ev, N. and Liu, Y. and Wong, R. and Semturs, C. and Mahdavi, S. S. and Barral, J. K. and Webster, D. R. and Corrado, G. S. and Matias, Y. and Azizi, S. and Karthikesalingam, A. and Natarajan, V.},
  title   = {Toward expert-level medical question answering with large language models},
  journal = {Nature Medicine},
  year    = {2025},
  volume  = {31},
  number  = {3},
  pages   = {943--950},
}

@article{
pmc_llama,
    author = {Wu, Chaoyi and Lin, Weixiong and Zhang, Xiaoman and Zhang, Ya and Xie, Weidi and Wang, Yanfeng},
    title = {{PMC-LLaMA}: Toward building open-source language models for medicine},
    journal = {Journal of the American Medical Informatics Association},
    volume = {31},
    number = {9},
    pages = {1833-1843},
    year = {2024},
}

@article{
shao2024deepseekmath,
  title={{DeepSeekMath}: Pushing the limits of mathematical reasoning in open language models},
  author={Shao, Zhihong and Wang, Peiyi and Zhu, Qihao and Xu, Runxin and Song, Junxiao and Bi, Xiao and Zhang, Haowei and Zhang, Mingchuan and Li, Y. K. and Wu, Yang and others},
  journal={arXiv preprint arXiv:2402.03300},
  year={2024}
}

@article{
xu2024survey,
  title={A survey on knowledge distillation of large language models},
  author={Xu, Xiaohan and Li, Ming and Tao, Chongyang and Shen, Tao and Cheng, Reynold and Li, Jinyang and Xu, Can and Tao, Dacheng and Zhou, Tianyi},
  journal={arXiv preprint arXiv:2402.13116},
  year={2024}
}

@article{
zhang2026instruction,
  title={Instruction tuning for large language models: A survey},
  author={Zhang, Shengyu and Dong, Linfeng and Li, Xiaoya and Zhang, Sen and Sun, Xiaofei and Wang, Shuhe and Li, Jiwei and Hu, Runyi and Zhang, Tianwei and Wang, Guoyin and others},
  journal={ACM Computing Surveys},
  volume={58},
  number={7},
  pages={1--36},
  year={2026},
  publisher={ACM New York, NY}
}

@article{pfister2025,
  title={{A representationalist, functionalist and naturalistic conception of intelligence as a foundation for AGI}},
  author={Pfister, Rolf},
  journal={arXiv preprint arXiv:2503.07600},
  year={2025}
}

@inproceedings{
wei2022cot,
author = {Wei, Jason and Wang, Xuezhi and Schuurmans, Dale and Bosma, Maarten and Ichter, Brian and Xia, Fei and Chi, Ed H. and Le, Quoc V. and Zhou, Denny},
title = {Chain-of-thought prompting elicits reasoning in large language models},
year = {2022},

booktitle = {Proceedings of the 36th International Conference on Neural Information Processing Systems},
}

@article{yao2023react,
  title={{ReAct: Synergizing reasoning and acting in language models}},
  author={Yao, Shunyu and Zhao, Jeffrey and Yu, Dian and Du, Nan and Shafran, Izhak and Narasimhan, Karthik and Cao, Yuan},
  journal={arXiv preprint arXiv:2210.03629},
  year={2023}
}

@inproceedings{
yao2023tot,
author = {Yao, Shunyu and Yu, Dian and Zhao, Jeffrey and Shafran, Izhak and Griffiths, Thomas L. and Cao, Yuan and Narasimhan, Karthik},
title = {Tree of thoughts: {D}eliberate problem solving with large language models},
year = {2023},
booktitle = {Proceedings of the 37th International Conference on Neural Information Processing Systems},
}

@article{zhou2023leasttomostprompting,
  title={{Least-to-most prompting enables complex reasoning in large language models}},
  author={Zhou, Denny and Sch{\"a}rli, Nathanael and Hou, Le and Wei, Jason and Scales, Nathan and Wang, Xuezhi and Schuurmans, Dale and Cui, Claire and Bousquet, Olivier and Le, Quoc and Chi, Ed},
  journal={arXiv preprint arXiv:2205.10625},
  year={2023}
}

@article{
goertzel_agi_concept,
author = {Goertzel, Ben},
year = {2009},
title = {{Artificial general intelligence: Concept, state of the art, and future prospects}},
journal = {Journal of Artificial General Intelligence},

}

@inbook{
Bennett_2025,
   title={What Is Artificial General Intelligence?},

   booktitle={Artificial General Intelligence},
   publisher={Springer Nature Switzerland},
   author={Bennett, Michael Timothy},
   year={2025},
   pages={30--42}
   }

@article{XU2025101370,
title = {Toward large reasoning models: {A} survey of reinforced reasoning with large language models},
journal = {Patterns},
volume = {6},
number = {10},
pages = {101370},
year = {2025},
author = {Fengli Xu and Qianyue Hao and Chenyang Shao and Zefang Zong and Yu Li and Jingwei Wang and Yunke Zhang and Jingyi Wang and Xiaochong Lan and Jiahui Gong and Tianjian Ouyang and Fanjin Meng and Yuwei Yan and Qinglong Yang and Yiwen Song and Sijian Ren and Xinyuan Hu and Jie Feng and Chen Gao and Yong Li},
}

@inproceedings{
Besta_2024,
   title={Graph of thoughts: {S}olving elaborate problems with large language models},
   volume={38},
   booktitle={Proceedings of the AAAI Conference on Artificial Intelligence},
   author={Besta, Maciej and Blach, Nils and Kubicek, Ales and Gerstenberger, Robert and Podstawski, Michal and Gianinazzi, Lukas and Gajda, Joanna and Lehmann, Tomasz and Niewiadomski, Hubert and Nyczyk, Piotr and Hoefler, Torsten},
   year={2024},
   pages={17682--17690} }

@article{patil2025advancingreasoningllms,
  title={{Advancing reasoning in large language models: Promising methods and approaches}},
  author={Patil, Avinash and Jadon, Aryan},
  journal={arXiv preprint arXiv:2502.03671},
  year={2025}
}

@inproceedings{
snell2025scaling,
title={{Scaling LLM test-time compute optimally can be more effective than scaling parameters for reasoning}},
author={Charlie Victor Snell and Jaehoon Lee and Kelvin Xu and Aviral Kumar},
booktitle={Proceedings of The Thirteenth International Conference on Learning Representations},
year={2025},
}

@inproceedings{
xie-etal-2024-efficient,
    title = {{Efficient continual pre-training for building domain specific large language models}},
    author = {Xie, Yong  and
      Aggarwal, Karan  and
      Ahmad, Aitzaz},
    booktitle = {Findings of the Association for Computational Linguistics: ACL 2024},
    year = {2024},
    pages = {10184--10201},
}

@article{
Wang2025PEFTSurvey,
  author       = {Wang, Luping and Chen, Sheng and Jiang, Linnan and Pan, Shu and Cai, Runze and Yang, Sen and Yang, Fei},
  title        = {Parameter-efficient fine-tuning in large language models: {A} survey of methodologies},
  journal      = {Artificial Intelligence Review},
  year         = {2025},
  volume       = {58},
  pages        = {227},
  note         = {Article number 227}
}

@article{
Ling2025_doaminspec,
author = {Ling, Chen and Zhao, Xujiang and Lu, Jiaying and Deng, Chengyuan and Zheng, Can and Wang, Junxiang and Chowdhury, Tanmoy and Li, Yun and Cui, Hejie and Zhang, Xuchao and Zhao, Tianjiao and Panalkar, Amit and Mehta, Dhagash and Pasquali, Stefano and Cheng, Wei and Wang, Haoyu and Liu, Yanchi and Chen, Zhengzhang and Chen, Haifeng and White, Chris and Gu, Quanquan and Pei, Jian and Yang, Carl and Zhao, Liang},
title = {{Domain specialization as the key to make large language models disruptive: A comprehensive survey}},
year = {2025},
volume = {58},
number = {3},
journal = {ACM Computing Surveys},
}

@inproceedings{
dziri2023faith,
title={{Faith and fate: Limits of transformers on compositionality}},
author={Nouha Dziri and Ximing Lu and Melanie Sclar and Xiang Lorraine Li and Liwei Jiang and Bill Yuchen Lin and Sean Welleck and Peter West and Chandra Bhagavatula and Ronan Le Bras and Jena D. Hwang and Soumya Sanyal and Xiang Ren and Allyson Ettinger and Zaid Harchaoui and Yejin Choi},
booktitle={Proceedings of the Thirty-seventh Conference on Neural Information Processing Systems},
year={2023},
}

@inproceedings{
patel-etal-2024-multi,
    title = "{Multi-LogiEval: Towards evaluating multi-step logical reasoning ability of large language models}",
    author = "Patel, Nisarg  and
      Kulkarni, Mohith  and
      Parmar, Mihir  and
      Budhiraja, Aashna  and
      Nakamura, Mutsumi  and
      Varshney, Neeraj  and
      Baral, Chitta",
    booktitle = "Proceedings of the Conference on Empirical Methods in Natural Language Processing",
    year = "2024",
    pages = "20856--20879"
}

@inproceedings{
sun2024thinkongraph,
title={{Think-on-Graph: Deep and responsible reasoning of large language model on knowledge graph}},
author={Jiashuo Sun and Chengjin Xu and Lumingyuan Tang and Saizhuo Wang and Chen Lin and Yeyun Gong and Lionel Ni and Heung-Yeung Shum and Jian Guo},
booktitle={Proceedings of The Twelfth International Conference on Learning Representations},
year={2024},
}

@inproceedings{
song2025a_survey,
title={{A survey on large language model reasoning failures}},
author={Peiyang Song and Pengrui Han and Noah Goodman},
booktitle={Proceedings of the 2nd AI for Math Workshop at ICML},
year={2025},
}

@inproceedings{
ICLR2024_0105f797,
 author = {Sharma, Mrinank and Tong, Meg and Korbak, Tomek and Duvenaud, David and Askell, Amanda and Bowman, Sam and Durmus, Esin and Hatfield-Dodds, Zac and Johnston, Scott and Kravec, Shauna and Maxwell, Timothy and McCandlish, Sam and Ndousse, Kamal and Rausch, Oliver and Schiefer, Nicholas and Yan, Da and Zhang, Miranda and Perez, Ethan},
 booktitle = {Proceedings of the International Conference on Learning Representations},
 pages = {110--144},
 title = {{Towards understanding sycophancy in language models}},
 volume = {2024},
 year = {2024}
}

@inproceedings{
2023rafailov_dpo,
author = {Rafailov, Rafael and Sharma, Archit and Mitchell, Eric and Ermon, Stefano and Manning, Christopher D. and Finn, Chelsea},
title = {Direct preference optimization: {Y}our language model is secretly a reward model},
year = {2023},
booktitle = {Proceedings of the 37th International Conference on Neural Information Processing Systems},
}

@article{
2023atlas,
  author  = {Gautier Izacard and Patrick Lewis and Maria Lomeli and Lucas Hosseini and Fabio Petroni and Timo Schick and Jane Dwivedi-Yu and Armand Joulin and Sebastian Riedel and Edouard Grave},
  title   = {{Atlas: Few-shot learning with retrieval-augmented language models}},
  journal = {Journal of Machine Learning Research},
  year    = {2023},
  volume  = {24},
  number  = {251},
  pages   = {1--43},
}

@article{
silver2016mastering,
  author  = {Silver, David and Huang, Aja and Maddison, Chris J. and Guez, Arthur and Sifre, Laurent and van den Driessche, George and Schrittwieser, Julian and Antonoglou, Ioannis and Panneershelvam, Veda and Lanctot, Marc and Dieleman, Sander and Grewe, Dominik and Nham, John and Kalchbrenner, Nal and Sutskever, Ilya and Lillicrap, Timothy and Leach, Madeleine and Kavukcuoglu, Koray and Graepel, Thore and Hassabis, Demis},
  title   = {Mastering the game of {G}o with deep neural networks and tree search},
  journal = {Nature},
  year    = {2016},
  volume  = {529},
  number  = {7587},
  pages   = {484--489},
}

@article{
silver2017mastering,
  author  = {Silver, David and Schrittwieser, Julian and Simonyan, Karen and Antonoglou, Ioannis and Huang, Aja and Guez, Arthur and Hubert, Thomas and Baker, Lucas and Lai, Matthew and Bolton, Adrian and Chen, Yutian and Lillicrap, Timothy and Hui, Fan and Sifre, Laurent and van den Driessche, George and Graepel, Thore and Hassabis, Demis},
  title   = {Mastering the game of {G}o without human knowledge},
  journal = {Nature},
  year    = {2017},
  volume  = {550},
  number  = {7676},
  pages   = {354--359},
}

@article{
doi:10.1126/science.aar6404,
author = {David Silver  and Thomas Hubert  and Julian Schrittwieser  and Ioannis Antonoglou  and Matthew Lai  and Arthur Guez  and Marc Lanctot  and Laurent Sifre  and Dharshan Kumaran  and Thore Graepel  and Timothy Lillicrap  and Karen Simonyan  and Demis Hassabis },
title = {A general reinforcement learning algorithm that masters chess, shogi, and Go through self-play},
journal = {Science},
volume = {362},
number = {6419},
pages = {1140-1144},
year = {2018},
}

@article{
abramson2024alphafold3,
  author  = {Abramson, Josh and Adler, Jonas and Dunger, Jack and Evans, Richard and Green, Tim and Pritzel, Alexander and Ronneberger, Olaf and Willmore, Lindsay and Ballard, Andrew J. and Bambrick, Joshua and Bodenstein, Sebastian W. and Evans, David A. and Hung, Chia-Chun and O'Neill, Michael and Reiman, David and Tunyasuvunakool, Kathryn and Wu, Zachary and {\v Z}emgulyt{\.e}, Akvil{\.e} and Arvaniti, Eirini and Beattie, Charles and Bertolli, Ottavia and Bridgland, Alex and Cherepanov, Alexey and Congreve, Miles and Cowen-Rivers, Alexander I. and Cowie, Andrew and Figurnov, Michael and Fuchs, Fabian B. and Gladman, Hannah and Jain, Rishub and Khan, Yousuf A. and Low, Caroline M. R. and Perlin, Kuba and Potapenko, Anna and Savy, Pascal and Singh, Sukhdeep and Stecula, Adrian and Thillaisundaram, Ashok and Tong, Catherine and Yakneen, Sergei and Zhong, Ellen D. and Zielinski, Michal and {\v Z}{\'i}dek, Augustin and Bapst, Victor and Kohli, Pushmeet and Jaderberg, Max and Hassabis, Demis and Jumper, John M.},
  title   = {Accurate structure prediction of biomolecular interactions with {AlphaFold}~3},
  journal = {Nature},
  year    = {2024},
  volume  = {630},
  number  = {8016},
  pages   = {493--500},

}

@article{
vinyals2019alphastar,
  author  = {Vinyals, Oriol and Babuschkin, Igor and Czarnecki, Wojciech M. and Mathieu, Micha{\"e}l and Dudzik, Andrew and Chung, Junyoung and Choi, David H. and Powell, Richard and Ewalds, Timo and Georgiev, Petko and Oh, Junhyuk and Horgan, Dan and Kroiss, Manuel and Danihelka, Ivo and Huang, Aja and Sifre, Laurent and Cai, Trevor and Agapiou, John P. and Jaderberg, Max and Vezhnevets, Alexander S. and Leblond, R{\'e}mi and Pohlen, Tobias and Dalibard, Valentin and Budden, David and Sulsky, Yury and Molloy, James and Paine, Tom L. and Gulcehre, Caglar and Wang, Ziyu and Pfaff, Tobias and Wu, Yuhuai and Ring, Roman and Yogatama, Dani and W{\"u}nsch, Dario and McKinney, Katrina and Smith, Oliver and Schaul, Tom and Lillicrap, Timothy and Kavukcuoglu, Koray and Hassabis, Demis and Apps, Chris and Silver, David},
  title   = {Grandmaster level in {StarCraft II} using multi-agent reinforcement learning},
  journal = {Nature},
  year    = {2019},
  volume  = {575},
  number  = {7782},
  pages   = {350--354},
}

@article{liang2025ai,
  title={{AI} reasoning in deep learning era: From symbolic {AI} to neural--symbolic {AI}},
  author={Liang, Baoyu and Wang, Yuchen and Tong, Chao},
  journal={Mathematics},
  volume={13},
  number={11},
  pages={1707},
  year={2025},
  publisher={MDPI}
}

@article{chervonyi2025gold,
  title={Gold-medalist performance in solving {Olympiad} geometry with {AlphaGeometry2}},
  author={Chervonyi, Yuri and Trinh, Trieu H. and Ol{\v{s}}{\'a}k, Miroslav and Yang, Xiaomeng and Nguyen, Hoang H. and Menegali, Marcelo and Jung, Junehyuk and Kim, Junsu and Verma, Vikas and Le, Quoc V. and others},
  journal={Journal of Machine Learning Research},
  volume={26},
  number={241},
  pages={1--39},
  year={2025}
}

@misc{deepmind2024_alphageometry,
  author       = {Trieu Trinh and Thang Luong and Google DeepMind},
  title        = {{AlphaGeometry: {A}n {O}lympiad-level {AI} System for Geometry}},
  year         = {2024},
  howpublished = {Google DeepMind blog. [Online.] Available: \url{https://deepmind.google/blog/alphageometry-an-olympiad-level-ai-system-for-geometry/}},
  note = {Accessed 2026-04-07}
}

@article{ren2025deepseek,
  title={{Deepseek-Prover-V2}: Advancing formal mathematical reasoning via reinforcement learning for subgoal decomposition},
  author={Ren, Z. Z. and Shao, Zhihong and Song, Junxiao and Xin, Huajian and Wang, Haocheng and Zhao, Wanjia and Zhang, Liyue and Fu, Zhe and Zhu, Qihao and Yang, Dejian and others},
  journal={arXiv preprint arXiv:2504.21801},
  year={2025}
}

@misc{alphaproof,
  author = {{Google DeepMind}},
  title = {{AI} achieves silver-medal standard solving {International Mathematical Olympiad} problems},
  howpublished = {[Online.] Available: \url{https://deepmind.google/blog/ai-solves-imo-problems-at-silver-medal-level/}},
  year = {2024},
  note = {Accessed 2026-04-07}
}

@article{yang2026formal_rlm_llms,
  title = {{Formal reasoning meets LLMs: Toward AI for mathematics and verification}},
  author = {Kaiyu Yang and Gabriel Poesia and Jingxuan He and Wenda Li and Kristin Lauter and Swarat Chaudhuri and Dawn Song},
  journal = {Communications of the ACM},
  year = {2026},
}

@article{lin2025goedel,
  title={{Goedel-Prover-V2}: Scaling formal theorem proving with scaffolded data synthesis and self-correction},
  author={Lin, Yong and Tang, Shange and Lyu, Bohan and Yang, Ziran and Chung, Jui-Hui and Zhao, Haoyu and Jiang, Lai and Geng, Yihan and Ge, Jiawei and Sun, Jingruo and others},
  journal={arXiv preprint arXiv:2508.03613},
  year={2025}
}

@article{liu2024deepseek,
  title={{DeepSeek-V3 technical report}},
  author={Liu, Aixin and Feng, Bei and Xue, Bing and Wang, Bingxuan and Wu, Bochao and Lu, Chengda and Zhao, Chenggang and Deng, Chengqi and Zhang, Chenyu and Ruan, Chong and others},
  journal={arXiv preprint arXiv:2412.19437},
  year={2024}
}

@inproceedings{moura2021lean,
  title={The {Lean} 4 theorem prover and programming language},
  author={Moura, Leonardo de and Ullrich, Sebastian},
  booktitle={Proceedings of the International Conference on Automated Deduction},
  pages={625--635},
  year={2021},
  organization={Springer}
}

@inproceedings{de2015lean,
  title={The {Lean} theorem prover (system description)},
  author={De Moura, Leonardo and Kong, Soonho and Avigad, Jeremy and Van Doorn, Floris and von Raumer, Jakob},
  booktitle={Proceedings of the International Conference on Automated Deduction},
  pages={378--388},
  year={2015},
  organization={Springer}
}

@inproceedings{bhagat2023sample,
  title={Sample-efficient learning of novel visual concepts},
  author={Bhagat, Sarthak and Stepputtis, Simon and Campbell, Joseph and Sycara, Katia},
  booktitle={Proceedings of the Conference on Lifelong Learning Agents},
  pages={637--657},
  year={2023},
  organization={PMLR}
}

@inproceedings{zhao2024physord,
  title={{PhysORD}: A neuro-symbolic approach for physics-infused motion prediction in off-road driving},
  author={Zhao, Zhipeng and Li, Bowen and Du, Yi and Fu, Taimeng and Wang, Chen},
  booktitle={Proceedings of the IEEE/RSJ International Conference on Intelligent Robots and Systems},
  pages={11670--11677},
  year={2024},
}

@inproceedings{zhang2024adaptable,
  title={Adaptable logical control for large language models},
  author={Zhang, Honghua and Kung, Po-Nien and Yoshida, Masahiro and Van den Broeck, Guy and Peng, Nanyun},
  booktitle={Proceedings of the 38th International Conference on Neural Information Processing Systems},
  pages={115563--115587},
  year={2024}
}

@inproceedings{west2022symbolic,
  title={Symbolic knowledge distillation: From general language models to commonsense models},
  author={West, Peter and Bhagavatula, Chandra and Hessel, Jack and Hwang, Jena and Jiang, Liwei and Le Bras, Ronan and Lu, Ximing and Welleck, Sean and Choi, Yejin},
  booktitle={Proceedings of the 2022 Conference of the North American Chapter of the Association for Computational Linguistics: Human Language Technologies},
  pages={4602--4625},
  year={2022}
}

@article{luo2025graph,
  title={{Graph-R1}: Towards agentic {GraphRAG} framework via end-to-end reinforcement learning},
  author={Luo, Haoran and Chen, Guanting and Lin, Qika and Guo, Yikai and Xu, Fangzhi and Kuang, Zemin and Song, Meina and Wu, Xiaobao and Zhu, Yifan and Tuan, Luu Anh and others},
  journal={arXiv preprint arXiv:2507.21892},
  year={2025}
}

@article{ma2024think,
  title={{Think-on-Graph} 2.0: Deep and faithful large language model reasoning with knowledge-guided retrieval augmented generation},
  author={Ma, Shengjie and Xu, Chengjin and Jiang, Xuhui and Li, Muzhi and Qu, Huaren and Yang, Cehao and Mao, Jiaxin and Guo, Jian},
  journal={arXiv preprint arXiv:2407.10805},
  year={2024}
}

@article{
belova2025graphmert,
title={Graph{MERT}: Efficient and scalable distillation of reliable knowledge graphs from unstructured data},
author={Margarita Belova and Jiaxin Xiao and Shikhar Tuli and Niraj Jha},
journal={Transactions on Machine Learning Research},
year={2026},
note={}
}

@article{luo2024graph,
  title={Graph-constrained reasoning: Faithful reasoning on knowledge graphs with large language models},
  author={Luo, Linhao and Zhao, Zicheng and Haffari, Gholamreza and Li, Yuan-Fang and Gong, Chen and Pan, Shirui},
  journal={arXiv preprint arXiv:2410.13080},
  year={2024}
}

@article{bader511042dimensions,
  title={Dimensions of neural-symbolic integration-a structured survey},
  year={2005},
  author={Bader, S. and Hitzler, P.},
  journal={arXiv preprint arXiv:cs/0511042}
}

@article{donadello2017logic,
  title={Logic tensor networks for semantic image interpretation},
  author={Donadello, Ivan and Serafini, Luciano and Garcez, Artur D'Avila},
  journal={arXiv preprint arXiv:1705.08968},
  year={2017}
}

@book{d2009neural,
  title={{Neural-Symbolic Cognitive Reasoning}},
  author={d’Avila Garcez, Artur S. and Lamb, Luis C. and Gabbay, Dov M.},
  year={2009},
  publisher={Springer}
}

@article{garcez2023neurosymbolic,
  title={Neurosymbolic {AI}: The 3rd wave},
  author={Garcez, Artur d’Avila and Lamb, Luis C.},
  journal={Artificial Intelligence Review},
  volume={56},
  number={11},
  pages={12387--12406},
  year={2023},
  publisher={Springer}
}

@article{graph_survey,
author = {Hogan, Aidan and Blomqvist, Eva and Cochez, Michael and D’amato, Claudia and Melo, Gerard De and Gutierrez, Claudio and Kirrane, Sabrina and Gayo, Jos\'{e} Emilio Labra and Navigli, Roberto and Neumaier, Sebastian and Ngomo, Axel-Cyrille Ngonga and Polleres, Axel and Rashid, Sabbir M. and Rula, Anisa and Schmelzeisen, Lukas and Sequeda, Juan and Staab, Steffen and Zimmermann, Antoine},
title = {{Knowledge graphs}},
year = {2021},
volume = {54},
number = {4},
journal = {ACM Computing Surveys},
publisher = {Association for Computing Machinery},
}

@inproceedings{yu2024cosmo,
  title={{COSMO}: A large-scale e-commerce common sense knowledge generation and serving system at {Amazon}},
  author={Yu, Changlong and Liu, Xin and Maia, Jefferson and Li, Yang and Cao, Tianyu and Gao, Yifan and Song, Yangqiu and Goutam, Rahul and Zhang, Haiyang and Yin, Bing and others},
  booktitle={Proceedings of the Companion of the International Conference on Management of Data},
  pages={148--160},
  year={2024}
}

@article{Jaradeh2023,
  author    = {Mohamed Yahya Jaradeh and Kuldeep Singh and Markus Stocker and Andreas Both and S{\"o}ren Auer},
  title     = {Information extraction pipelines for knowledge graphs},
  journal   = {Knowledge and Information Systems},
  year      = {2023},
  volume    = {65},
  number    = {5},
  pages     = {1989--2016},
}

@misc{brissette2024LLMKG,
  author    = {Rohan Rao and Benika Hall and Sunil Patel and Christopher Brissette and Gordana Neskovic},
  title     = {Insights, Techniques, and Evaluation for {LLM}‑Driven Knowledge Graphs},
  year      = {2024},
  organization = {NVIDIA Technical Blog},
  howpublished = {[Online.] Available: \url{https://developer.nvidia.com/blog/insights-techniques-and-evaluation-for-llm-driven-knowledge-graphs/}},
  note = {Accessed 2026-04-06}
}

@article{ns_methods_kg_reasoning,
author = {Cheng, Kewei and Ahmed, Nesreen K. and Rossi, Ryan A. and Willke, Theodore and Sun, Yizhou},
title = {{Neural-symbolic methods for knowledge graph reasoning: A survey}},
year = {2024},
volume = {18},
number = {9},
journal = {ACM Transactions on Knowledge Discovery from Data}
}

@inproceedings{reasoning_over_kg_with_logic_2025,
  title={{Improving complex reasoning over knowledge graph with logic-aware curriculum tuning}},
  author={Xia, Tianle and Ding, Liang and Wan, Guojia and Zhan, Yibing and Du, Bo and Tao, Dacheng},
  booktitle={Proceedings of the AAAI Conference on Artificial Intelligence},
  volume={39},
  pages={12881--12889},
  year={2025}
}

@article{LLM_for_kg_construction,
author = {Zhu, Yuqi and Wang, Xiaohan and Chen, Jing and Qiao, Shuofei and Ou, Yixin and Yao, Yunzhi and Deng, Shumin and Chen, Huajun and Zhang, Ningyu},
title = {{LLMs} for knowledge graph construction and reasoning: Recent capabilities and future opportunities},
year = {2024},
volume = {27},
number = {5},
journal = {World Wide Web}
}

@article{LI2025104769,
title = {{BiomedRAG}: A retrieval augmented large language model for biomedicine},
journal = {Journal of Biomedical Informatics},
volume = {162},
pages = {104769},
year = {2025},
author = {Mingchen Li and Halil Kilicoglu and Hua Xu and Rui Zhang}}

@inproceedings{cao-2021-knowledgeable,
  title={Knowledgeable or educated guess? {Revisiting} language models as knowledge bases},
  author={Cao, Boxi and Lin, Hongyu and Han, Xianpei and Sun, Le and Yan, Lingyong and Liao, Meng and Xue, Tong and Xu, Jin},
  booktitle={Proceedings of the 59th Annual Meeting of the Association for Computational Linguistics and the 11th International Joint Conference on Natural Language Processing (Volume 1: Long Papers)},
  pages={1860--1874},
  year={2021}
}

@inproceedings{mousavi-etal-2024-dyknow,
  title={{DyKnow: Dynamically verifying time-sensitive factual knowledge in LLMs}},
  author={Mousavi, Seyed Mahed and Alghisi, Simone and Riccardi, Giuseppe},
  booktitle={Findings of the Association for Computational Linguistics: EMNLP 2024},
  pages={8014--8029},
  year={2024}
}

@article{kim2025medicalhallucinations,
  title={{Medical hallucinations in foundation models and their impact on healthcare}},
  author={Kim, Yubin and Jeong, Hyewon and Chen, Shan and Li, Shuyue Stella and Lu, Mingyu and Alhamoud, Kumail and Mun, Jimin and Grau, Cristina and Jung, Minseok and Gameiro, Rodrigo and Fan, Lizhou and Park, Eugene and Lin, Tristan and Yoon, Joonsik and Yoon, Wonjin and Sap, Maarten and Tsvetkov, Yulia and Liang, Paul and Xu, Xuhai and Liu, Xin and McDuff, Daniel and Lee, Hyeonhoon and Park, Hae Won and Tulebaev, Samir and Breazeal, Cynthia},
  journal={arXiv preprint arXiv:2503.05777},
  year={2025}
}

@article{2024reversalcurse,
  title={The reversal curse: {LLMs} trained on {``A is B''} fail to learn {``B is A''}},
  author={Berglund, Lukas and Tong, Meg and Kaufmann, Max and Balesni, Mikita and Stickland, Asa Cooper and Korbak, Tomasz and Evans, Owain},
  journal={arXiv preprint arXiv:2309.12288},
  year={2024}
}

@inproceedings{min-factscore,
  title={{FActScore: Fine-grained atomic evaluation of factual precision in long form text generation}},
  author={Min, Sewon and Krishna, Kalpesh and Lyu, Xinxi and Lewis, Mike and Yih, Wen-Tau and Koh, Pang and Iyyer, Mohit and Zettlemoyer, Luke and Hajishirzi, Hannaneh},
  booktitle={Proceedings of the Conference on Empirical Methods in Natural Language Processing},
  year={2023}
}

@article{agents_multiagent_collab,
  title={Multi-agent collaboration mechanisms: A survey of {LLM}s},
  author={Tran, Khanh-Tung and Dao, Dung and Nguyen, Minh-Duong and Pham, Quoc-Viet and O'Sullivan, Barry and Nguyen, Hoang D.},
  journal={arXiv preprint arXiv:2501.06322},
  year={2025}
}

@article{agents_foundation_survey,
  title={Advances and challenges in foundation agents: From brain-inspired intelligence to evolutionary, collaborative, and safe systems},
  author={Liu, Bang and Li, Xinfeng and Zhang, Jiayi and Wang, Jinlin and He, Tanjin and Hong, Sirui and Liu, Hongzhang and Zhang, Shaokun and Song, Kaitao and Zhu, Kunlun and others},
  journal={arXiv preprint arXiv:2504.01990},
  year={2025}
}

@article{agents_coala,
  title={Cognitive architectures for language agents},
  author={Sumers, Theodore and Yao, Shunyu and Narasimhan, Karthik R. and Griffiths, Thomas L.},
  journal={Transactions on Machine Learning Research},
  year={2023}
}

@article{agents_masrouter,
  title={{MasRouter}: Learning to route {LLM}s for multi-agent systems},
  author={Yue, Yanwei and Zhang, Guibin and Liu, Boyang and Wan, Guancheng and Wang, Kun and Cheng, Dawei and Qi, Yiyan},
  journal={arXiv preprint arXiv:2502.11133},
  year={2025}
}

@article{agents_protocol_survey,
  title={A survey of {AI} agent protocols},
  author={Yang, Yingxuan and Chai, Huacan and Song, Yuanyi and Qi, Siyuan and Wen, Muning and Li, Ning and Liao, Junwei and Hu, Haoyi and Lin, Jianghao and Chang, Gaowei and others},
  journal={arXiv preprint arXiv:2504.16736},
  year={2025}
}

@article{agents_scientist_survey,
  title={{A survey of AI scientists}},
  author={Tie, Guiyao and Zhou, Pan and Sun, Lichao},
  journal={arXiv preprint arXiv:2510.23045},
  year={2025}
}

@article{agents_ai_scientist_sakana,
  title={The {AI} scientist: Towards fully automated open-ended scientific discovery},
  author={Lu, Chris and Lu, Cong and Lange, Robert Tjarko and Foerster, Jakob and Clune, Jeff and Ha, David},
  journal={arXiv preprint arXiv:2408.06292},
  year={2024}
}

@article{agents_self_evolving_survey,
  title={{A survey of self-evolving agents: What, when, how, and where to evolve on the path to artificial super intelligence}},
  author={Gao, Huan-ang and Geng, Jiayi and Hua, Wenyue and Hu, Mengkang and Juan, Xinzhe and Liu, Hongzhang and Liu, Shilong and Qiu, Jiahao and Qi, Xuan and Wu, Yiran and others},
  journal={Transactions on Machine Learning Research},
  year={2026}
}

@article{agents_embodied_systems,
  title={{Orchestrating embodied systems through the Embodied Context Protocol: Motivation, progress, and directions}},
  author={Ma, Fuyu and Li, Dong and Chen, Yizhe and Xing, Jiahui and Liu, Yu and Lan, Dapeng and Shao, Jinyan and Pang, Zhibo},
  journal={Research},
  volume={8},
  pages={1047},
  year={2025},
  publisher={AAAS}
}

@inproceedings{lightman2023let,
  title={Let's verify step by step},
  author={Lightman, Hunter and Kosaraju, Vineet and Burda, Yuri and Edwards, Harrison and Baker, Bowen and Lee, Teddy and Leike, Jan and Schulman, John and Sutskever, Ilya and Cobbe, Karl},
  booktitle={Proceedings of The Twelfth International Conference on Learning Representations},
  year={2024}
}

@inproceedings{agents_metagpt,
  title={{MetaGPT}: Meta programming for a multi-agent collaborative framework},
  author={Hong, Sirui and Zhuge, Mingchen and Chen, Jonathan and Zheng, Xiawu and Cheng, Yuheng and Wang, Jinlin and Zhang, Ceyao and Wang, Zili and Yau, Steven Ka Shing and Lin, Zijuan and others},
  booktitle={Proceedings of The Twelfth International Conference on Learning Representations},
  year={2024}
}

@inproceedings{agents_chatdev,
  title={{ChatDev}: Communicative agents for software development},
  author={Qian, Chen and Liu, Wei and Liu, Hongzhang and Chen, Nuo and Dang, Yufan and Li, Jiahao and Yang, Cheng and Chen, Weize and Su, Yusheng and Cong, Xin and others},
  booktitle={Proceedings of the 62nd Annual Meeting of the Association for Computational Linguistics (Volume 1: Long Papers)},
  pages={15174--15186},
  year={2024}
}

@article{agents_project_sid,
  title={Project {Sid}: Many-agent simulations toward {AI} civilization},
  author={AL, Altera and Ahn, Andrew and Becker, Nic and Carroll, Stephanie and Christie, Nico and Cortes, Manuel and Demirci, Arda and Du, Melissa and Li, Frankie and Luo, Shuying and others},
  journal={arXiv preprint arXiv:2411.00114},
  year={2024}
}

@book{agents_minsky_society,
  author    = {Minsky, Marvin},
  title     = {{The Society of Mind}},
  publisher = {Simon \& Schuster},
  year      = {1988}
}

@inproceedings{agents_reconcile,
  title={{ReConcile}: Round-table conference improves reasoning via consensus among diverse {LLM}s},
  author={Chen, Justin and Saha, Swarnadeep and Bansal, Mohit},
  booktitle={Proceedings of the 62nd Annual Meeting of the Association for Computational Linguistics (Volume 1: Long Papers)},
  pages={7066--7085},
  year={2024}
}

@inproceedings{agents_mad_debate,
  title={Encouraging divergent thinking in large language models through multi-agent debate},
  author={Liang, Tian and He, Zhiwei and Jiao, Wenxiang and Wang, Xing and Wang, Yan and Wang, Rui and Yang, Yujiu and Shi, Shuming and Tu, Zhaopeng},
  booktitle={Proceedings of the Conference on Empirical Methods in Natural Language Processing},
  pages={17889--17904},
  year={2024}
}

@inproceedings{agents_toolformer,
  title={Toolformer: Language models can teach themselves to use tools},
  author={Schick, Timo and Dwivedi-Yu, Jane and Dess{\`\i}, Roberto and Raileanu, Roberta and Lomeli, Maria and Hambro, Eric and Zettlemoyer, Luke and Cancedda, Nicola and Scialom, Thomas},
  booktitle={Proceedings of the 37th International Conference on Neural Information Processing Systems},
  pages={68539--68551},
  year={2023}
}

@article{agents_chemcrow,
  title={{ChemCrow}: Augmenting large-language models with chemistry tools},
  author={Bran, Andres M. and Cox, Sam and Schilter, Oliver and Baldassari, Carlo and White, Andrew D. and Schwaller, Philippe},
  journal={arXiv preprint arXiv:2304.05376},
  year={2023}
}

@article{agents_honeycomb,
  title={{HoneyComb}: A flexible {LLM}-based agent system for materials science},
  author={Zhang, Huan and Song, Yu and Hou, Ziyu and Miret, Santiago and Liu, Bang},
  journal={arXiv preprint arXiv:2409.00135},
  year={2024}
}

@article{agents_si_llm_ideas,
  title={Can {LLM}s generate novel research ideas? {A} large-scale human study with 100+ {NLP} researchers},
  author={Si, Chenglei and Yang, Diyi and Hashimoto, Tatsunori},
  journal={arXiv preprint arXiv:2409.04109},
  year={2024}
}

@article{agents_sciagents,
  author  = {Ghafarollahi, Alireza and Buehler, Markus J.},
  title   = {{SciAgents: Automating scientific discovery through multi-agent intelligent graph reasoning}},
  journal = {arXiv preprint arXiv:2409.05556},
  year    = {2024}
}

@article{agents_genesis,
  title={Genesis: Towards the automation of systems biology research},
  author={Tiukova, Ievgeniia A. and Brunns{\aa}ker, Daniel and Bjurstr{\"o}m, Erik Y. and Gower, Alexander H. and Kronstr{\"o}m, Filip and Reder, Gabriel K. and Reiserer, Ronald S. and Korovin, Konstantin and Soldatova, Larisa B. and Wikswo, John P. and others},
  journal={arXiv preprint arXiv:2408.10689},
  year={2024}
}

@article{agents_alphageometry,
  title={Solving {O}lympiad geometry without human demonstrations},
  author={Trinh, Trieu H. and Wu, Yuhuai and Le, Quoc V. and He, He and Luong, Thang},
  journal={Nature},
  volume={625},
  number={7995},
  pages={476--482},
  year={2024},
  publisher={Nature Publishing Group UK London}
}

@article{agents_chemagent,
  title={{ChemAgent}: Self-updating library in large language models improves chemical reasoning},
  author={Tang, Xiangru and Hu, Tianyu and Ye, Muyang and Shao, Yanjun and Yin, Xunjian and Ouyang, Siru and Zhou, Wangchunshu and Lu, Pan and Zhang, Zhuosheng and Zhao, Yilun and others},
  journal={arXiv preprint arXiv:2501.06590},
  year={2025}
}

@article{agents_agent_laboratory,
  title={Agent laboratory: Using {LLM} agents as research assistants},
  author={Schmidgall, Samuel and Su, Yusheng and Wang, Ze and Sun, Ximeng and Wu, Jialian and Yu, Xiaodong and Liu, Jiang and Moor, Michael and Liu, Zicheng and Barsoum, Emad},
  journal={Findings of the Association for Computational Linguistics: EMNLP 2025},
  pages={5977--6043},
  year={2025},
  publisher={Association for Computational Linguistics}
}

@article{agents_delocalized_discovery,
  title={Delocalized, asynchronous, closed-loop discovery of organic laser emitters},
  author={Strieth-Kalthoff, Felix and Hao, Han and Rathore, Vandana and Derasp, Joshua and Gaudin, Th{\'e}ophile and Angello, Nicholas H. and Seifrid, Martin and Trushina, Ekaterina and Guy, Mason and Liu, Junliang and others},
  journal={Science},
  volume={384},
  number={6697},
  pages={eadk9227},
  year={2024},
  publisher={American Association for the Advancement of Science}
}

@article{agents_chemos2,
  title={ChemOS 2.0: An orchestration architecture for chemical self-driving laboratories},
  author={Sim, Malcolm and Vakili, Mohammad Ghazi and Strieth-Kalthoff, Felix and Hao, Han and Hickman, Riley J. and Miret, Santiago and Pablo-Garc{\'\i}a, Sergio and Aspuru-Guzik, Al{\'a}n},
  journal={Matter},
  volume={7},
  number={9},
  pages={2959--2977},
  year={2024},
  publisher={Elsevier}
}

@article{agents_gmemory,
  title={{G-Memory: Tracing hierarchical memory for multi-agent systems}},
  author={Zhang, Guibin and Fu, Muxin and Wan, Guancheng and Yu, Miao and Wang, Kun and Yan, Shuicheng},
  journal={arXiv preprint arXiv:2506.07398},
  year={2025}
}

@inproceedings{agents_evoagent,
  title={{EvoAgent}: Towards automatic multi-agent generation via evolutionary algorithms},
  author={Yuan, Siyu and Song, Kaitao and Chen, Jiangjie and Tan, Xu and Li, Dongsheng and Yang, Deqing},
  booktitle={Proceedings of the Conference of the Nations of the Americas Chapter of the Association for Computational Linguistics: Human Language Technologies (Volume 1: Long Papers)},
  pages={6192--6217},
  year={2025}
}

@article{agent_mlat,
  title={{Machine learning as a tool (MLAT): A framework for integrating statistical ML models as callable tools within LLM agent workflows}},
  author={Chen, Edwin and Bibi, Zulekha},
  journal={arXiv preprint arXiv:2602.14295},
  year={2026}
}

@article{lifelong_learning,
  title={{Towards lifelong learning of large language models: A survey}},
  author={Zheng, Junhao and Qiu, Shengjie and Shi, Chengming and Ma, Qianli},
  journal={arXiv preprint arXiv:2406.06391},
  year={2024}
}

@article{cl_survey,
author = {Shi, Haizhou and Xu, Zihao and Wang, Hengyi and Qin, Weiyi and Wang, Wenyuan and Wang, Yibin and Wang, Zifeng and Ebrahimi, Sayna and Wang, Hao},
title = {{Continual learning of large language models: A comprehensive survey}},
year = {2025},
volume = {58},
number = {5},
journal = {ACM Computing Surveys},
}

@article{zheng2026lifelong,
  title={Lifelong learning of large language model based agents: A roadmap},
  author={Zheng, Junhao and Shi, Chengming and Cai, Xidi and Li, Qiuke and Zhang, Duzhen and Li, Chenxing and Yu, Dong and Ma, Qianli},
  journal={IEEE Transactions on Pattern Analysis and Machine Intelligence},
  year={2026},
}

@article{ai_hungry_benchmark_2025,
  title={How hungry is {AI}? {B}enchmarking energy, water, and carbon footprint of {LLM} inference},
  author={Jegham, Nidhal and Abdelatti, Marwan and Koh, Chan Young and Elmoubarki, Lassad and Hendawi, Abdeltawab},
  journal={arXiv preprint arXiv:2505.09598},
  year={2025}
}

@inproceedings{strubell2019energy,
  title={{Energy and policy considerations for deep learning in NLP}},
  author={Strubell, Emma and Ganesh, Ananya and McCallum, Andrew},
  booktitle={Proceedings of the 57th Annual Meeting of the Association for Computational Linguistics},
  pages={3645--3650},
  year={2019},
}

@article{kaplan2020scaling,
  title={{Scaling laws for neural language models}},
  author={Kaplan, Jared and McCandlish, Sam and Henighan, Tom and Brown, Tom B. and Chess, Benjamin and Child, Rewon and Gray, Scott and Radford, Alec and Wu, Jeffrey and Amodei, Dario},
  journal={arXiv preprint arXiv:2001.08361},
  year={2020},
}

@misc{pew_datacenter_energy,
  title={{What We Know about Energy Use at {US} Data Centers amid the {AI} Boom}},
  author={{Pew Research Center}},
  year={2025},
  howpublished = {[Online.] Available: \url{https://www.pewresearch.org/short-reads/2025/10/24/what-we-know-about-energy-use-at-us-data-centers-amid-the-ai-boom/}},
  note = {Accessed 2026-04-06}
}

@inproceedings{jevons_paradox_ai,
  title={From efficiency gains to rebound effects: The problem of {Jevons'} paradox in {AI}'s polarized environmental debate},
  author={Luccioni, Alexandra Sasha and Strubell, Emma and Crawford, Kate},
  booktitle={Proceedings of the ACM Conference on Fairness, Accountability, and Transparency},
  pages={76--88},
  year={2025}
}

@misc{staffing_ai_jobs,
  title={{Generative {AI} to Affect Blue-collar Jobs Less than White-collar Jobs}},
  author={{Staffing Industry Analysts}},
  year={2023},
  howpublished = {[Online.] Available: \url{https://www.staffingindustry.com/Editorial/Industrial-Staffing-Report/Dec.-21-2023/Generative-AI-to-affect-blue-collar-jobs-less-than-white-collar-jobs}},
  note = {Accessed 2026-04-06}
}

@misc{claude_pricing,
  title={Pricing -- {C}laude Documentation},
  author={{Anthropic}},
  year={2026},
  howpublished = {[Online.] Available: \url{https://platform.claude.com/docs/en/about-claude/pricing}},
  note = {Accessed 2026-04-06}
}

@misc{ier_orion_energy,
  title={Putting {AI}'s Insatiable Electricity Demand in Perspective},
  author={{Institute for Energy Research}},
  year={2025},
  howpublished = {[Online.] Available: \url{https://www.instituteforenergyresearch.org/the-grid/putting-ais-insatiable-electricity-demand-in-perspective/}},
  note = {Accessed 2026-04-06}
}

@misc{deepseek_cost,
  title={{DeepSeek}'s Low Inference Cost Explained: {MoE} and Strategy},
  author={{Intuition Labs}},
  year={2025},
  howpublished = {[Online.] Available: \url{https://intuitionlabs.ai/articles/deepseek-inference-cost-explained}},
  note = {Accessed 2026-04-06}
}

@misc{google_env_2025,
  title={2025 Environmental Report},
  author={{Google Sustainability}},
  year={2025},
  howpublished = {[Online.] Available: \url{https://sustainability.google/google-2025-environmental-report/}},
  note = {Accessed 2026-04-06}
}

@misc{ai_job_impact,
  title={{AI}'s Impact on Jobs: White-Collar, Blue-Collar, and Beyond},
  author={{LowTouch AI}},
  year={2025},
  howpublished = {[Online.] Available: \url{https://www.lowtouch.ai/ais-impact-on-jobs-white-collar-blue-collar-and-beyond/}},
  note = {Accessed 2026-04-06}
}

@misc{qualcomm_llama,
  title={{Llama-v3.1-8B-Instruct} -- {Q}ualcomm {AI} Hub},
  author={{Qualcomm}},
  year={2025},
  howpublished = {[Online.] Available: \url{https://aihub.qualcomm.com/models/llama_v3_1_8b_instruct}},
  note = {Accessed 2026-04-06}
}

@misc{snapdragon_8_elite,
  title={{Qualcomm's {S}napdragon 8 {E}lite Brings {O}ryon Cores to Mobile for the First Time}},
  author={{XDA Developers}},
  year={2024},
  howpublished = {[Online.] Available: \url{https://www.xda-developers.com/qualcomm-snapdragon-8-elite/}},
  note = {Accessed 2026-04-06}
}

@misc{apple_a18_pro,
  title={{A}pple {A18 Pro} Processor -- Benchmarks and Specs},
  author={{Notebookcheck}},
  year={2024},
  howpublished = {[Online.] Available: \url{https://www.notebookcheck.net/Apple-A18-Pro-Processor-Benchmarks-and-Specs.891556.0.html}},
  note = {Accessed 2026-04-06}
}

@misc{offline_ai_guide,
  title={Offline {AI} Assistant: The Definitive Guide to Choosing, Building and Deploying On-Device Intelligence},
  author={{A-bots}},
  year={2025},
  howpublished = {[Online.] Available: \url{https://a-bots.com/blog/Offline-AI-Assistant-Guide}},
  note = {Accessed 2026-04-06}
}

@misc{deepseek_r1_pricing,
  title={{DeepSeek} Model \& Pricing},
  author={{DeepSeek}},
  year={2026},
  howpublished = {[Online.] Available: \url{https://api-docs.deepseek.com/quick_start/pricing}},
  note = {Accessed 2026-04-06}
}

@misc{datacenter_cost_structure,
  title={A Look at the Cost Structure Igniting the {AI} Boom},
  author={{Alpha Matica}},
  year={2025},
  howpublished = {[Online.] Available: \url{https://www.alpha-matica.com/post/deconstructing-the-data-center-a-look-at-the-cost-structure-1}},
  note = {Accessed 2026-04-06}
}

@misc{jevons_paradox_digitopoly,
  title={Artificial Intelligence and the {J}evons Paradox},
  author={{Digitopoly}},
  year={2025},
  howpublished = {[Online.] Available: \url{https://digitopoly.org/2025/10/14/artificial-intelligence-and-the-jevons-paradox/}},
  note = {Accessed 2026-04-06}
}

@misc{deepseek_vs_gpt_leanware,
  title={{DeepSeek} vs. {GPT} o1 | Features, Performance, and Pricing},
  author={{Leanware}},
  year={2025},
  howpublished = {[Online.] Available: \url{https://www.leanware.co/insights/deepseek-r1-vs-gpt-o1}},
  note = {Accessed 2026-04-06}
}

@misc{ai_inference_costs_2025,
  title={{AI} Inference Costs in 2025: The \$255{B} Market's Energy Crisis and Path to Sustainable Scaling},
  author={{TensorMesh AI}},
  year={2025},
  howpublished = {[Online.] Available: \url{https://www.tensormesh.ai/blog-posts/ai-inference-costs-2025-energy-crisis}},
  note = {Accessed 2026-04-06}
}

@article{dss_graphrag_comparison,
  author    = {Peng, Boci and Zhu, Yun and Liu, Yongchao and Bo, Xiaohe and Shi, Haizhou and Hong, Chuntao and Zhang, Yan and Tang, Siliang},
  title     = {{Graph retrieval-augmented generation: A survey}},
  journal   = {arXiv preprint arXiv:2408.08921},
  year      = {2024},
}

@misc{dss_agent_protocols,
  title={{AI} Agents in Action: Foundations for Evaluation and Governance},
  author={{World Economic Forum}},
  year={2025},
  howpublished = {[Online.] Available: \url{https://reports.weforum.org/docs/WEF_AI_Agents_in_Action_Foundations_for_Evaluation_and_Governance_2025.pdf}},
  note = {Accessed 2026-04-06}
}

@article{dss_legal_ai,
  author    = {Koenecke, Allison and Stiglitz, Jed and Mimno, David and Wilkens, Matthew},
  title     = {{Tasks and roles in legal AI: Data curation, annotation, and verification}},
  journal   = {arXiv preprint arXiv:2504.01349},
  year      = {2025},
}

@article{dss_internet_of_agents,
  author    = {Chen, Weize and You, Ziming and Li, Ran and Guan, Yitong and Qian, Chen and Zhao, Chenyang and Yang, Cheng and Xie, Ruobing and Liu, Zhiyuan and Sun, Maosong},
  title     = {{Internet of agents: Weaving a web of heterogeneous agents for collaborative intelligence}},
  journal   = {arXiv preprint arXiv:2407.07061},
  year      = {2024},
}

@article{dss_graphrag_pharma,
  author    = {Edge, Darren and Trinh, Ha and Cheng, Newman and Bradley, Joshua and Chao, Alex and Mody, Apurva and Truitt, Steven and Metropolitansky, Dasha and Ness, Robert Osazuwa and Larson, Jonathan},
  title     = {{From local to global: A Graph RAG approach to query-focused summarization}},
  journal   = {arXiv preprint arXiv:2404.16130},
  year      = {2024},
}

@article{dss_graphrag_benchmark,
  author    = {Ahmad, Sarat and Nezami, Zeinab and Hafeez, Maryam and Zaidi, Syed Ali Raza},
  title     = {Benchmarking vector, graph and hybrid retrieval augmented generation},
  journal   = {arXiv preprint arXiv:2507.03608},
  year      = {2025},
}

@article{dss_proof_of_reasoning,
  author    = {Calo, James and Lo, Benny},
  title     = {Proof of reasoning for privacy enhanced federated blockchain learning at the edge},
  journal   = {arXiv preprint arXiv:2601.07134},
  year      = {2026},
}

@article{dss_neurosymbolic_audit,
  title={Neuro-symbolic {AI} for auditable cognitive information extraction from medical reports},
  author={Prenosil, George A. and Weitzel, Thilo K. and Bello, Sandra C. and Mingels, Clemens and Manzini, Giulia and Meier, Lorenz P. and Shi, Kuang-Yu and Rominger, Axel and Afshar-Oromieh, Ali},
  journal={Communications Medicine},
  volume={5},
  number={1},
  pages={491},
  year={2025},
  publisher={Nature Publishing Group UK London}
}

@article{dss_synthllm,
  author    = {Qin, Zeyu and Dong, Qingxiu and Zhang, Xingxing and Dong, Li and Huang, Xiaolong and Yang, Ziyi and Khademi, Mahmoud and Zhang, Dongdong and Awadalla, Hany Hassan and Fung, Yi R. and Chen, Weizhu and Cheng, Minhao and Wei, Furu},
  title     = {{Scaling laws of synthetic data for language models}},
  journal   = {arXiv preprint arXiv:2503.19551},
  year      = {2025},
}

@article{dss_textbooks_all_you_need,
  author    = {Gunasekar, Suriya and Zhang, Yi and Aneja, Jyoti and Mendes, Caio C{\'e}sar Teodoro and Del Giorno, Allie and Gopi, Sivakanth and Javaheripi, Mojan and Kauffmann, Piero and de Rosa, Gustavo and Saarikivi, Olli and Salim, Adil and Shah, Shital and Behl, Harkirat Singh and Wang, Xin and Bubeck, S{\'e}bastien and Eldan, Ronen and Kalai, Adam Tauman and Lee, Yin Tat and Li, Yuanzhi},
  title     = {{Textbooks are all you need}},
  journal   = {arXiv preprint arXiv:2306.11644},
  year      = {2023},
}

@misc{dss_train_ai_side_hustle,
  title={Get Paid to Train {AI}: Could Your Next Side Hustle Be Training {AI}?},
  author={{Built In}},
  year={2025},
  howpublished = {[Online.] Available: \url{https://builtin.com/articles/train-ai-side-hustle}},
  note = {Accessed 2026-04-06}
}

@misc{dss_expert_data,
  title={Why Expert Data Is Becoming the New Fuel for {AI} Models},
  author={{SignalFire}},
  year={2025},
  howpublished = {[Online.] Available: \url{https://www.signalfire.com/blog/expert-data-is-new-fuel-for-ai-models}},
  note = {Accessed 2026-04-06}
}

@misc{dss_data_labeling,
  title={How The {AI} Data Labeling Industry Works},
  author={{O-Mega AI}},
  year={2025},
  howpublished = {[Online.] Available: \url{https://o-mega.ai/articles/how-the-data-labeling-industry-works-full-insider-guide-2025}},
  note = {Accessed 2026-04-06}
}

@misc{dss_data_coalitions,
  title={Data Coalitions \& Escrow Agents},
  author={{RadicalxChange Foundation}},
  year={2023},
  howpublished = {[Online.] Available: \url{https://www.radicalxchange.org/updates/documents/data-coalitions-and-escrow-agents.pdf}},
  note = {Accessed 2026-04-06}
}

@article{dss_compute_divide,
  author    = {Ahmed, Nur and Wahed, Muntasir},
  title     = {The de-democratization of {AI}: Deep learning and the compute divide in artificial intelligence research},
  journal   = {arXiv preprint arXiv:2010.15581},
  year      = {2020},
}

@misc{dss_digital_divide,
  title={{AI Risks Sparking a New Era of Divergence as Development Gaps between Countries Widen, {UNDP} Report Finds}},
  author={{UNDP Asia-Pacific}},
  year={2025},
  howpublished = {[Online.] Available: \url{https://www.undp.org/asia-pacific/press-releases/ai-risks-sparking-new-era-divergence-development-gaps-between-countries-widen-undp-report-finds}},
  note = {Accessed 2026-04-06}
}

@article{dss_sovereign_ai,
  author    = {Singh, Shalabh Kumar and Sengupta, Shubhashis},
  title     = {{Sovereign AI: Rethinking autonomy in the age of global interdependence}},
  journal   = {arXiv preprint arXiv:2511.15734},
  year      = {2025},
}

@misc{dss_agriculture_ai,
  title={The Fourth Agricultural Revolution: {AI}'s Role in Feeding the Future},
  author={{Netafim}},
  year={2025},
  howpublished = {[Online.] Available: \url{https://www.netafim.com/en/blog/the-fourth-agricultural-revolution-ais-role-in-feeding-the-future/}},
  note = {Accessed 2026-04-06}
}

@misc{dss_indigenous_languages,
  title={Can Small Language Models Revitalize Indigenous Languages?},
  author={{Brookings Institution}},
  year={2025},
  howpublished = {[Online.] Available: \url{https://www.brookings.edu/articles/can-small-language-models-revitalize-indigenous-languages/}},
  note = {Accessed 2026-04-06}
}

@misc{dss_deepmind_materials,
  title={Millions of New Materials Discovered with Deep Learning},
  author={{Google DeepMind}},
  year={2023},
  howpublished = {[Online.] Available: \url{https://deepmind.google/blog/millions-of-new-materials-discovered-with-deep-learning/}},
  note = {Accessed 2026-04-06}
}

@article{Szymanski2026,
  author  = {Szymanski, Nathan J. and Rendy, Bernardus and Fei, Yuxing and Kumar, Rishi E. and He, Tanjin and Milsted, David and McDermott, Matthew J. and Gallant, Max and Cubuk, Ekin Dogus and Merchant, Amil and Kim, Haegyeom and Jain, Anubhav and Bartel, Christopher J. and Persson, Kristin and Zeng, Yan and Ceder, Gerbrand},
  title   = {Author Correction: An autonomous laboratory for the accelerated synthesis of inorganic materials},
  journal = {Nature},
  year    = {2026},
  month   = feb,
  volume  = {650},
  number  = {8100},
  pages   = {E1},
  url     = {https://doi.org/10.1038/s41586-025-09992-y}
}

@article{dss_autonomous_agents,
  author    = {Zhou, Lianhao and Ling, Hongyi and Fu, Cong and Huang, Yepeng and Sun, Michael and Yu, Wendi and Wang, Xiaoxuan and Li, Xiner and Su, Xingyu and Zhang, Junkai and Chen, Xiusi and Liang, Chenxing and Qian, Xiaofeng and Ji, Heng and Wang, Wei and Zitnik, Marinka and Ji, Shuiwang},
  title     = {{Autonomous agents for scientific discovery: Orchestrating scientists of the future}},
  journal   = {arXiv preprint arXiv:2510.09901},
  year      = {2025},
}

@misc{dss_knowledge_graphs_clarivate,
  title={What is a Knowledge Graph and how are they useful in pharma and life sciences?},
  author={{Clarivate}},
  year={2026},
  howpublished = {[Online.] Available: \url{https://clarivate.com/life-sciences-healthcare/consulting-services/research-and-development-consulting/knowledge-graphs/}},
  note = {Accessed 2026-04-06}
}

@misc{dss_hitl_future,
  title={Future of Human-in-the-Loop {AI} (2026): Emerging Trends \& Hybrid Automation Insights},
  author={{Parseur}},
  year={2025},
  howpublished = {[Online.] Available: \url{https://parseur.com/blog/future-of-hitl-ai}},
  note = {Accessed 2026-04-06}
}

@misc{dss_augmented_worker,
  title={Augmented Worker: How Digital Technology Can Power Your Workforce},
  author={{Tulip}},
  year={2026},
  howpublished = {[Online.] Available: \url{https://tulip.co/ebooks/augmented-worker/}},
  note = {Accessed 2026-04-06}
}
